\pdfoutput=1
\documentclass{article}

\usepackage[preprint,nonatbib]{neurips_2022}

\usepackage[hidelinks]{hyperref}
\usepackage{array}
\usepackage{tabularx}
\usepackage{wrapfig,lipsum}
\usepackage{url}            % simple URL typesetting
\usepackage{booktabs}       % professional-quality tables
\usepackage{amsfonts}       % blackboard math symbols
\usepackage{amsmath}
\usepackage{amssymb}
\usepackage{mathtools}
\usepackage{amsthm}
\usepackage{paralist, tabularx}
\usepackage{bm}
\usepackage{nicefrac}       % compact symbols for 1/2, etc.
\usepackage{microtype}      % microtypography
\usepackage[dvipsnames]{xcolor}% colors
\usepackage{multirow}
\theoremstyle{plain}

\theoremstyle{definition}

\theoremstyle{remark}

\usepackage[textsize=tiny]{todonotes}
\newcommand{\bx}{\bm{x}}
\newcommand{\by}{\bm{y}}
\newcommand{\bmu}{\bm{\mu}}
\newcommand{\bSigma}{\bm{\Sigma}}
\newcommand{\loss}{\mathcal{L}}

\newcommand{\abc}[1]{\textcolor{black}{#1}}
\newcommand{\ziquan}[1]{\textcolor{black}{#1}}
\setdefaultleftmargin{2em}{}{}{}{0.5em}{0.5em}
\usepackage{authblk}

\title{An Empirical Study on Distribution Shift Robustness From the Perspective of Pre-Training and Data Augmentation}

\author{\textbf{Ziquan Liu}$^1$\thanks{Work done during an internship at DAMO Academy, Alibaba Group.},  \textbf{Yi Xu}$^2$, \textbf{Yuanhong Xu}$^3$, \textbf{Qi Qian}$^3$, \textbf{Hao Li}$^3$, \textbf{Rong Jin}$^3$, \textbf{Xiangyang Ji}$^4$, \textbf{Antoni B. Chan}$^1$\\
$^1$Department of Computer Science, City University of Hong Kong \\
$^2$School of Artificial Intelligence, Dalian University of Technology \\
$^3$DAMO Academy, Alibaba Group \\
$^4$Department of Automation, Tsinghua University \par
\texttt{ziquanliu2-c@my.cityu.edu.hk}, \texttt{yxu@dlut.edu.cn}, \texttt{\{yuanhong.xuyh, qi.qian, lihao.lh, jinrong.jr\}@alibaba-inc.com},  \texttt{xyji@tsinghua.edu.cn}, \texttt{abchan@cityu.edu.hk}
}

\begin{document}

\maketitle

\begin{abstract}
  The performance of machine learning models under distribution shift has been the focus of the community in recent years. Most of current methods have been proposed to improve the robustness to distribution shift from the algorithmic perspective, i.e., designing better training algorithms to help the generalization in shifted test distributions. This paper studies the distribution shift problem from the perspective of pre-training and data augmentation, two important factors in the practice of deep learning that have not been systematically investigated by existing work. By evaluating seven pre-trained models, including ResNets \cite{he2016deep} and ViT's \cite{dosovitskiy2020image} with self-supervision and supervision mode, on five important distribution-shift datasets, from WILDS \cite{koh2021wilds} and DomainBed \cite{gulrajani2021in} benchmarks, with five different learning algorithms, we provide the first comprehensive empirical study focusing on pre-training and data augmentation. With our empirical result obtained from 1,330 models, we provide the following main observations: 1) ERM combined with data augmentation can achieve state-of-the-art performance if we choose a proper pre-trained model respecting the data property; 2) specialized algorithms further improve the robustness on top of ERM when handling a specific type of distribution shift, e.g., GroupDRO \cite{sagawa2019distributionally} for spurious correlation and CORAL \cite{sun2016deep} for large-scale out-of-distribution data; 3) Comparing different pre-training modes, architectures and data sizes, we provide novel observations about pre-training on distribution shift, which sheds light on designing or selecting pre-training strategy for different kinds of distribution shifts. In summary, our empirical study provides a comprehensive baseline for a wide range of pre-training models fine-tuned with data augmentation, which potentially inspires research exploiting the power of pre-training and data augmentation in the future of distribution shift study.
  %in the future of distribution shift study.
\end{abstract}
% Introduction to the problem of distribution shift. The severe consequence of non-robustness to distribution shift. 

% Algorithms that can handle OOD.

% Big picture of machine learning.

% The problem of OOD
\begin{table}[t]
\renewcommand{\arraystretch}{1.2}
    \centering
    \tiny
    \newcolumntype{S}{>{\centering\arraybackslash} m{0.72cm}}
    \newcolumntype{T}{>{\centering\arraybackslash} p{2.0cm}}
    \begin{tabular}{c|S|S|S|S|c|S|S|T}
    \hline
         &  \multicolumn{2}{c|}{\scriptsize WILDS-Waterbirds} & \multicolumn{2}{c|}{\scriptsize WILDS-FMoW} & \scriptsize WILDS-Camelyon & \multicolumn{2}{c|}{\scriptsize WILDS-iWildCam} & \scriptsize DomainNet \\
         \hline
         \small Training & \multicolumn{2}{c|}{\begin{minipage}{.09\textwidth}
      \includegraphics[width=\linewidth]{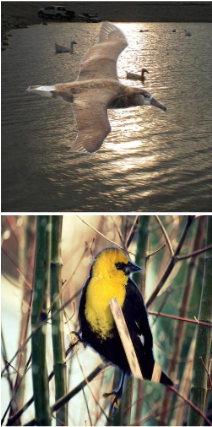}
    \end{minipage}} & \multicolumn{2}{c|}{\begin{minipage}{.09\textwidth}
      \includegraphics[width=\linewidth]{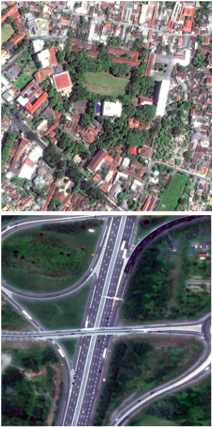}
    \end{minipage}} & \begin{minipage}{.172\textwidth}
      \includegraphics[width=\linewidth]{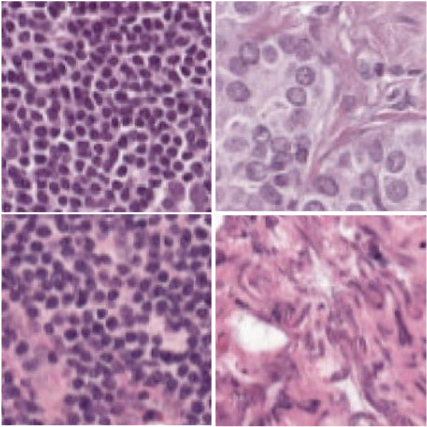}
    \end{minipage}& \multicolumn{2}{c|}{\begin{minipage}{.11\textwidth}
      \includegraphics[width=\linewidth]{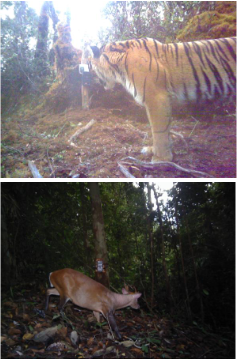}
    \end{minipage}}& \begin{minipage}{.13\textwidth}\vspace{2pt}
      \includegraphics[width=\linewidth]{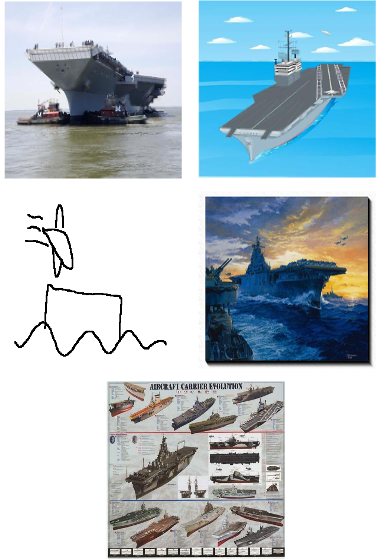}\vspace{2pt}
    \end{minipage}\\
    \hline
         \small Test & %\multicolumn{2}{m{2.4cm}|}{\includegraphics[width=0.5in]{figure/waterbird_test.png}}
         \multicolumn{2}{c|}{\begin{minipage}{.09\textwidth}\vspace{2pt}
      \includegraphics[width=\linewidth]{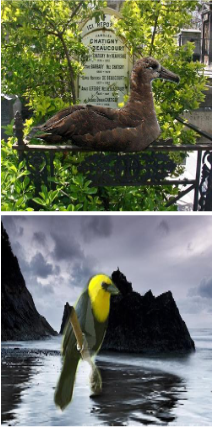}\vspace{2pt}
    \end{minipage}} 
    & \multicolumn{2}{c|}{\begin{minipage}{.09\textwidth}
      \includegraphics[width=\linewidth]{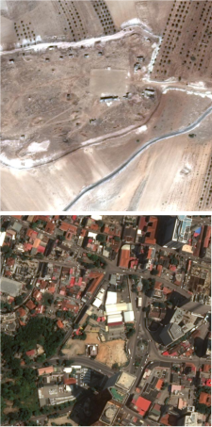}
    \end{minipage}} & \begin{minipage}{.172\textwidth}
      \includegraphics[width=\linewidth]{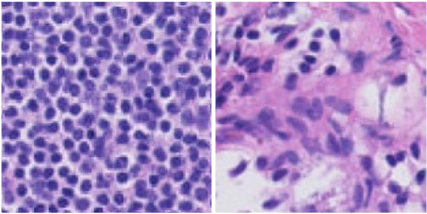}
    \end{minipage}& \multicolumn{2}{c|}{\begin{minipage}{.11\textwidth}
      \includegraphics[width=\linewidth]{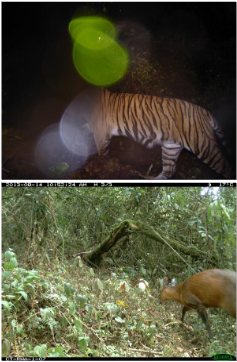}
    \end{minipage}}& \begin{minipage}{.1\textwidth}
      \includegraphics[width=\linewidth]{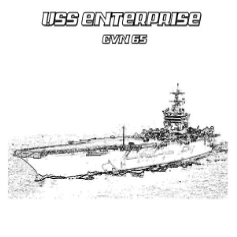}
    \end{minipage} \\
         \hline
    Metric & WG Acc. & Avg. Acc. & WG Acc. & Avg. Acc. & OOD Acc. & OOD Macro F1 & ID Macro F1 & Avg. Acc.  \\
        %\hline
         %Method & \multicolumn{2}{|c|}{GroupDRO} & \multicolumn{2}{|c|}{LISA} & MBDG & ERM & CADG \\
         \hline
         SotA & 91.4(1.1) \cite{sagawa2019distributionally} & 93.5(0.3) \cite{sagawa2019distributionally}& 35.5(0.8) \cite{yao2022improving} & 52.8(1.2) \cite{yao2022improving}& 93.3(1.0) \cite{robey2021model} & 38.5(0.6) \cite{miller2021accuracy} & 52.8(1.4) \cite{miller2021accuracy} &49.8(0.1) \cite{Dai2022CADGAM}\\
         \hline
         Our Best & \textbf{92.6(0.8)} & \textbf{94.1(0.7)} & \textbf{40.7(1.0)} & \textbf{57.4(2.1)} & \textbf{94.7(0.2)} & \textbf{43.2(0.9)} & 52.1(2.4) & \textbf{52.1(1.1)} \\
         \hline
         Our ERM Best & \textbf{92.6(0.8)} & \textbf{94.1(0.7)} &
         40.2(1.6) &
         57.1(1.0) &
         94.5(0.5) &
         41.3(2.4) & \textbf{55.7(2.1)} &
         49.8(0.1) \\
         \hline
    \end{tabular}
    \caption{The datasets with distribution shift evaluated in this paper. Spurious correlation (Waterbirds), subpopulation shift (FMoW) and out-of-distribution generalization (Camelyon, iWildCam and DomainNet) are considered. Our best result achieved by ERM combined with data augmentation matches state-of-the-art results on the five datasets, demonstrating the importance of selecting pre-trained models in improving distribution shift robustness. The state-of-the-art result is from \cite{domainnetbench} and \cite{wildsbench} for a single model without model averaging and ensemble. }
    \vspace{-0.9cm}
    \label{tab:datset}
\end{table}
\vspace{-0.3cm}
\section{Introduction}
\vspace{-0.3cm}
Machine learning (ML) has received much success in many computer vision applications such as image classification. However, it heavily depends on an in-distribution (ID) assumption that the training data and testing data are identically and independently from the same data distribution. Unfortunately, this ID assumption may be hardly satisfied in practice, which leads to distribution shifts. For example, an ML model used in self-driving cars may encounter surroundings that it hasn't seen in its training dataset. In this situation, generalizing to out-of-distribution (OOD) data becomes a challenging problem, leading to severe consequences if not well handled, because the models trained by traditional ML techniques will usually have degraded performance~\cite{arjovsky2019invariant,sagawa2019distributionally}. With the increasing use of models deployed on a data distribution that is different from its training data distribution, it is critical to evaluate and improve the performance under distribution shift. 

In the past several years, there has been a line of work proposed to study distribution shifts from different perspectives~\cite{arjovsky2019invariant,sagawa2019distributionally,krueger2021out,sun2016deep}. Arjovsky et al.~\cite{arjovsky2019invariant} proposed Invariant Risk Minimization (IRM) to learn invariant prediction across multiple training environments, which does not need the assumption of i.i.d. in training and testing. Follow-up works of IRM consider both theoretical and empirical aspects (e.g., see~\cite{ahuja2020invariant,rosenfeld2020risks,choe2020empirical,ahuja2020empirical}). On the other hand, robust optimization is a popular technique for distribution shift problems, which minimizes the worst-case loss. Examples include Distributionally Robust Optimization (DRO)~\cite{duchi2021learning,namkoong2016stochastic,sagawa2019distributionally},  minimax Risk Extrapolation (MM-REx)~\cite{krueger2021out}, and Heterogeneous Risk Minimization (HRM) ~\cite{liu2021heterogeneous}. However, most existing methods focus on developing new specialized learning algorithms to improve the generalization under distribution shifts, without considering deep learning (DL) characteristics such as training procedure and model architecture which are the important components to performance improvement in deep learning. 

In the paper, we empirically investigate the importance of \ziquan{pre-training} and model architectures for generalization under distribution shift, with a focus on image classification tasks. To this end, we consider Empirical Risk Minimization (ERM)~\cite{vapnik2013nature}, the most commonly-used learning paradigm in DL. We identify the source of naturally arising distribution shifts and study the impact of pre-training, which is carried out on a well-controlled training set without distribution shifts such as ImageNet \cite{ILSVRC15}, on downstream datasets with distribution shifts when fine-tuning with data augmentations. The contributions of our paper are:
\vspace{-0.2cm}
\begin{compactenum}
\item We provide the first extensive empirical study on how pre-training and data augmentation affect robustness to distribution shifts on various computer vision datasets. We consider seven pre-training models (including self-supervised and supervised ResNets and ViT’s), three data augmentations (including a group-aware mixup), and five datasets with distribution shifts (e.g., spurious correlation, subpopulation shift and OOD generalization), totaling 1,330 trained models. %We provide an extensive empirical study on how the pre-training and data augmentation affect robustness to distribution shifts in various computer vision datasets. Seven pre-training models and three data augmentations, including a group-aware mixup, are considered on five datasets with distribution shifts, e.g., spurious correlation, subpopulation shift and out-of-distribution generalization. 
\item We find that specialized algorithms such as GroupDRO \cite{sagawa2019distributionally} and CORAL \cite{sun2016deep} for distribution shifts achieve the state-of-the-art performance with proper pre-trained models on some datasets, while ERM with data augmentation is also a quite competitive baseline.
\item Comparing different pre-training strategies, models and data sizes, we summarize key observations about the impact of pre-training on distribution shifts, and provide practical tips on selecting pre-trained models under different types of distribution shifts.
\end{compactenum}

\vspace{-0.5cm}
\section{Related Work}
\vspace{-0.2cm}
\textbf{Spurious Correlation and Subpopulation Shift.} The spurious correlation problem, originated in statistics \cite{yule1926we,simon1954spurious}, is introduced to the general ML community by \cite{arjovsky2019invariant}. In fair machine learning, some works study the correlation between prediction and protected attributes to achieve \emph{demographic parity} \cite{zemel2013learning,hardt2016equality}. Subpopulation shift sometimes leads to spurious correlation, e.g., WILDS-Waterbirds, because the proportion of different groups reflects a correlation between protected and target attributes. There are several approaches to address the spurious correlation problem, such as invariant learning \cite{arjovsky2019invariant,ahuja2020invariant,creager2021environment,krueger2021out,liu2021integrated,liu2021heterogeneous}, causal learning \cite{peters2016causal,michael2021regularizing,mouli2022asymmetry} and distributionally robust optimization \cite{sagawa2019distributionally,oren2019distributionally}. In our work, we evaluate two datasets with spurious correlation and subpopulation shift, e.g., WILDS-Waterbirds and WILDS-FMoW \cite{koh2021wilds}, using ERM, GroupDRO \cite{sagawa2019distributionally} and CORAL \cite{sun2016deep}, and investigate the impact of pre-training and data augmentation on this problem. The experiment shows that GroupDRO is quite competitive in addressing spurious correlation, and CORAL is sometimes good at subpopulation shift data, while ERM with data augmentation is a strong baseline when using a proper pre-trained model.
%The spurious correlation problem, originated in statistics \cite{yule1926we,simon1954spurious}, is introduced to the general machine learning community by \cite{arjovsky2019invariant}. In fair machine learning, some works study the correlation between prediction and protected attributes to achieve \emph{demographic parity}~\cite{zemel2013learning,hardt2016equality}. Subpopulation shift sometimes leads to spurious correlation, e.g., WILDS-Waterbirds, because the proportion of different groups reflects correlation between protected and target attributes. There are several approaches to address the spurious correlation problem, such as invariant learning \cite{arjovsky2019invariant,ahuja2020invariant,creager2021environment,krueger2021out,liu2021integrated,liu2021heterogeneous}, causal learning \cite{peters2016causal,michael2021regularizing,mouli2022asymmetry} and distributionally robust optimization \cite{sagawa2019distributionally,oren2019distributionally}. In our work, we evaluate two datasets with spurious correlation and subpopulation shift, e.g., WILDS-Waterbirds and WILDS-FMoW \cite{koh2021wilds}, using ERM, GroupDRO \cite{sagawa2019distributionally} and CORAL \cite{sun2016deep}, and investigate the impact of pre-training and data augmentation on the problem. The experiment shows that GroupDRO are quite competitive in addressing spurious correlation and CORAL is sometimes good at subpopulation shift data, while ERM with data augmentation is a strong baseline when using a proper pre-trained model.

%and domain generalization \cite{sun2016deep,ganin2016domain}

\textbf{Out-of-Distribution Generalization.} The OOD generalization task assumes that the generation processes of training and test samples conditioned on the target label are different. For example, the environment or background of objects changes between training and test (WILDS-Camelyon, WILDS-iWildCam, ImageNet-C \cite{hendrycks2019benchmarking}) or the visual features of objects are changed during test (DomainNet, ImageNet-A \cite{hendrycks2021natural}). Various methods have been proposed to improve the generalization under OOD, including domain adversarial learning \cite{ganin2016domain,ganin2015unsupervised} and domain feature aligning \cite{sun2016deep,tzeng2014deep}. We propose to study the performance of different OOD generalization algorithms including the popular method CORAL \cite{ganin2016domain}, when different pre-trained models and data augmentation tricks are used. Similar to the finding for spurious correlation, different pre-trained models have substantially different impacts on OOD generalization. Key observations and tips about OOD generalization are provided based on our empirical results. Please refer to the survey~\cite{shen2021towards} for more methods on OOD generalization.
%The OOD generalization assumes that the generation processes of training and test samples conditioned on the target label are different. For example, the environment or background of objects changes from training to test (WILDS-Camelyon, WILDS-iWildCam, ImageNet-C \cite{hendrycks2019benchmarking}) and the visual features of objects are changed during test (DomainNet, ImageNet-A \cite{hendrycks2021natural}). Various methods have been proposed to improve the generalization under OOD condition, including domain adversarial learning \cite{ganin2016domain,ganin2015unsupervised} and domain feature aligning \cite{sun2016deep,tzeng2014deep}. We propose to study the performance of different OOD generalization algorithms including the popular domain generalization method CORAL \cite{ganin2016domain}, when different pre-trained models and data augmentation tricks are used. Similar to the finding in spurious correlation,different pre-trained models have a substantially different impacts on OOD generalization. Key observations and tips about OOD generalization are provided based on our empirical results. More methods for OOD generalization please refer to the survey paper~\cite{shen2021towards} and references therein. 

\textbf{The Impact of Pre-Training and Data Augmentation.} \cite{wiles2022fine} summarizes a framework for distribution shift, consisting of spurious correlation, low-data shift and unseen data, and evaluates the performance of representation learning, data augmentation and neural architectures. However, \cite{wiles2022fine} analyzed those factors \emph{independently} and only evaluated the pre-trained ResNet50. In contrast, our paper investigates the de facto training setting in distribution shift problem, i.e., fine-tuning a pre-trained model on a target task, and we comprehensively evaluate the performances of different pre-training models and neural architectures when combined with data augmentation and fine-tuning. \cite{gulrajani2021in} focuses on the role of data augmentation on domain generalization, where the visual features of objects are shifted during test such as DomainNet \cite{peng2019moment} (Fig.~\ref{tab:datset}), and draws a similar conclusion that ERM is a strong baseline in their context. Our work uses the largest dataset in \cite{gulrajani2021in} as representative of domain shift and studies the impact of pre-training in addition to the data augmentation. \cite{santurkar2021breeds} provides a nice benchmark on distribution shift consisting of different levels of semantic hierarchies with ImageNet \cite{ILSVRC15}. We do not evaluate on \cite{santurkar2021breeds} because our study focuses on the performance of pre-trained models on ImageNet in various downstream applications, instead of the subpopulation shift in ImageNet data.
%\cite{wiles2022fine} summarizes a framework for distribution shift, consisting of spurious correlation, low-data shift and unseen data, and evaluates the performance of representation learning, data augmentation and neural architectures. However, \cite{wiles2022fine} analyzed those factors \emph{independently} and only evaluated the pre-trained ResNet50. In contrast, our paper investigates the de facto training setting in distribution shift problem, i.e., fine-tuning a pre-trained model on a target task. The performances of different pre-training modes and neural architectures when combined with data augmentation in the fine-tuning are evaluated. \cite{gulrajani2021in} focuses on the role of data augmentation on domain generalization, where the visual features of objects are shifted during test such as DomainNet \cite{peng2019moment} (Fig.~\ref{tab:datset}), and draws a similar conclusion that ERM is a strong baseline in their context. Our work uses the largest dataset in \cite{gulrajani2021in} as a representative of domain shift and studies the impact of pre-training in addition to the data augmentation. \cite{santurkar2021breeds} provides a nice benchmark on distribution shift consisting of different level of semantic hierarchies with ImageNet \cite{ILSVRC15}. We do not evaluate on \cite{santurkar2021breeds} because our study focuses on the performance of pre-trained models on ImageNet in various downstream applications, instead of the subpopulation shift in ImageNet data. 

\vspace{-0.25cm}
\section{An Overview of Distribution Shift}
\vspace{-0.4cm}
We summarize two common sources of distribution shift in practice.
%s that often happen in practice. 
The first source is the bias in data, as a result of real-world bias. Assuming the data distribution is $p(\bx,\by)$ where $\by$ is an attribute vector, the conditional distribution of one attribute given another $p(y^i|y^j)$ may be high, indicating a spurious correlation between $y^i$ and $y^j$, e.g., the background and bird species in WILDS-Waterbirds. Some marginal distributions of $p(y^i)$ may be relatively small or large, leading to a subpopulation shift, e.g., the size of African images in WILDS-FMoW. The second source is the scarcity of data, i.e., the dataset cannot contain all possible images from the support of $p(\bx,\by)$ as a result of \ziquan{the high dimensionality of the continuous space}. In other words, the training set contains data from a certain distribution $p_{tr}(\bx|\by)$ but the test set have a different distribution $p_{te}(\bx|\by) \neq p_{tr}(\bx|\by)$, e.g., WILDS-iWildCam \cite{koh2021wilds} and DomainNet \cite{peng2019moment}. % It is worth noting 
Note that these two types of distribution shifts could be conceptually addressed by deliberately sampling a balanced dataset to avoid spurious correlation and subpopulation shift, and collecting all possible data from the distribution support. But in reality, we cannot obtain such a perfect dataset to train the model %so the problem exists 
in many real-world applications. A pre-trained model on a balanced large-scale dataset is often used as the initialization when training on such downstream tasks, because the pre-trained model learns good representations and benefits the optimization. Our work mainly investigates the impact of pre-training on distribution shift, and aims to answer the following question: How does the pre-training, i.e., training on the \ziquan{standard large-scale dataset that is more carefully curated}, affect the fine-tuning result on downstream datasets with distribution shift? To this end, we study different aspects of pre-training including training strategies, neural architectures and pre-training data sizes in the following Sections and summarize our findings in Section~\ref{sec:obs_tips}.

% The pre-training models experiment shows that more data helps the distribution shift. 

% Different datasets favor different neural architectures. So designing better models is also key to the distribution shift.

% Data augmentation generally improves the performance, indicating that generating more data in the fine-tuning process is helpful to the distribution shift robustness.
\begin{table}[t]
    \centering
    \footnotesize
    \vspace{-0.5cm}
    \begin{tabular}{c|c|c|c|c|c|c}
    \hline
              &  \multicolumn{2}{c|}{Waterbirds} & \multicolumn{2}{c|}{FMoW}& Camelyon & iWildCam \\
              \hline
        Metric & WG Acc. & Avg. Acc. & WG Acc. & Avg. Acc. & OOD Acc.  & OOD Macro F1 \\
        \hline
         ERM & 63.7(1.9) & \textbf{97.0(0.2)} & 	34.8(1.90) & 	\textbf{55.6(0.23)} & 	70.8(7.2) & 	32.0(1.5) \\
         GroupDRO & \textbf{91.4(1.1)} & 93.5(0.3) & 30.8(0.18)  & 	52.1(0.50)  & 68.4(7.3) & 23.9(2.1)  \\
         IRM & 67.4(5.2) & 73.4(9.7) & 30.0(1.37) & 50.8(0.13) & 64.2(8.1) & 15.1(4.9) \\
         %Fish & - & - & 34.6(0.18) & 51.8 (0.32) & 74.7(7.1) & 22.0(1.8)  \\
         CORAL & 79.4(1.9)&94.1(0.9) & 31.7(1.24) & 	50.5(0.36) & 59.5(7.7) & \textbf{32.8(0.1)}  \\
         \hline
         ERM+data aug  & 80.7(1.1) & 93.6(1.4) & 34.8(1.48) & 	55.4(0.52) & 82.0(7.4) & 	32.2(1.2)  \\
         ERM+Mixup & 82.6(2.5)&93.7(1.3) & 36.8(0.93)&55.0(0.73) & 91.8(0.7) & \textbf{32.8(1.4)} \\
         ERM+GroupMix & 85.0(2.3)&89.6(2.6) & \textbf{39.0(1.22)}&54.8(0.92) & \textbf{92.7(0.2)} & 31.9(0.9) \\
         \hline
        \end{tabular}
        \caption{WILDS benchmarks with data augmentation, mixup and GroupMix, using the default pre-trained model in each dataset. With mixup or GroupMix, the ERM outperforms GroupDRO, IRM and CORAL on FMoW, Camelyon and iWildCam.}
        \label{tab:original_wilds_bench}
        \vspace{-0.7cm}
\end{table}

\vspace{-0.3cm}
\section{Experimental Settings}
\vspace{-0.4cm}
\abc{In this section we discuss our experiment settings, including datasets, pre-trained models, and learning algorithms.} 
\vspace{-0.4cm}
\subsection{Datasets}
\vspace{-0.4cm}
To have a comprehensive evaluation on the effect of pre-training and data augmentation, we use five datasets that represent different types of distribution shifts. A general description of the five datasets follows. For more details about the datasets, see the original papers \cite{koh2021wilds} and \cite{peng2019moment}.

\textbf{WILDS-Waterbirds} \cite{koh2021wilds,sagawa2019distributionally} is a bird classification task and has two kinds of birds in two types of environments. Specifically, bird images from CUB \cite{wah2011caltech} are cropped and pasted to scene images from Places \cite{zhou2017places}. The training set of Waterbirds has 3,498 landbirds on land, 1,057 waterbirds on water, 184 landbirds on water and 56 waterbirds on land, inducing a spurious correlation in the data: the environment label (land/water) is correlated with the target label (waterbird/landbird). If the classical ERM is used, the environment features will be exploited and the performance of minor groups will be substantially worse than of major groups. Following the common practice \cite{sagawa2019distributionally,creager2021environment,liu2021just,zhou2021examining,krueger2021out}, we use the worse-group (WG) accuracy as a metric for robustness to spurious correlation and report the average accuracy weighted by the \emph{training} group size. %The input size is fixed to 224$\times$224.
%in all experiments.

\textbf{WILDS-FMoW} \cite{koh2021wilds,christie2018functional} is a satellite imagery data consisting of dense objects in the image. We use the satellite images to predict land usage as in \cite{koh2021wilds,shi2021gradient,kumar2022fine}. The images are divided into five groups according to their regions (Asia, Europe, Africa, Americas and Oceania). Due to historical and economic reasons, developing regions have fewer satellite images than developed regions. So in the FMoW data, the Africa region (1,582 training images) has much fewer images than Europe (34,816 training images) or Americas (20,973 training images). With the unbalanced groups and the fact that image features often vary across different continents, training with classical ERM may have a lower accuracy in minor groups compared with major groups, and has the risk of learning \ziquan{biased} models \ziquan{with discriminatory prediction}. %Similar to Waterbirds, we use worse-group accuracy as the metric on FMoW. The images are fixed to 224$\times$224.
%size across all experiments.

%{'train': tensor([17809., 34816.,  1582., 20973.,  1641.,     0.], device='cuda:0'), 'val': tensor([4121., 7732.,  803., 6562.,  693.,    0.], device='cuda:0'), 'test': tensor([4963., 5858., 2593., 8024.,  666.,    0.], device='cuda:0'), 'train_n_g': 6, 'val_n_g': 6, 'test_n_g': 6}
% region = Asia
% region = Europe
% region = Africa
% region = Americas
% region = Oceania
% region = Other

\textbf{WILDS-Camelyon} \cite{koh2021wilds,bandi2018detection} contains tissue slide images from five hospitals, and the task is to predict the presence of tumor tissue. The ways to collect and process the slide images differ among hospitals, while it is often desired that the trained model on limited data from a few hospitals generalizes to unseen data from a novel environment. Three hospitals are used for training, one hospital is for validation and the other is for test. %We follow \cite{koh2021wilds} and use 96$\times$96 as the input size during training. 

\textbf{WILDS-iWildCam} \cite{koh2021wilds,beery2020iwildcam} is collected by cameras in the wild, and the task is to predict wild animal species.
%and help the research on biodiversity and ecology. 
The data are split into 323 groups according to the location of cameras since the environment has an impact on the image features. The trained model on certain groups is expected to generalize to OOD, i.e., images from unseen or new cameras. As in \cite{koh2021wilds}, we use the Macro F1 score as the metric of the performance on iWildCam. %As in \cite{koh2021wilds}, we use 243, 32 and 48 groups for training, validation and test, and the images are resized to 448$\times$448. 

\textbf{DomainNet} \cite{peng2019moment} is often used to test domain generalization or adaptation, and consists of six domains that have the same categories but different image styles, e.g., painting, real and clipart. There are several other domain generalization datasets but we use the DomainNet because it is the largest dataset among its counterparts, e.g., PACS \cite{li2017deeper}, VLCS \cite{fang2013unbiased}, OfficeHome \cite{venkateswara2017deep} and TerraIncognita \cite{beery2018recognition}. We follow the common practice \cite{gulrajani2021in} to train a model on five domains and evaluate the performance on the held-out domain. Thus, we train six models since we repeat the process to test on each domain, and report the average results.
%and report the performance of each model and the average. 
%The image has an input size of 224$\times$224.

\vspace{-0.4cm}
\subsection{Pre-Trained Models} 
\vspace{-0.3cm}
The benchmarks in WILDS and DomainNet use a pre-trained ResNet \cite{he2016deep} on ImageNet-1k \cite{ILSVRC15} as the initialization for training. Specifically, Waterbirds and iWildCam use ResNet50 \cite{he2016deep}, FMoW and Camelyon use DenseNet121 \cite{huang2017densely}. In a domain generalization benchmark \cite{gulrajani2021in}, the pre-trained ResNet50 is the default model for DomainNet. In this work, the following seven pre-trained models are evaluated to study the impact of pre-training strategies, neural architectures and pre-training data size on the distribution shifts. 

\textbf{MoCo-R50 and MoCo-ViT-B/16} \cite{he2020momentum,chen2020improved} are self-supervised pre-trained models using contrastive learning to learn generic visual features. We use pre-trained ResNet50 (R50) and ViT-B/16 from \cite{chen2020improved}. 

\textbf{MAE-ViT-B/16} \cite{he2021masked} uses a masked image modeling task for self-supervised pre-training of the patch-based transformer architecture. We use the ViT-B/16 version of MAE. 

\textbf{Supervised ViT-B/16 and R50} that are trained on ImageNet-1k (IN-1k) and ImageNet-21k (IN-21k) are evaluated. \ziquan{There are 1,281,167 images and 1,000 classes in IN-1k, and 14,197,122 images and 21,841 classes in IN-21k.}
We use ViT-IN-1k and ViT-IN-21k model from \cite{steiner2021train}. The R50-IN-21k model is from \cite{ridnik2021imagenet}. Since all pre-trained ViT models have the same architecture, i.e., ViT-B/16, we use ViT for short in this paper. By comparing the performance of models pre-trained on 1k and 21k, we can investigate the impact of pre-training data scale on downstream distribution shifts. 

\begin{table}[t]
    \centering
    \small
    \vspace{-0.6cm}
    \begin{tabular}{c|c|c|c|c|c|c|c}
    \hline
            &  & & \multicolumn{3}{c|}{ERM} & \multirow{2}{*}{GroupDRO} & \multirow{2}{*}{CORAL} \\
              \cline{1-6}
        PT mode & Model & PT data & Data Aug. & Mixup & GroupMix &  &  \\
        \hline
         MoCo & ViT & IN-1k & 84.7(1.3) & 83.5(2.8) & 85.8(1.4) & \textbf{86.7(0.8)} & 83.3(1.9) \\
         MoCo & R50 & IN-1k & 82.5(0.8) & 86.7(0.7) & 87.7(0.7) & \textbf{88.1(0.8)} & 83.2(1.8) \\
         MAE & ViT & IN-1k & 81.7(2.0) & 80.3(1.5) & 82.2(2.5) & \textbf{87.4(1.0)} & 80.0(2.3) \\
         \hline
         \multirow{4}{*}{Sup.} & ViT & IN-1k & 76.3(2.4) & 79.5(4.5) & 84.1(1.9) & \textbf{87.0(1.1)} & 81.4(1.7) \\
          & R50 & IN-1k & 80.7(1.1) & 82.6(2.5) & 85.0(2.3) & \textbf{87.6(0.4)} & 79.4(1.9) \\
          & ViT & IN-21k & \textcolor{ForestGreen}{88.5(0.6)} & \textcolor{ForestGreen}{88.2(1.2)} & \textcolor{ForestGreen}{91.5(0.8)} & \textcolor{ForestGreen}{\textbf{92.6(0.5)}} & \textcolor{ForestGreen}{86.5(1.3)} \\
          & R50 & IN-21k & 82.9(1.4) & 86.5(1.3) & \textbf{87.5(1.6)} & 86.6(1.1) & 83.2(1.2) \\
          \hline
          \multicolumn{3}{c|}{Avg. Over Models} & 82.5 & 83.9 & 86.3 & \textbf{88.0} & 82.4\\
         \hline
        \end{tabular}
        \caption{WILDS-Waterbirds Result. We report the worse-group accuracy. The bold numbers are the best performance in the row and the green numbers are the best performance in a column. GroupDRO is a strong algorithm for spurious correlation and GroupMix substantially improves the DA and Mixup baseline. }
        \label{tab:wilds_waterbirds}
        \vspace{-0.5cm}
\end{table}
\begin{table}[t]
    \centering
    \small
    \vspace{-0.2cm}
    \begin{tabular}{c|c|c|c|c|c|c|c}
    \hline
            &  & & \multicolumn{3}{c|}{ERM} & \multirow{2}{*}{GroupDRO} & \multirow{2}{*}{CORAL} \\
              \cline{1-6}
        PT mode & Model & PT data & Data Aug. & Mixup & GroupMix &  &  \\
        \hline
         MoCo & ViT & IN-1k & 36.2(1.4) & \textbf{37.0(1.7)} & 36.5(1.5) & 35.2(1.6) & 35.8(1.7) \\
         MoCo & R50 & IN-1k & 36.4(2.3) & 36.0(1.2) & \textbf{37.1(1.4)} & 37.2(1.1) & 36.1(0.7) \\
         MAE & ViT & IN-1k & \textcolor{ForestGreen}{39.1(0.5)} & \textcolor{ForestGreen}{38.8(1.1)} & \textcolor{ForestGreen}{40.2(1.6)} & 37.9(1.4) & \textbf{40.4(0.8)} \\
         \hline
         \multirow{4}{*}{Sup.} & ViT & IN-1k & 34.5(0.8) & 34.4(0.3) & 35.8(0.7) & 35.8(1.8) & \textbf{35.9(1.2)} \\
          & R50 & IN-1k & 34.9(1.9) & 35.9(1.8) & \textbf{36.8(1.3)} & 36.1(1.6) & 34.2(1.2) \\
          & ViT & IN-21k & 37.9(1.3) & 38.5(0.9) & 38.6(1.5) & \textcolor{ForestGreen}{39.0(1.0)} & \textcolor{ForestGreen}{\textbf{40.7(1.0)}} \\
          & R50 & IN-21k & 37.2(2.9) & 38.4(0.6) & \textbf{39.0(0.8)} & 37.1(1.9) & 36.6(0.8) \\
         \hline
         \multicolumn{3}{c|}{Avg. Over Models} & 36.6 & 37.0& \textbf{37.7}& 36.9& 37.1\\
         \hline
        \end{tabular}
        \caption{WILDS-FMoW Result. The worse-group accuracy is reported. ERM is a quite competitive baseline and the self-supervised pre-trained model MAE is more suitable to the satellite imaging data than other models.}
        \label{tab:wilds_fmow}
        \vspace{-0.8cm}
\end{table}
\begin{table}[t]
    \centering
    \small
    \begin{tabular}{c|c|c|c|c|c|c|c}
    \hline
            &  & & \multicolumn{3}{c|}{ERM} & \multirow{2}{*}{GroupDRO} & \multirow{2}{*}{CORAL} \\
              \cline{1-6}
        PT mode & Model & PT data & Data Aug. & Mixup & GroupMix &  &  \\
        \hline
         MoCo & ViT & IN-1k & 89.9(2.1) & 92.6(0.7) & 92.6(0.9) & 89.7(2.0) & \textbf{92.7(0.4)} \\
         MoCo & R50 & IN-1k & 90.9(1.3) & \textbf{91.8(1.8)} & 90.6(2.9) & 91.2(1.6) & 88.2(4.6) \\
         MAE & ViT & IN-1k & 93.7(0.7) & \textcolor{ForestGreen}{94.4(0.4)} & \textcolor{ForestGreen}{94.5(0.5)} & \textcolor{ForestGreen}{94.4(0.2)} & \textcolor{ForestGreen}{\textbf{94.7(0.2)}} \\
         \hline
         \multirow{4}{*}{Sup.} & ViT & IN-1k & 90.1(1.2) & 92.2(1.8) & 93.3(0.8) & \textbf{93.3(0.5)} & 92.9(1.1) \\
         & R50 & IN-1k & 85.5(4.6) & 78.9(11.6) & 80.3(8.5) & \textbf{87.5(3.9)} & 78.9(11.6) \\
         & ViT & IN-21k & 93.9(0.7) & 93.5(0.7) & \textbf{94.2(0.3)} & 90.8(2.3) & 93.5(0.7) \\
         & R50 & IN-21k & \textcolor{ForestGreen}{\textbf{93.7(0.3)}} & 91.1(1.1) & 89.6(2.7) & 93.1(1.7) & 91.1(1.1) \\
         \hline
         \multicolumn{3}{c|}{Avg. Over Models} & 91.1& 90.6& 90.7& \textbf{91.4}& 90.3\\
         \hline
        \end{tabular}
        \caption{WILDS-Camelyon Result. We report the OOD average accuracy. GroupMix and CORAL perform the best on Camelyon and MAE is the best pre-trained model for the tissue slide data in most cases. }
        \label{tab:wilds_camelyon}
        \vspace{-0.7cm}
\end{table}

\vspace{-0.4cm}
\subsection{\abc{Learning} Algorithms} 
\vspace{-0.2cm}
\textbf{ERM} is the classical objective function used in ML. We note that in stochastic gradient descent, the ERM is related to batch sampling. On Waterbirds and DomainNet, we follow the common practice \cite{sagawa2019distributionally,gulrajani2021in} to use the weighted sampling so that each batch has the same number of samples for all groups. On other datasets, the standard batch sampling is used. 

\textbf{Data augmentation}. On Waterbirds, we use the standard data augmentation, i.e., random resizing and cropping, and random horizontal flipping. On Camelyon and FMoW, we use the 2-level of randaug \cite{cubuk2020randaug}. On iWildCam, the 1-step randaug is used. On DomainNet, we use the standard data augmentation in Waterbirds, along with color jittering and random grayscale.

% transforms.RandomResizedCrop(224, scale=(0.7, 1.0)),
%             transforms.RandomHorizontalFlip(),
%             transforms.ColorJitter(0.3, 0.3, 0.3, 0.3),
%             transforms.RandomGrayscale(),

\textbf{Mixup} \cite{zhang2017mixup} optimizes a convex hull of training samples to improve the generalization, by augmenting the training data with random combination of input images and labels. Specifically, two samples $\bx_1$ and $\bx_2$ with labels $y_1$ and $y_2$ will generate a new training sample as 
%. The mixup is done as 
follows
\setlength{\belowdisplayskip}{1.8pt} \setlength{\belowdisplayshortskip}{1.8pt}
\setlength{\abovedisplayskip}{1.9pt} \setlength{\abovedisplayshortskip}{1.9pt}
\footnotesize
\begin{align}
    \lambda &\sim Beta(\alpha,\beta), \quad \bx' = \lambda  \bx_1 + (1-\lambda)  \bx_2,\quad y'= \lambda y_1+(1-\lambda) y_2,
\end{align}
\normalsize
where $\lambda$ is sampled from a Beta distribution with parameters %, 
$\alpha$ and $\beta$, 
%is the parameter in the Beta distribution controlling the randomness of mixing, 
%$f(\cdot)$ denotes the neural network and $XE$ represents the cross-entropy loss. 
%The mixed pair $(\bx',y')$ are input to the network and trained directly.

\textbf{Group-Based Mixup.} Motivated by GroupDRO, we investigate a variant of mixup which incorporates the group information into the mixing steps. Here is a generalized mixup with group information,
\setlength{\belowdisplayskip}{1.8pt} \setlength{\belowdisplayshortskip}{1.8pt}
\setlength{\abovedisplayskip}{1.9pt} \setlength{\abovedisplayshortskip}{1.9pt}
\footnotesize
\begin{align}
    \lambda &\sim Beta(g_1^{(b)}\alpha,g_2^{(b)}\beta), \quad \bx'= g_1^{(x)}\lambda  \bx_1 + g_2^{(x)}(1-\lambda)  \bx_2, \quad y' = g^{(l)}_1\lambda  y_1 + g^{(l)}_2(1-\lambda)  y_2
%    , l = l(\bx',y'),
\end{align}
\normalsize
where $g^{(b)},g^{(x)},g^{(l)}$ denote the group-related weight for beta distribution sampling, input and label. The group-related weight is computed from group weight $g(\bx_i)$, which is related to the size of group $\bx_i$ belong to. Specifically for group $j$, the group weight $g_j=\text{softmax}(C/\sqrt{n_j})$ and we use the function $g(\cdot)$ to denote $g_j$ for $\bx_i\in G_j$. We use one simple version of GroupMix, where $g_1^{(b)}=g_2^{(b)}=1.0$, $g_1^{(x)}=g_2^{(x)}=1.0$ and $g^{(l)}_1=g(\bx_1),g^{(l)}_2=g(\bx_2)$, if not mentioned otherwise. In Section~\ref{sec:variant_groupmix}, we evaluate five different variants of GroupMix. 

% 1) $g_1=g_2=1.0$, $g'_1=g'_2=1.0$, $g^*_1=g(\bx_1),g^*_2=g(\bx_2)$, 92.2(0.5).\\
% 2) $g_1=g(\bx_1),g_2=g(\bx_2)$, $g'_1=g'_2=g^*_1=g^*_2=1.0$, 92.0(1.1)\\
% 3) $g_1=g(\bx_1),g_2=g(\bx_2)$, $g'_1=g'_2=1.0$, $g^*_1=g(\bx_1),g^*_2=g(\bx_2)$, normalize the weight for batch, 92.6(0.7)\\
% 4) $g_1=g_2=1.0$, $g'_1=g'_2=1.0$, $g^*_1=g(\bx_1),g^*_2=g(\bx_2)$, normalize the weight for batch, 91.9(0.9)\\
% 5) Try $g_1=g_2=1.0$, $g'_1=g^*_1=g(\bx_1)$, $g'_1=g^*_1=g(\bx_1)$, normalize the weight for instance, 91.35\\
% 6) Try $g_1=g(\bx_1),g_2=g(\bx_2)$, $g'_1=g^*_1=g(\bx_1)$, $g'_1=g^*_1=g(\bx_1)$, normalize the weight for instance,  91.3(1.1)\\

\textbf{GroupDRO} \cite{sagawa2019distributionally}  collects group losses from past SGD iterations, based on which the current iteration's losses are re-weighted. 

\textbf{CORAL} \cite{sun2016deep} minimizes a distance measure of the mean and covariance estimates of features from two different groups, so that the learned features are invariant to domain or distribution shift. 
\vspace{-0.2cm}
\subsection{Experimental Settings}
\vspace{-0.2cm}
We use the official training/validation/testing split for WILDS datasets. For DomainNet, if a domain is used for training, then the data is randomly split into 80\% and 20\% as training and validation set following \cite{gulrajani2021in}. As the de facto setting in WILDS and DomainBed benchmark, the training starts with a pre-trained model and optimizes all parameters with the training data. In other words, we study the performance of fine-tuning a pre-trained model on distribution-shift datasets. The training epochs/steps are set as the same in the WILDS \cite{koh2021wilds} or DomainBed \cite{gulrajani2021in} benchmark. On WILDS datasets, we use the official (out-of-distribution) validation set to do the model selection. On DomainNet, we use the training domain validation set in the model selection, as suggested by \cite{gulrajani2021in}. In Table \ref{tab:original_wilds_bench}, the optimizer is the same as the benchmark. In all other experiments, AdamW \cite{loshchilov2018decoupled} is used as the optimizer. We search the hyperparameters for each setting by the grid search, where the parameter space is determined by either existing training practice or our pilot study on those datasets and models. For the search space for each dataset, please see the appendix. In total, we trained about 3000 models during the hyperparameter searching. Following the common practice \cite{koh2021wilds,gulrajani2021in}, on WILDS datasets, we report the mean and standard deviation of 5 trials; on DomainNet, we report the result of 3 trials.

%Mixup augmentation. Group-based mixup. Besides the data augmentation, we also evaluate two strong baselines in WILDS and DomainNet, i.e., GroupDRO and CORAL. 

\begin{table}[t]
    \centering
    \small
    \vspace{-0.3cm}
    \begin{tabular}{c|c|c|c|c|c|c|c}
    \hline
            &  & & \multicolumn{3}{c|}{ERM} & \multirow{2}{*}{GroupDRO} & \multirow{2}{*}{CORAL} \\
              \cline{1-6}
        PT mode & Model & PT data & Data Aug. & Mixup & GroupMix &  &  \\
        \hline
         MoCo & ViT & IN-1k & 34.0(0.5) & 31.7(2.1) & 32.8(0.7) & 13.5(1.0) & \textbf{35.2(0.6)} \\
         MoCo & R50 & IN-1k & 34.5(1.5) & 36.8(1.6) & 36.0(1.1) & 20.8(0.8) & \textbf{37.2(1.1)} \\
         MAE & ViT & IN-1k & 28.4(2.0) & 27.3(2.2) & 28.0(2.9) & 9.2(0.9) & \textbf{31.4(2.6)} \\
         \hline
         \multirow{4}{*}{Sup.} & ViT & IN-1k & 38.9(0.7) & 41.0(0.7) & 40.9(1.2) & 20.5(0.5) & \textbf{41.9(0.9)} \\
         & R50 & IN-1k & 32.2(1.2) & 32.8(1.4) & 31.9(0.9) & 18.9(0.8) & \textbf{33.4(0.5)} \\
         & ViT & IN-21k & 39.0(2.7) & \textcolor{ForestGreen}{\textbf{41.3(2.4)}} & 41.1(1.4) & 19.0(1.2) & 36.4(1.6) \\
         & R50 & IN-21k & \textcolor{ForestGreen}{40.9(1.3)} & 40.8(0.4) & \textcolor{ForestGreen}{41.2(1.3)} & \textcolor{ForestGreen}{24.0(0.8)} & \textcolor{ForestGreen}{\textbf{43.2(0.9)}} \\
         \hline
         \multicolumn{3}{c|}{Avg. Over Models} & 35.4 & 36.0 & 36.0 & 18.0 &\textbf{37.0}\\
         \hline
        \end{tabular}
        \caption{WILDS-iWildCam Result. The macro F1 score of OOD data is reported as in WILDS benchmark. CORAL is the best algorithm in most cases while GroupMix is also a strong baseline. In the wild recognition task, supervised pre-training is generally better than the  self-supervised counterpart.}
        \label{tab:wilds_iwildcam}
        \vspace{-0.6cm}
\end{table}

\begin{table}[t]
    \centering
    \small
    \begin{tabular}{c|c|c|c|c|c|c|c}
    \hline
            &  & & \multicolumn{3}{c|}{ERM} & \multirow{2}{*}{GroupDRO} & \multirow{2}{*}{CORAL} \\
              \cline{1-6}
        PT mode & Model & PT data & Data Aug. & Mixup & GroupMix &  &  \\
        \hline
         MoCo & ViT & IN-1k & \textbf{47.5(0.1)} & 46.7(0.1) & 46.3(0.2) & 37.6(0.0) & 46.2(1.1) \\
         MoCo & R50 & IN-1k & \textbf{44.2(0.1)} & 41.6(0.1) & 41.3(0.2) & 35.9(0.1) & 42.6(0.3) \\
         MAE & ViT & IN-1k & \textbf{43.6(0.2)} & 41.6(0.5) & 41.5(0.6) & 31.7(0.1) & 43.5(0.1) \\
         \hline
         \multirow{4}{*}{Sup.} & ViT & IN-1k & 47.7(0.1) & 47.3(0.1) & 47.3(0.2) & 42.4(0.0) & \textbf{48.0(0.0)} \\
         & R50 & IN-1k & 40.9(0.1) & 39.2(0.1) & 37.3(0.2) & 33.3(0.2) & \textbf{41.5(0.1)} \\
         & ViT & IN-21k & 47.4(0.2) & 47.8(0.2) & 47.6(0.3) & 40.8(0.2) & \textcolor{ForestGreen}{\textbf{52.1(1.1)}} \\
         & R50 & IN-21k & \textcolor{ForestGreen}{\textbf{49.8(0.1)}} & \textcolor{ForestGreen}{48.5(0.1)} & \textcolor{ForestGreen}{48.2(0.2)} & \textcolor{ForestGreen}{43.1(0.0)} & 48.8(0.1) \\
         \hline
         \multicolumn{3}{c|}{Avg. Over Models} & \textbf{45.8}& 44.7& 44.2& 37.8& 45.7\\
         \hline
        \end{tabular}
        \caption{Experimental Results on DomainNet. The six domains are set as held-out target domain respectively and the averaged accuracy of six experimental results is reported. On the object recognition data, supervised pre-training on IN-21k triumphs over other models.}
        \label{tab:domainnet}
        \vspace{-0.7cm}
\end{table}

\begin{table}[t]
    \centering
    \small
    \newcolumntype{S}{>{\centering\arraybackslash} m{1.0cm}}
    \vspace{-0.6cm}
    \begin{tabular}{c|c|S|S|S|S|c}
    \hline
     \multirow{2}{*}{PT Model}&\multirow{2}{*}{Metric}& \multicolumn{4}{c|}{GroupMix} & \multirow{2}{*}{GroupDRO} \\
     \cline{3-6}
            & & V1 & V2 & V3 & V4 & \\
        \hline
        \multirow{2}{*}{Sup.,ViT,IN-21k} & WG Acc. & 91.5(0.8) & 92.0(1.1)& \textbf{92.6(0.8)} &91.9(0.9) & \textbf{92.6(0.5)}\\
        \cline{2-7}
        & Avg. Acc.& 93.6(1.0) &95.6(0.6)& 94.1(0.7) &96.0(0.5)  & 93.2(0.5)\\
         \hline
        \end{tabular}
        \caption{Worse-group (Row 3) and average (Row 4) test accuracy of different versions of GroupMix on Waterbirds, compared with GroupDRO. GroupMix-v3 achieves the same worse-group accuracy as GroupDRO, demonstrating the potential of GroupMix. Also note that the average accuracy of GroupMix is generally higher than that of GroupDRO.}
        \label{tab:variant_groupmix}
        \vspace{-0.4cm}
\end{table}
\vspace{-0.4cm}
\section{Experiment Results}
\label{sec:exp_result}
\vspace{-0.4cm}
We report the experimental results in this section. Data augmentation (DA), MixUp and GroupMix are included as the \emph{generalized} ERM algorithm since they are often considered as general data augmentation methods in DL and simple to implement. Thus, we denote the DA, mixup and GroupMix as \emph{general DA} in the following content. In our experiment, 
%i.e., last three rows of Tab.~\ref{tab:original_wilds_bench} and Tab.~\ref{tab:wilds_waterbirds}-\ref{tab:domainnet}, 
DA is used with ERM (mixup and GroupMix), GroupDRO and CORAL to make a fair comparison. We first compare ERM with four baselines on WILDS with default models, then the result of seven pre-trained models on the five datasets is reported. Here we only report worse-case or OOD accuracy, and the complete result is in the supplemental.
\vspace{-0.2cm}
\subsection{Default Models and Data Augmentation}
\vspace{-0.2cm}
Tab.~\ref{tab:original_wilds_bench} shows the performance of ERM with DA, mixup and GroupMix, compared on several WILDS baselines \abc{using the default models}. Except for Waterbirds, ERM with general DA achieves the best result. On Waterbirds, GroupMix substantially improves ERM with other types of data augmentation. For the result of ERM on DomainNet, \cite{gulrajani2021in} draws a similar conclusion that ERM with data augmentation is a quite strong baseline on all popular domain generalization datasets. 
\vspace{-0.2cm}
\subsection{Different Pre-Trained Models and Data Augmentation}
\vspace{-0.2cm}
On four WILDS datasets and DomainNet, we report the result of fine-tuning seven pre-trained models with three different algorithms in Tabs.~\ref{tab:wilds_waterbirds}-\ref{tab:domainnet}. The bold numbers indicate the algorithm with the best performance when fixing the pre-trained model, while the green number denotes the best pre-trained model for each algorithm. Next we analyze the empirical result for the five datasets respectively.

\textbf{Simulated Spurious Correlation: Waterbirds.} GroupDRO achieves the best result on six pre-trained models while GroupMix has the highest WG accuracy with R50 pre-trained on IN-21k. CORAL is generally worse than GroupMix and GroupDRO. The supervised pre-trained ViT on IN-21k achieves the best performance among all the settings and substantially improves the runner-up model. This may suggest that to overcome the spurious correlation, we need more data and fine-grained labels in pre-training to reduce ambiguity. In contrast, with the smaller pre-training data, self-supervised models are better than supervised ones, suggesting that when the pre-training data is not enough, the low-level features learned by self-supervision are more useful in data with spurious correlation data.

\textbf{Subpopulation Shift: FMoW.} ERM with general DA achieves the best performance on four pre-trained models and CORAL achieves the best on the other three models, demonstrating the effectiveness of DA, especially GroupMix. The MAE is quite competitive among these models, even outperforming models pre-trained on IN-21k with labels. Compared with MoCo, it suggests that for the dense object images in FMoW, the patch-based reconstruction pre-training is more effective than the image-based contrastive learning. 

\textbf{OOD Generalization: Camelyon and iWildCam.} On Camelyon, MAE is the best model on all algorithms, since the tissue slide contains dense features as in FMoW. Moreover, MoCo-R50 is better than Supervised R50 with IN-1k. For training methods, there is no significant difference among different algorithms. In contrast, supervised pre-training ViT's are substantially better than the self-supervised ones on iWildCam. Since iWildCam is an object recognition dataset while Camelyon and FMoW are dense image classification data, where no dominant object is shown in the image, the learned semantics during pre-training is crucial to learning invariant object features in the downstream task. In terms of algorithms, CORAL achieves the best performance in most cases and mixup is also competitive when using ViT-IN-21k.

\textbf{Domain Generalization: DomainNet.} Based on Tab.~\ref{tab:domainnet}, we have the conclusion consistent with \cite{gulrajani2021in} that ERM is a strong baseline combined with data augmentation, which outperforms others on four models. However, mixup and GroupMix do not work as well as CORAL on this dataset. Comparing different pre-trained models, supervised ViT outperforms R50 with a large margin when pre-trained on IN-1k, while R50 catches up and beats ViT when IN-21k is used. The phenomenon is also observed in iWildCam (Tab.~\ref{tab:wilds_iwildcam}). In self-supervised models, MoCo-ViT outperforms the other two and MoCo-R50 achieves the better performance than the supervised R50 as in iWildCam. 
\vspace{-0.3cm}

\subsubsection{Variants of GroupMix}
\label{sec:variant_groupmix}
\vspace{-0.2cm}
We evaluate four variants of GroupMix on Waterbirds. In addition to original V1, we consider the following 3 variants: 2) the weight $g_1^{(b)}, g_2^{(b)}$ is proportional to $g(\bx_1), g(\bx_2)$, which gives larger $\lambda$ to a sample from minor groups, and keep other weights 1; 3) $(g_1^{(b)}, g_2^{(b)})\propto (g(\bx_1), g(\bx_2))$,  $g^{(x)}_1=g^{(x)}_2=1.0,g^{(l)}_1=g(\bx_1),g^{(l)}_2=g(\bx_2)$ and normalize the loss weight so that the weight sums to 1 as in ERM; 4) $g^{(b)}_1=g^{(b)}_2=g^{(x)}_1=g^{(x)}_2=1.0$, $g^{(l)}_1=g(\bx_1),g^{(l)}_2=g(\bx_2)$ and normalize the weight as in 3). Tab.~\ref{tab:variant_groupmix} shows the result of four variants, where the GroupMix-V3 is the best one in terms of worse-case accuracy and matches the performance of GroupDRO. It indicates that the group-condition sampling can further improve the performance of GroupMix. Moreover, all versions of GroupMix have a higher average accuracy than GroupDRO, because mixup alleviates the overfitting in training. More details about GroupMix are in the supplemental.

% 1) $g_1=g_2=1.0$, $g'_1=g'_2=1.0$, $g^*_1=g(\bx_1),g^*_2=g(\bx_2)$, 92.2(0.5).\\
% 2) $g_1=g(\bx_1),g_2=g(\bx_2)$, $g'_1=g'_2=g^*_1=g^*_2=1.0$, 92.0(1.1)\\
% 3) $g_1=g(\bx_1),g_2=g(\bx_2)$, $g'_1=g'_2=1.0$, $g^*_1=g(\bx_1),g^*_2=g(\bx_2)$, normalize the weight for batch, 92.6(0.7)\\
% 4) $g_1=g_2=1.0$, $g'_1=g'_2=1.0$, $g^*_1=g(\bx_1),g^*_2=g(\bx_2)$, normalize the weight for batch, 91.9(0.9)\\

\begin{figure}
    \centering
    \vspace{-0.25cm}
    \includegraphics[width=1.0\linewidth]{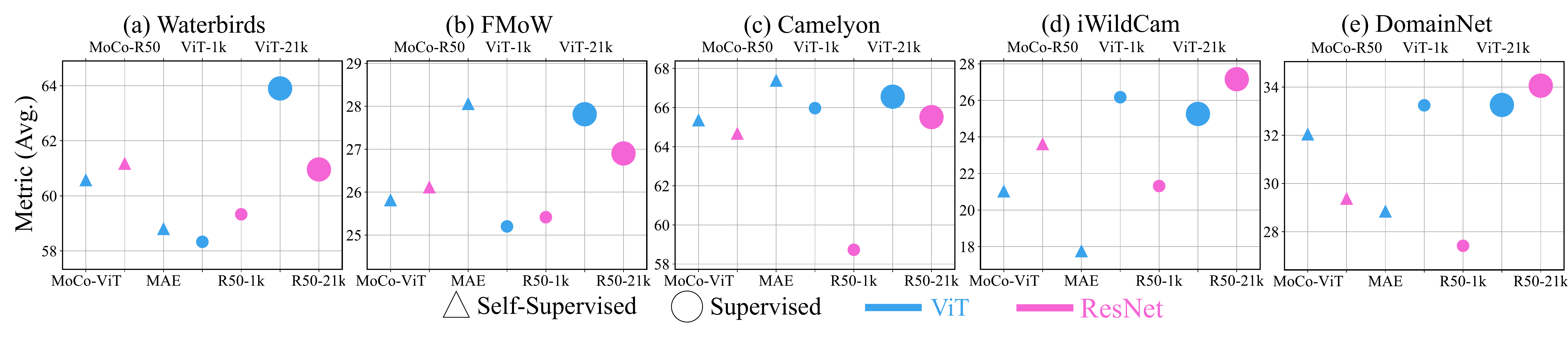}
    \vspace{-0.85cm}
    \caption{The performance of seven pre-trained models on five datasets, averaged over 5five algorithms. The triangle and circle denote self-supervised and supervised pre-trained model, while blue and pink denote ViT and R50. The large and small marker mean pre-training with IN-21k and IN-1k respectively. See our analysis in Sections~\ref{sec:exp_result} and \ref{sec:obs_tips}. }
    \label{fig:pre_train_model}
    \vspace{-0.6cm}
\end{figure}

% \begin{table}[t]
%     \centering
%     \scriptsize
%     \begin{tabular}{c|c|c|c|c|c|c|c}
%     \hline
%           Dataset  &\multicolumn{2}{c|}{Waterbirds} & \multicolumn{2}{c|}{FMoW} & Camelyon & iWildCam & DomainNet \\
%               \hline
%         Metric & WG Acc. & Avg. Acc. & WG Acc. & Avg. Acc. & OOD Acc. & OOD Macro F1 & Avg. Acc.  \\
%         %\hline
%          %Method & \multicolumn{2}{|c|}{GroupDRO} & \multicolumn{2}{|c|}{LISA} & MBDG & ERM & CADG \\
%          \hline
%          Best on BM& 91.4(1.1) \cite{sagawa2019distributionally} & 93.5(0.3) & 35.5(0.8) \cite{yao2022improving} & 52.8(1.2) & 93.3(1.0) \cite{robey2021model} & 38.5(0.6) \cite{miller2021accuracy} & \textbf{49.8} \cite{Dai2022CADGAM}\\
%          \hline
%          Our Best & \textbf{92.6(0.8)} & \textbf{94.1(0.7)} & \textbf{40.7(1.0)} & \textbf{57.4(2.1)} & \textbf{94.7(0.2)} & \textbf{43.2(1.5)} & \textbf{49.8} \\
%          \hline
%          Our ERM & \textbf{92.6(0.8)} & \textbf{94.1(0.7)} &
%          40.2(1.6) &
%          57.1(1.0) &
%          94.5(0.5) &
%          42.6(2.5) &
%          49.8 \\
%          \hline
%         \end{tabular}
%         \caption{Comparison with state-of-the-art. }
%         \label{tab:sota_compare}
% \end{table}

\vspace{-0.3cm}
\section{Key Observations and Tips}
\label{sec:obs_tips}
\vspace{-0.3cm}
\textbf{Self-supervised or supervised pre-training?} Fig.~\ref{fig:pre_train_model} shows the accuracy of seven pre-trained models, averaged over learning algorithms. For spurious correlation and subpopulation shift (Waterbirds and FMoW), self-supervised pre-training has a clear benefit over the supervised pre-training, when using the IN-1k. It indicates that the high-level features learned in supervised training are not robust to spurious correlation and subpopulation shift. On Camelyon, MAE is better than the remaining models including ViT and R50 trained on IN-21k, since the tissue images only contain low-level features. In contrast, the self-supervised model has not much benefit over supervised ones on iWildCam and DomainNet for ViT models. The OOD generalization in object recognition may need high-level features from the labels in pre-training. For ResNet, when the pre-training data is insufficient, MoCo-R50 is better than Sup. R50 on IN-1k. 

\textbf{ResNet or ViT?} On Waterbirds, FMoW and Camelyon, ViT's achieve the best performance, while on iWildCam and DomainNet, the supervised pre-trained R50 on IN-21 achieves a better result than the ViT counterpart. On Waterbirds, ViT's benefit is quite substantial when pre-trained on IN-21k, indicating that with more fine-grained labels, the spurious correlation problem can be alleviated as a result of narrowed ambiguity. On OOD generalization datasets (Camelyon, iWildCam and DomainNet), an interesting phenomenon is that ViT is more data-efficient in supervised pre-training than R50: ViT outperforms R50 with a large margin using IN-1k while the benefit is washed out if using the larger IN-21k.

\textbf{Larger pre-training dataset leads to stronger robustness?} Comparing the performance of different pretraining sizes (corresponding to marker sizes in Fig.~\ref{fig:pre_train_model}), %it is clear that 
the larger dataset IN-21k substantially improves the performance of IN-1k in most cases. Thus, it is validated that generalization under distribution shift in a downstream can be achieved by pre-training on a standard large-scale dataset. 

\begin{wraptable}{r}{7.8cm}
    \centering
    \scriptsize
    \vspace{-0.3cm}
    \begin{tabular}{c|c|c|c|c|c}
    \hline
         \multirow{2}{*}{Dataset} &  \multicolumn{3}{c|}{ERM} & \multirow{2}{*}{GroupDRO} & \multirow{2}{*}{CORAL}\\
         \cline{2-4}
         & DA & Mixup & GroupMix & & \\
         \hline
         Waterbirds & 82.5 & 83.9 & 86.3 & \textbf{88.0} & 82.4\\
         %\hline
         FMoW & 36.6 & 37.0& \textbf{37.7}& 36.9& 37.1 \\
         %\hline
         Camelyon & 91.1& 90.6& 90.7& \textbf{91.4}& 90.3 \\
         %\hline
         iWildCam & 35.4 & 36.0 & 36.0 & 18.0 &\textbf{37.0} \\
         %\hline
         DomainNet & \textbf{45.8}& 44.7& 44.2& 37.8& 45.7 \\
         \hline
    \end{tabular}
    \caption{The performance of five learning algorithms on all datasets, averaged over pre-trained models.}
    \vspace{-0.2cm}
    \label{tab:learning_algorithm}
\end{wraptable}
\textbf{Which \abc{learning} algorithm is best?} Tab.~\ref{tab:learning_algorithm} shows the averaged accuracy of each algorithm over pre-trained models. GroupDRO is the best training method on average when evaluated on Waterbirds and Camelyon, where the data size is quite small and the difficulty of classification is not high (both are binary classification task). In contrast, on large-scale datasets such as iWildCam and DomainNet, GroupDRO is the worse one compared with others, indicating that GroupDRO is specialized for small-scale data. CORAL is quite competitive on iWildCam and DomainNet but not on small datasets such as Waterbirds. The possible reason is that CORAL relies on estimated means and variances of features, whose stability and accuracy will be affected if the data is scarce. ERM with GroupMix is a strong baseline on Waterbirds and FMoW, showing the importance of using group-information in spurious correlation with spurious correlation and subpopulation shift. On Camelyon, iWildCam and DomainNet, ERM with general DA is also quite competitive, ranking the first or second on the three datasets. 

\textbf{Tips.} Finally, from these observations, we provide tips for practitioners for improving %interested in 
robustness to distribution shift:
\begin{compactenum}
\vspace{-0.2cm}
\item For OOD generalization and spurious correlation in object recognition,  use a supervised pre-trained model with as much pre-training data as possible. When the pre-training data is limited, self-supervised pre-trained models can be better.
\item %\textbf{2)} 
If the target data is small, use algorithms respecting the group information such as GroupDRO and GroupMix. If the target data is large, CORAL and ERM with general DA are quite competitive.
\item %\textbf{3)} 
For distribution shift in dense image classification, use a patch-based self-supervised pre-trained model, e.g., MAE.
\end{compactenum}
\vspace{-0.5cm}
\section{Conclusion}
\label{sec:conclusion}
\vspace{-0.4cm}
Our paper provide the first empirical study focusing on the impact of pre-training and data augmentation on distribution shift robustness. Our evaluation is done with seven pre-trained models and five training methods on five datasets including three kinds of common distribution shifts. The empirical observation shows that ERM combined with data augmentation is a very strong baseline in distribution shift, when choosing a suitable pre-trained model. By analyzing different factors in pre-training, e.g. training modes, models and data size, we summarize that the distribution shift robustness can be substantially improved by designing better pre-training. We hope that the empirical study could motivate researchers to invent fine-tuning algorithms leveraging the power of pre-trained models in the downstream task, and design pre-training strategies targeting for distribution shift robustness. The main limitation is that our current work extensively studies the distribution shift on the image classification task, which can be improved by considering more computer vision tasks such as object detection and semantic segmentation.

{\small
\bibliographystyle{unsrt}
\bibliography{ref}

\begin{thebibliography}{10}

\bibitem{he2016deep}
Kaiming He, Xiangyu Zhang, Shaoqing Ren, and Jian Sun.
\newblock Deep residual learning for image recognition.
\newblock In {\em Proceedings of the IEEE conference on computer vision and
  pattern recognition}, pages 770--778, 2016.

\bibitem{dosovitskiy2020image}
Alexey Dosovitskiy, Lucas Beyer, Alexander Kolesnikov, Dirk Weissenborn,
  Xiaohua Zhai, Thomas Unterthiner, Mostafa Dehghani, Matthias Minderer, Georg
  Heigold, Sylvain Gelly, et~al.
\newblock An image is worth 16x16 words: Transformers for image recognition at
  scale.
\newblock {\em arXiv preprint arXiv:2010.11929}, 2020.

\bibitem{koh2021wilds}
Pang~Wei Koh, Shiori Sagawa, Henrik Marklund, Sang~Michael Xie, Marvin Zhang,
  Akshay Balsubramani, Weihua Hu, Michihiro Yasunaga, Richard~Lanas Phillips,
  Irena Gao, et~al.
\newblock Wilds: A benchmark of in-the-wild distribution shifts.
\newblock In {\em International Conference on Machine Learning}, pages
  5637--5664. PMLR, 2021.

\bibitem{gulrajani2021in}
Ishaan Gulrajani and David Lopez-Paz.
\newblock In search of lost domain generalization.
\newblock In {\em International Conference on Learning Representations}, 2021.

\bibitem{sagawa2019distributionally}
Shiori Sagawa, Pang~Wei Koh, Tatsunori~B Hashimoto, and Percy Liang.
\newblock Distributionally robust neural networks for group shifts: On the
  importance of regularization for worst-case generalization.
\newblock {\em arXiv preprint arXiv:1911.08731}, 2019.

\bibitem{sun2016deep}
Baochen Sun and Kate Saenko.
\newblock Deep coral: Correlation alignment for deep domain adaptation.
\newblock In {\em European conference on computer vision}, pages 443--450.
  Springer, 2016.

\bibitem{yao2022improving}
Huaxiu Yao, Yu~Wang, Sai Li, Linjun Zhang, Weixin Liang, James Zou, and Chelsea
  Finn.
\newblock Improving out-of-distribution robustness via selective augmentation.
\newblock {\em CoRR}, abs/2201.00299, 2022.

\bibitem{robey2021model}
Alexander Robey, George~J. Pappas, and Hamed Hassani.
\newblock Model-based domain generalization.
\newblock In M.~Ranzato, A.~Beygelzimer, Y.~Dauphin, P.S. Liang, and J.~Wortman
  Vaughan, editors, {\em Advances in Neural Information Processing Systems},
  volume~34, pages 20210--20229. Curran Associates, Inc., 2021.

\bibitem{miller2021accuracy}
John~P Miller, Rohan Taori, Aditi Raghunathan, Shiori Sagawa, Pang~Wei Koh,
  Vaishaal Shankar, Percy Liang, Yair Carmon, and Ludwig Schmidt.
\newblock Accuracy on the line: on the strong correlation between
  out-of-distribution and in-distribution generalization.
\newblock In Marina Meila and Tong Zhang, editors, {\em Proceedings of the 38th
  International Conference on Machine Learning}, volume 139 of {\em Proceedings
  of Machine Learning Research}, pages 7721--7735. PMLR, 18--24 Jul 2021.

\bibitem{Dai2022CADGAM}
Chengqiu Dai, Fan Li, Xiyao Li, and Don Xie.
\newblock Cadg: A model based on cross attention for domain generalization.
\newblock {\em ArXiv}, abs/2203.17067, 2022.

\bibitem{domainnetbench}
Papers with code: {D}omain{N}et {B}enchmark (domain generalization).
\newblock
  \url{https://paperswithcode.com/sota/domain-generalization-on-domainnet}.

\bibitem{wildsbench}
{WILDS} {B}enchmark.
\newblock \url{https://wilds.stanford.edu/leaderboard/}.

\bibitem{arjovsky2019invariant}
Martin Arjovsky, L{\'e}on Bottou, Ishaan Gulrajani, and David Lopez-Paz.
\newblock Invariant risk minimization.
\newblock {\em arXiv preprint arXiv:1907.02893}, 2019.

\bibitem{krueger2021out}
David Krueger, Ethan Caballero, Joern-Henrik Jacobsen, Amy Zhang, Jonathan
  Binas, Dinghuai Zhang, Remi Le~Priol, and Aaron Courville.
\newblock Out-of-distribution generalization via risk extrapolation (rex).
\newblock In {\em International Conference on Machine Learning}, pages
  5815--5826. PMLR, 2021.

\bibitem{ahuja2020invariant}
Kartik Ahuja, Karthikeyan Shanmugam, Kush Varshney, and Amit Dhurandhar.
\newblock Invariant risk minimization games.
\newblock In {\em International Conference on Machine Learning}, pages
  145--155. PMLR, 2020.

\bibitem{rosenfeld2020risks}
Elan Rosenfeld, Pradeep Ravikumar, and Andrej Risteski.
\newblock The risks of invariant risk minimization.
\newblock {\em arXiv preprint arXiv:2010.05761}, 2020.

\bibitem{choe2020empirical}
Yo~Joong Choe, Jiyeon Ham, and Kyubyong Park.
\newblock An empirical study of invariant risk minimization.
\newblock {\em arXiv preprint arXiv:2004.05007}, 2020.

\bibitem{ahuja2020empirical}
Kartik Ahuja, Jun Wang, Amit Dhurandhar, Karthikeyan Shanmugam, and Kush~R
  Varshney.
\newblock Empirical or invariant risk minimization? a sample complexity
  perspective.
\newblock {\em arXiv preprint arXiv:2010.16412}, 2020.

\bibitem{duchi2021learning}
John~C Duchi and Hongseok Namkoong.
\newblock Learning models with uniform performance via distributionally robust
  optimization.
\newblock {\em The Annals of Statistics}, 49(3):1378--1406, 2021.

\bibitem{namkoong2016stochastic}
Hongseok Namkoong and John~C Duchi.
\newblock Stochastic gradient methods for distributionally robust optimization
  with f-divergences.
\newblock {\em Advances in neural information processing systems}, 29, 2016.

\bibitem{liu2021heterogeneous}
Jiashuo Liu, Zheyuan Hu, Peng Cui, Bo~Li, and Zheyan Shen.
\newblock Heterogeneous risk minimization.
\newblock In {\em International Conference on Machine Learning}, pages
  6804--6814. PMLR, 2021.

\bibitem{vapnik2013nature}
Vladimir Vapnik.
\newblock {\em The nature of statistical learning theory}.
\newblock Springer science \& business media, 2013.

\bibitem{ILSVRC15}
Olga Russakovsky, Jia Deng, Hao Su, Jonathan Krause, Sanjeev Satheesh, Sean Ma,
  Zhiheng Huang, Andrej Karpathy, Aditya Khosla, Michael Bernstein,
  Alexander~C. Berg, and Li~Fei-Fei.
\newblock {ImageNet Large Scale Visual Recognition Challenge}.
\newblock {\em International Journal of Computer Vision (IJCV)},
  115(3):211--252, 2015.

\bibitem{yule1926we}
G~Udny Yule.
\newblock Why do we sometimes get nonsense-correlations between time-series?--a
  study in sampling and the nature of time-series.
\newblock {\em Journal of the royal statistical society}, 89(1):1--63, 1926.

\bibitem{simon1954spurious}
Herbert~A Simon.
\newblock Spurious correlation: A causal interpretation.
\newblock {\em Journal of the American statistical Association},
  49(267):467--479, 1954.

\bibitem{zemel2013learning}
Rich Zemel, Yu~Wu, Kevin Swersky, Toni Pitassi, and Cynthia Dwork.
\newblock Learning fair representations.
\newblock In Sanjoy Dasgupta and David McAllester, editors, {\em Proceedings of
  the 30th International Conference on Machine Learning}, volume~28 of {\em
  Proceedings of Machine Learning Research}, pages 325--333, Atlanta, Georgia,
  USA, 17--19 Jun 2013. PMLR.

\bibitem{hardt2016equality}
Moritz Hardt, Eric Price, Eric Price, and Nati Srebro.
\newblock Equality of opportunity in supervised learning.
\newblock In D.~Lee, M.~Sugiyama, U.~Luxburg, I.~Guyon, and R.~Garnett,
  editors, {\em Advances in Neural Information Processing Systems}, volume~29.
  Curran Associates, Inc., 2016.

\bibitem{creager2021environment}
Elliot Creager, J{\"o}rn-Henrik Jacobsen, and Richard Zemel.
\newblock Environment inference for invariant learning.
\newblock In {\em International Conference on Machine Learning}, pages
  2189--2200. PMLR, 2021.

\bibitem{liu2021integrated}
Jiashuo Liu, Zheyuan Hu, Peng Cui, Bo~Li, and Zheyan Shen.
\newblock Kernelized heterogeneous risk minimization.
\newblock In A.~Beygelzimer, Y.~Dauphin, P.~Liang, and J.~Wortman Vaughan,
  editors, {\em Advances in Neural Information Processing Systems}, 2021.

\bibitem{peters2016causal}
J.~Peters, P.~B{\"u}hlmann, and N.~Meinshausen.
\newblock Causal inference using invariant prediction: identification and
  confidence intervals.
\newblock {\em Journal of the Royal Statistical Society, Series B (Statistical
  Methodology)}, 78(5):947--1012, 2016.
\newblock (with discussion).

\bibitem{michael2021regularizing}
Michael Oberst, Nikolaj Thams, Jonas Peters, and David~A. Sontag.
\newblock Regularizing towards causal invariance: Linear models with proxies.
\newblock In Marina Meila and Tong Zhang, editors, {\em Proceedings of the 38th
  International Conference on Machine Learning, {ICML} 2021, 18-24 July 2021,
  Virtual Event}, volume 139 of {\em Proceedings of Machine Learning Research},
  pages 8260--8270. {PMLR}, 2021.

\bibitem{mouli2022asymmetry}
S~Chandra Mouli and Bruno Ribeiro.
\newblock Asymmetry learning for counterfactually-invariant classification in
  {OOD} tasks.
\newblock In {\em International Conference on Learning Representations}, 2022.

\bibitem{oren2019distributionally}
Yonatan Oren, Shiori Sagawa, Tatsunori~B. Hashimoto, and Percy Liang.
\newblock Distributionally robust language modeling.
\newblock In {\em Proceedings of the 2019 Conference on Empirical Methods in
  Natural Language Processing and the 9th International Joint Conference on
  Natural Language Processing (EMNLP-IJCNLP)}, pages 4227--4237, Hong Kong,
  China, November 2019. Association for Computational Linguistics.

\bibitem{hendrycks2019benchmarking}
Dan Hendrycks and Thomas Dietterich.
\newblock Benchmarking neural network robustness to common corruptions and
  perturbations.
\newblock {\em arXiv preprint arXiv:1903.12261}, 2019.

\bibitem{hendrycks2021natural}
Dan Hendrycks, Kevin Zhao, Steven Basart, Jacob Steinhardt, and Dawn Song.
\newblock Natural adversarial examples.
\newblock In {\em Proceedings of the IEEE/CVF Conference on Computer Vision and
  Pattern Recognition}, pages 15262--15271, 2021.

\bibitem{ganin2016domain}
Yaroslav Ganin, Evgeniya Ustinova, Hana Ajakan, Pascal Germain, Hugo
  Larochelle, Fran\c{c}ois Laviolette, Mario Marchand, and Victor Lempitsky.
\newblock Domain-adversarial training of neural networks.
\newblock {\em J. Mach. Learn. Res.}, 17(1):2096–2030, jan 2016.

\bibitem{ganin2015unsupervised}
Yaroslav Ganin and Victor Lempitsky.
\newblock Unsupervised domain adaptation by backpropagation.
\newblock In {\em International conference on machine learning}, pages
  1180--1189. PMLR, 2015.

\bibitem{tzeng2014deep}
Eric Tzeng, Judy Hoffman, Ning Zhang, Kate Saenko, and Trevor Darrell.
\newblock Deep domain confusion: Maximizing for domain invariance.
\newblock {\em arXiv preprint arXiv:1412.3474}, 2014.

\bibitem{shen2021towards}
Zheyan Shen, Jiashuo Liu, Yue He, Xingxuan Zhang, Renzhe Xu, Han Yu, and Peng
  Cui.
\newblock Towards out-of-distribution generalization: A survey.
\newblock {\em arXiv preprint arXiv:2108.13624}, 2021.

\bibitem{wiles2022fine}
Olivia Wiles, Sven Gowal, Florian Stimberg, Sylvestre-Alvise Rebuffi, Ira
  Ktena, Krishnamurthy~Dj Dvijotham, and Ali~Taylan Cemgil.
\newblock A fine-grained analysis on distribution shift.
\newblock In {\em International Conference on Learning Representations}, 2022.

\bibitem{peng2019moment}
Xingchao Peng, Qinxun Bai, Xide Xia, Zijun Huang, Kate Saenko, and Bo~Wang.
\newblock Moment matching for multi-source domain adaptation.
\newblock In {\em Proceedings of the IEEE International Conference on Computer
  Vision}, pages 1406--1415, 2019.

\bibitem{santurkar2021breeds}
Shibani Santurkar, Dimitris Tsipras, and Aleksander Madry.
\newblock Breeds: Benchmarks for subpopulation shift.
\newblock In {\em International Conference on Learning Representations}, 2021.

\bibitem{wah2011caltech}
Catherine Wah, Steve Branson, Peter Welinder, Pietro Perona, and Serge
  Belongie.
\newblock The caltech-ucsd birds-200-2011 dataset.
\newblock 2011.

\bibitem{zhou2017places}
Bolei Zhou, Agata Lapedriza, Aditya Khosla, Aude Oliva, and Antonio Torralba.
\newblock Places: A 10 million image database for scene recognition.
\newblock {\em IEEE transactions on pattern analysis and machine intelligence},
  40(6):1452--1464, 2017.

\bibitem{liu2021just}
Evan~Z Liu, Behzad Haghgoo, Annie~S Chen, Aditi Raghunathan, Pang~Wei Koh,
  Shiori Sagawa, Percy Liang, and Chelsea Finn.
\newblock Just train twice: Improving group robustness without training group
  information.
\newblock In {\em International Conference on Machine Learning}, pages
  6781--6792. PMLR, 2021.

\bibitem{zhou2021examining}
Chunting Zhou, Xuezhe Ma, Paul Michel, and Graham Neubig.
\newblock Examining and combating spurious features under distribution shift.
\newblock In {\em International Conference on Machine Learning}, pages
  12857--12867. PMLR, 2021.

\bibitem{christie2018functional}
Gordon Christie, Neil Fendley, James Wilson, and Ryan Mukherjee.
\newblock Functional map of the world.
\newblock In {\em Proceedings of the IEEE Conference on Computer Vision and
  Pattern Recognition}, 2018.

\bibitem{shi2021gradient}
Yuge Shi, Jeffrey Seely, Philip~HS Torr, N~Siddharth, Awni Hannun, Nicolas
  Usunier, and Gabriel Synnaeve.
\newblock Gradient matching for domain generalization.
\newblock {\em arXiv preprint arXiv:2104.09937}, 2021.

\bibitem{kumar2022fine}
Ananya Kumar, Aditi Raghunathan, Robbie Jones, Tengyu Ma, and Percy Liang.
\newblock Fine-tuning can distort pretrained features and underperform
  out-of-distribution.
\newblock {\em arXiv preprint arXiv:2202.10054}, 2022.

\bibitem{bandi2018detection}
Peter Bandi, Oscar Geessink, Quirine Manson, Marcory Van~Dijk, Maschenka
  Balkenhol, Meyke Hermsen, Babak~Ehteshami Bejnordi, Byungjae Lee, Kyunghyun
  Paeng, Aoxiao Zhong, et~al.
\newblock From detection of individual metastases to classification of lymph
  node status at the patient level: the camelyon17 challenge.
\newblock {\em IEEE Transactions on Medical Imaging}, 2018.

\bibitem{beery2020iwildcam}
Sara Beery, Elijah Cole, and Arvi Gjoka.
\newblock The iwildcam 2020 competition dataset.
\newblock {\em arXiv preprint arXiv:2004.10340}, 2020.

\bibitem{li2017deeper}
Da~Li, Yongxin Yang, Yi-Zhe Song, and Timothy~M Hospedales.
\newblock Deeper, broader and artier domain generalization.
\newblock In {\em Proceedings of the IEEE international conference on computer
  vision}, pages 5542--5550, 2017.

\bibitem{fang2013unbiased}
Chen Fang, Ye~Xu, and Daniel~N Rockmore.
\newblock Unbiased metric learning: On the utilization of multiple datasets and
  web images for softening bias.
\newblock In {\em Proceedings of the IEEE International Conference on Computer
  Vision}, pages 1657--1664, 2013.

\bibitem{venkateswara2017deep}
Hemanth Venkateswara, Jose Eusebio, Shayok Chakraborty, and Sethuraman
  Panchanathan.
\newblock Deep hashing network for unsupervised domain adaptation.
\newblock In {\em Proceedings of the IEEE conference on computer vision and
  pattern recognition}, pages 5018--5027, 2017.

\bibitem{beery2018recognition}
Sara Beery, Grant Van~Horn, and Pietro Perona.
\newblock Recognition in terra incognita.
\newblock In {\em Proceedings of the European conference on computer vision
  (ECCV)}, pages 456--473, 2018.

\bibitem{huang2017densely}
Gao Huang, Zhuang Liu, Laurens Van Der~Maaten, and Kilian~Q Weinberger.
\newblock Densely connected convolutional networks.
\newblock In {\em Proceedings of the IEEE conference on computer vision and
  pattern recognition}, pages 4700--4708, 2017.

\bibitem{he2020momentum}
Kaiming He, Haoqi Fan, Yuxin Wu, Saining Xie, and Ross Girshick.
\newblock Momentum contrast for unsupervised visual representation learning.
\newblock In {\em Proceedings of the IEEE/CVF conference on computer vision and
  pattern recognition}, pages 9729--9738, 2020.

\bibitem{chen2020improved}
Xinlei Chen, Haoqi Fan, Ross Girshick, and Kaiming He.
\newblock Improved baselines with momentum contrastive learning.
\newblock {\em arXiv preprint arXiv:2003.04297}, 2020.

\bibitem{he2021masked}
Kaiming He, Xinlei Chen, Saining Xie, Yanghao Li, Piotr Doll{\'a}r, and Ross
  Girshick.
\newblock Masked autoencoders are scalable vision learners.
\newblock {\em arXiv preprint arXiv:2111.06377}, 2021.

\bibitem{steiner2021train}
Andreas Steiner, Alexander Kolesnikov, Xiaohua Zhai, Ross Wightman, Jakob
  Uszkoreit, and Lucas Beyer.
\newblock How to train your vit? data, augmentation, and regularization in
  vision transformers.
\newblock {\em arXiv preprint arXiv:2106.10270}, 2021.

\bibitem{ridnik2021imagenet}
Tal Ridnik, Emanuel Ben-Baruch, Asaf Noy, and Lihi Zelnik-Manor.
\newblock Imagenet-21k pretraining for the masses.
\newblock {\em arXiv preprint arXiv:2104.10972}, 2021.

\bibitem{cubuk2020randaug}
Ekin~Dogus Cubuk, Barret Zoph, Jon Shlens, and Quoc Le.
\newblock Randaugment: Practical automated data augmentation with a reduced
  search space.
\newblock In H.~Larochelle, M.~Ranzato, R.~Hadsell, M.F. Balcan, and H.~Lin,
  editors, {\em Advances in Neural Information Processing Systems}, volume~33,
  pages 18613--18624. Curran Associates, Inc., 2020.

\bibitem{zhang2017mixup}
Hongyi Zhang, Moustapha Cisse, Yann~N Dauphin, and David Lopez-Paz.
\newblock mixup: Beyond empirical risk minimization.
\newblock {\em arXiv preprint arXiv:1710.09412}, 2017.

\bibitem{loshchilov2018decoupled}
Ilya Loshchilov and Frank Hutter.
\newblock Decoupled weight decay regularization.
\newblock In {\em International Conference on Learning Representations}, 2018.

\bibitem{paszke2019pytorch}
Adam Paszke, Sam Gross, Francisco Massa, Adam Lerer, James Bradbury, Gregory
  Chanan, Trevor Killeen, Zeming Lin, Natalia Gimelshein, Luca Antiga, et~al.
\newblock Pytorch: An imperative style, high-performance deep learning library.
\newblock {\em Advances in neural information processing systems}, 32, 2019.

\bibitem{chen2021empirical}
Xinlei Chen, Saining Xie, and Kaiming He.
\newblock An empirical study of training self-supervised vision transformers.
\newblock In {\em Proceedings of the IEEE/CVF International Conference on
  Computer Vision}, pages 9640--9649, 2021.

\end{thebibliography}
}

\newpage
\appendix

\renewcommand{\thetable}{S\arabic{table}} 
\renewcommand{\thefigure}{S\arabic{figure}} 

\section{Details of Learning algorithms}
We first briefly introduce the algorithm of GroupDRO \cite{sagawa2019distributionally}. GroupDRO takes $\{(\bx_i,y_i,g_i)\}$ as the input, where the $g_i$ is the group label for $\bx_i$, and updates the network by minimizing a group-weighted loss. Specifically, the group weight of $j$th group is updated as
\begin{align}
    q_j(t+1) = q_j(t)\exp(\eta_q\frac{1}{N_{j}(t+1)}\sum_{\bx_i\in g_j}l(\bx_i,y_i))=q_j(t)\exp(\eta_ql_j(t+1)),
    \label{eqn:groupdro}
\end{align}
where the $\eta_q$ is the step size for group weight update, $N_j(t+1)$ is the number of samples from $j$th group in the current batch. For GroupDRO with adjustment, there is an extra term $C_{DRO}/\sqrt{n_j}$ added in $\eta_ql_j(t+1)$ of Equation \ref{eqn:groupdro}. We use the shorthand notation $l_j(t+1)$ for the $j$th group's loss in the batch. Then the loss is weighted in a group-wise way,
\begin{align}
    \loss(t+1)=q_j(t+1)\sum_{j=1}^{G}l_j(t+1).
\end{align}
In GroupMix, we add the group information in the mixup step,
\footnotesize
\begin{align}
    \lambda &\sim Beta(g_1^{(b)}\alpha,g_2^{(b)}\beta),\\ 
    \bx'&= g_1^{(x)}\lambda  \bx_1 + g_2^{(x)}(1-\lambda)  \bx_2,\\
    y' &= g^{(l)}_1\lambda  y_1 + g^{(l)}_2(1-\lambda)  y_2.
%    , l = l(\bx',y'),
\end{align}
\normalsize
The default GroupMix in our paper is to set $g_1^{(b)}=g_2^{(b)}=g_1^{(x)}=g_2^{(x)}=1$ and $g_{1,2}^{(l)}$ is related to group size, i.e., the number of training samples in a group. Assume the number of training samples of $j$th group is $n_j$, the group weight is defined as the output of a softmax function $g_j=\text{softmax}(C/\sqrt{n_j})$ where the $C$ is a hyperparameter. So the summation of group weights is one and smaller groups have larger weights. As in the main paper, we use the function $g(\cdot)$ for $\bx$ to denote the group weight of $\bx$.

We also investigate three variants of GroupMix, i.e.,

\textbf{2)} $g_1^{(b)}=\gamma g(\bx_1)$ and $g_2^{(b)}=\gamma g(\bx_2)$, keep other four weights to be 1. The $\gamma>0$ is a constant hyperparameter to control the parameters in Beta distribution. To keep a stable sampling for two input images from the same group, we fix the $\alpha,\beta$ to be 0.1 if two images are from the same group. This variant uses the property of Beta distribution to give a larger $\lambda$ for an image from minor groups.

\textbf{3)} $g_1^{(b)}=\gamma g(\bx_1)$ and $g_2^{(b)}=\gamma g(\bx_2)$, $g_1^{(x)}=g_2^{(x)}=1.0$ and $g^{(l)}_1=g(\bx_1),g^{(l)}_2=g(\bx_2)$. Then normalize each sample's weight so that the summation of weight for a batch is 1. The difference between this method and 2) is that the label is multiplied by group weight and the coefficient/weight for losses is normalized. 

\textbf{4)} $g_1^{(x)}=g_2^{(x)}=g_1^{(b)}=g_2^{(b)}=1$, $g_1^{(l)}=g(\bx_1),g_2^{(l)}=g(\bx_2)$ and normalize the loss weights as in the third variant. This is the normalized version of the default GroupMix. 

CORAL \cite{sun2016deep} aligns the feature means and covariances of different groups by minimizing the squared L2/Frobenius norm of the difference in means/covariances, i.e.,
\begin{align}
    \lambda(\|\bmu_j-\bmu_k\|_2^2 + \|\bSigma_j-\bSigma_k\|_F^2),
\end{align}
where the $j$ and $k$ denote two groups and the mean and covariance are computed from the features of two groups, i.e., the last backbone layer's output.

\begin{table}[]
    \centering
    %\small
    \begin{tabular}{c|c|c|c|c}
    \hline
              &  \multicolumn{2}{c|}{Waterbirds} & \multicolumn{2}{c}{FMoW} \\
              \hline
        Metric & WG Acc. & Avg. Acc. & WG Acc. & Avg. Acc.  \\
        \hline
         ERM & 63.7(1.9) & \textbf{97.0(0.2)} & 	34.8 (1.9) & 	55.6 (0.2)  \\
         ERM+data aug  & 80.7(1.1) & 93.6(1.4) & 34.8(1.5) & 	55.4 (0.5)  \\
         GroupDRO & 91.4(1.1) & 93.5(0.3) & 30.8(0.2)  & 	52.1 (0.5)    \\
         IRM & 67.4(5.2) & 73.4(9.7) & 30.0(1.4) & 50.8 (0.1) \\
         %Fish & - & - & 34.6(0.18) & 51.8 (0.32) & 74.7(7.1) & 22.0(1.8)  \\
         CORAL & 79.4(1.9)&94.1(0.9) & 31.7(1.2) & 	50.5 (0.4)  \\
         %Mixup & 82.61(2.46)&93.74(1.3) & 36.75(0.93)&55.00(0.73) \\
         %Group-Mix & 85.03(2.34)&89.62(2.55) & 38.96(1.22)&54.84(0.92)  \\
         %Group-Mix-Fixed-Lambda & 89.10 & - & 38.96(1.22) & - & 92.67(0.2) & 32.36(0.8) & NA\\
         %Group-Mix-Random-Lambda & 89.88 & - & 38.96(1.22) & - & 92.67(0.2) & 32.36(0.8) & NA\\
         %LDAM-Mix &90.8(0.2) & - & 36.12(2.20) & - & 92.43(0.3) & 30.69(0.7)  \\
         %Vanilla Mixup (DeiT-S) & - & - & 36.06 & - & 94.61 & 32.40  \\
         %Group-Mix (DeiT-S) & 91.27 & - & 39.80 & - & 95.34 & 31.34\\
         %LDAM-Mix (DeiT-S) & 90.95 & - & 36.56 & - & 95.88 & 31.33 \\
         \hline
         ERM (MoCo, ViT-B, IN-1k) & 84.7(1.3)&90.1(1.94) & 36.2(1.4)&55.8(0.5) \\
         ERM (MoCo, R50, IN-1k) & 82.5(0.8)&92.7(0.9) & 36.4(2.3)&53.8(1.0)\\
         ERM (MAE, ViT-B, IN-1k) & 81.7(2.0)&92.8(2.1) & 39.1(0.5)&58.1(0.4) \\
         ERM (Sup., ViT-B, IN-1k) & 76.3(2.4)&94.7(0.6) & 34.5(0.8)&56.4(0.4) \\
         ERM (Sup., R50, IN-1k) & 80.7(1.1)&93.6(1.4) & 34.9(1.9)&52.3(0.7) \\
         ERM (Sup., ViT-B, IN-21k) & 88.5(0.6)&96.2(0.7) & 37.9(1.3)&59.2(0.8) \\
         ERM (Sup., R50, IN-21k) & 82.9(1.4)&96.0(0.5) & 37.2(2.9)&55.1(1.3) \\
         \hline
        %  ERM (MoCo, ViT-B, IN-1k, 76.7\%) & 86.29 & - & - & - & - & -\\
        %  ERM (MoCo, R50, IN-1k, 74.6\%) & 85.63 & - & - & - & - & -\\
        %  ERM (MAE, ViT-B, IN-1k, 73.5\%) & 84.79 & - & - & - & - & -\\
        %  ERM (Sup., ViT-B, IN-1k, 78.2\%) & 83.64 & - & - & - & - & -\\
        %  ERM (Sup., R50, IN-1k, 76.15\%) & 83.96 & - & - & - & - & -\\
        %  ERM (Sup., ViT-B, IN-21k, 85.5\%) & 90.82 & - & - & - & - & -\\
        %  ERM (Sup., R50, IN-21k, 82.0\%) & 85.67 & - & - & - & - & -\\
         GroupDRO (MoCo, ViT-B, IN-1k) & 86.7(0.8)&90.1(0.9) & 35.2(1.6)&55.5(1.0) \\
         GroupDRO (MoCo, R50, IN-1k) & 88.1(0.9)&90.6(0.6) & 37.2(1.1)&54.5(0.7) \\
         GroupDRO (MAE, ViT-B, IN-1k) & 87.4(1.0)&88.6(0.8) & 37.9(1.4)&57.8(0.6) \\
         GroupDRO (Sup., ViT-B, IN-1k) & 87.0(1.1)&88.8(1.4) & 35.8(1.8)&56.5(0.4) \\
         GroupDRO (Sup., R50, IN-1k) & 87.6(0.4)&88.5(0.7) & 36.1(1.6)&52.0(0.7) \\
         GroupDRO (Sup., ViT-B, IN-21k) & \textbf{92.6(0.5)}&93.2(0.5) & 39.0(1.0)&59.2(0.5) \\
         GroupDRO (Sup., R50, IN-21k) & 86.6(1.1)&93.5(0.3) & 37.1(1.9)&55.9(0.7) \\
         \hline
         CORAL (MoCo, ViT-B, IN-1k) & 83.3(1.9)&93.1(1.2) &35.8(1.7)&53.5(1.7)\\
         CORAL (MoCo, R50, IN-1k) & 83.2(1.8)&92.6(1.7) & 36.1(0.7)&53.3(0.7) \\
         CORAL (MAE, ViT-B, IN-1k) & 80.0(2.3)&93.7(0.6) & 40.4(1.0)&54.8(0.8) \\
         CORAL (Sup., ViT-B, IN-1k) & 81.4(1.7)&94.2(1.3) & 35.9(1.2)&53.1(1.5) \\
         CORAL (Sup., R50, IN-1k) & 79.4(1.9)&94.1(0.9) & 34.2(1.2)&52.0(1.3) \\
         CORAL (Sup., ViT-B, IN-21k) & 86.5(1.3)&95.7(0.6) & \textbf{40.7(1.0)}&57.4(2.1) \\
         CORAL (Sup., R50, IN-21k) & 83.2(1.2)&96.1(0.5) & 36.6(0.8)&54.6(1.4) \\
         \hline
         Mixup (MoCo, ViT-B, IN-1k) & 83.5(2.7)&92.7(1.9) & 37.0(1.7)&56.4(0.3) \\
         Mixup (MoCo, R50, IN-1k) & 86.7(0.7)&93.4(0.7) & 36.0(1.2)&56.7(0.2)\\
         Mixup (MAE, ViT-B, IN-1k) & 80.3(1.5)&94.1(1.1) & 38.8(1.1)&58.4(0.7) \\
         Mixup (Sup., ViT-B, IN-1k) &79.5(4.5)&94.5(1.8) & 34.4(0.3)&56.5(0.5)\\
         Mixup (Sup., R50, IN-1k) & 82.6(2.5)&93.7(1.3) & 35.9(1.8)&54.6(0.7) \\
         Mixup (Sup., ViT-B, IN-21k) & 88.2(1.2)&96.9(1.3) & 38.5(0.9)&60.0(0.2)\\
         Mixup (Sup., R50, IN-21k) & 86.5(1.3)&95.3(0.5) & 38.4(0.6)&57.6(0.7) \\
         \hline
         GroupMix (MoCo, ViT-B, IN-1k) & 85.8(1.4)&89.6(1.2) & 36.5(1.5)&56.1(0.6) \\
         GroupMix (MoCo, R50, IN-1k) & 87.7(0.7)&91.0(0.7) & 37.1(1.4)&55.2(1.3) \\
         GroupMix (MAE, ViT-B, IN-1k) & 82.2(2.5)&92.4(1.6) & 40.2(1.6)&57.1(1.0)\\
         GroupMix (Sup., ViT-B, IN-1k) & 84.1(1.9)&90.8(1.9) & 35.8(0.7)&54.4(0.8) \\
         GroupMix (Sup., R50, IN-1k) & 85.0(2.3)&89.6(2.6) & 36.8(1.3)&53.5(1.7) \\
         GroupMix (Sup., ViT-B, IN-21k) & 91.5(0.8) & 93.6(1.0) & 38.6(1.5)& \textbf{60.1(0.2)}\\
         %Group-Mix-2 (Sup., ViT-B, IN-21k)  & 92.0(1.1) & 95.6(0.6) & - & - & - & -\\
         %Group-Mix-3 (Sup., ViT-B, IN-21k)  & \textbf{92.6(0.8)} & 94.1(0.7) & -  & - & - & -\\
         %Group-Mix-4 (Sup., ViT-B, IN-21k)  & 91.9(0.9) & 96.0(0.5) & - & - & - & -\\
         %Group-Mix-5 (Sup., ViT-B, IN-21k)  & 91.1(1.0) & 95.9(1.3) & - & - & - & -\\
         %Group-Mix-6 (Sup., ViT-B, IN-21k)  & 91.2(1.1) & 94.4(1.2) & - & - & - & -\\
         GroupMix (Sup., R50, IN-21k) & 87.5(1.6)&93.5(0.6) & 39.0(0.8)&58.1(0.5)\\
         %Group-Mix (MoCo,ViT-B) & - & - & - & - & - & -  \\
         %Group-Mix (MoCo,ViT-B,warmup) &- & - & - & - & - & -  \\
         %Group-Mix (MAE,ViT-B) &- & - & - & - & - & -  \\
         %Group-Mix (MAE,ViT-B,warmup) &- & - & - & - & - & -  \\
         %Group-Mix (CLIP,ViT-B) &- & - & - & - & - & -  \\
         %Group-Mix (CLIP,ViT-B,warmup) &- & - & - & - & - & -  \\
         %Group-Mix (DeiT,ViT-B) &- & - & - & - & - & -  \\
         %Group-Mix (DeiT,ViT-B,warmup) &- & - & - & - & - & -  \\
         %Group-Mix (ViT-21k,ViT-B) &- & - & - & - & - & -  \\
         %Group-Mix (ViT-21k,ViT-B,warmup) &- & - & - & - & - & - \\
         \hline
    \end{tabular}
    \caption{The full result of Waterbirds and FMoW with worse-group accuracy and averaged accuracy.}
    \label{tab:waterbirds_fmow}
\end{table}

\section{Experimental Settings}
\textbf{Waterbirds.} As in WILDS benchmark \cite{koh2021wilds}, we use the input resolution of 224$\times$224, the training epoch of 200 and batch size of 128 in this paper. We search the learning rate from \{0.000001,0.000002, 0.000004, 0.000008, 0.00001, 0.00002, 0.00004, 0.00008\} and weight decay (WD) from \{0.0,1e-3,1e-2,1e-1,1.0\}, $\alpha$ of mixup from \{0.1,0.3,1.0\}. For GroupMix, we search the constant $C$ from \{1.0,3.0,10.0,30.0\}. For GroupDRO, the constant $C_{DRO}$ is searched from \{1.0,2.0,4.0,8.0,16.0\} and the step size $\eta_{q}$ is fixed as 0.01 for all datasets. For CORAL, the $\lambda$ is searched from \{0.3,1.0,3.0\}. Thus, in the hyperparameter search stage, we train 960 models. 
%For GroupMix, search the learning rate from \{0.000001,0.000002, 0.000004, 0.000008, 0.00001, 0.00002, 0.00004, 0.00008\}, WD from \{0.0,1e-3,1e-2,1e-1,1.0\}, $\alpha$ from \{0.1,0.3,1.0\}, $C$ from \{1.0,3.0,10.0,30.0\}. For GroupDRO, search the $C$ from \{1.0,2.0,4.0,8.0,16.0\}. For CORAL, the $\lambda$ in CORAL is searched from \{0.3,1.0,3.0\}. ERM: 8*5=, Mixup: 8*5*3=, GroupMix: 8*5*3*4=, GroupDRO: 8*5*5=, CORAL: 8*5*3=. \textbf{960}.

\textbf{FMoW.} The input resolution is set as 224$\times$224, the training epoch as 50 and batch size as 72. We fix the WD as 0.0 following existing practice and search the learning rate from \{0.00001, 0.0001, 0.001\} and the $\alpha$ from \{0.1,0.3,1.0\}. We do not use the adjustment GroupDRO on this dataset and the following datasets. For GroupMix, we search $C$ from \{3.0,10.0,30.0,100.0\}. The $\lambda$ in CORAL is searched from \{0.3,1.0,3.0\}. We trained 60 models for hyperparameter search on FMoW. 

%For GroupMix, search the learning rate from \{0.00001, 0.0001, 0.001\}, $\alpha$ from \{0.1,0.3,1.0\}, $C$ from \{3.0,10.0,30.0,100.0\}. Fix the WD to be 0. For CORAL, the $\lambda$ in CORAL is searched from \{0.3,1.0,3.0\}. ERM: 3=, Mixup: 3*3=, GroupMix: 3*3*4=, GroupDRO: 3=, CORAL: 3*3=. \textbf{60}
\begin{table}[t]
    \centering
    
    \small{A) Model selection: Out-of-distribution validation set}
    \tiny
    \newcolumntype{S}{>{\centering\arraybackslash} p{0.8cm}}
    \begin{tabular}{c|c|c|c|c|c|c|c|c|c|c}
    \hline
            &  \multicolumn{6}{c|}{ERM} & \multicolumn{2}{c|}{\multirow{2}{*}{GroupDRO}} & \multicolumn{2}{c}{\multirow{2}{*}{CORAL}} \\
              \cline{1-7}
        PT Model & \multicolumn{2}{c|}{Data Aug.} & \multicolumn{2}{c|}{Mixup} & \multicolumn{2}{c|}{GroupMix} & \multicolumn{2}{c|}{} & \multicolumn{2}{c}{} \\
        \hline
         MoCo-ViT-IN-1k & 34.0(0.5)&51.1(1.0) & 31.7(2.1)&49.3(1.1) & 32.8(0.7)&48.7(1.7) & 13.5(1.0)&22.6(1.2) & 35.2(0.6)& 47.5(2.1) \\
         MoCo-R50-IN-1k & 34.5(1.5)&51.7(1.2) & 36.8(1.6)&51.2(1.4) & 36.0(1.1)&50.2(2.0) & 20.8(0.8)&30.2(1.0) & 37.2(1.1)&51.1(1.6) \\
         MAE-ViT-IN-1k & 28.4(2.0)&44.3(1.7) & 27.3(2.2)&44.6(1.9) & 28.0(2.9)&44.5(2.0) & 9.2(0.9)&14.7(1.1) & 31.4(2.6)&42.1(1.2) \\
         \hline
         Sup-ViT-IN-1k & 38.9(0.7)&53.6(1.3) &  41.0(0.7)&54.7(1.2)& 40.9(1.2)&54.6(1.1) & 20.5(0.5)&30.2(1.1) & 41.9(0.9)&51.8(1.5) \\
         Sup-R50-IN-1k & 32.2(1.2)&47.0(1.4) & 32.8(1.4)&49.6(0.6) & 31.9(0.9)&48.4(1.2) & 18.9(0.8)&28.9(0.7) & 33.4(0.5)&48.1(1.2) \\
         Sup-ViT-IN-21k & 39.0(2.7)&55.4(2.4) & \textbf{41.3(2.4)}&\textbf{55.7(2.1)} & 41.1(1.4)&\textbf{57.2(0.4)} & 19.0(1.2)&28.7(1.5) & 36.4(1.6)&47.3(1.8) \\
         Sup-R50-IN-21k & \textbf{40.9(1.3)} & \textbf{55.5(0.9)} & 40.8(0.4)&55.1(1.0) & \textbf{41.2(1.3)}&55.0(1.0) & \textbf{24.0(0.8)}&\textbf{33.4(0.7)} & \textbf{43.2(0.9)} & \textbf{52.1(2.4)} \\
         \hline
         %\vspace{0.4cm}
         %\multicolumn{3}{c|}{Avg. Over Models} & 35.5& 35.6& 36.5& 18.3&\textbf{37.1}\\
         %\hline
         %\vspace{0.4cm}
        \end{tabular}
        
        \small{\vspace{0.4cm}B) Model selection: In-distribution (ID) validation set}
        \tiny
        \begin{tabular}{c|c|c|c|c|c|c|c|c|c|c}
    \hline
             & \multicolumn{6}{c|}{ERM} & \multicolumn{2}{c|}{\multirow{2}{*}{GroupDRO}} & \multicolumn{2}{c}{\multirow{2}{*}{CORAL}} \\
              \cline{1-7}
        PT Model & \multicolumn{2}{c|}{Data Aug.} & \multicolumn{2}{c|}{Mixup} & \multicolumn{2}{c|}{GroupMix} & \multicolumn{2}{c|}{} & \multicolumn{2}{c}{} \\
        \hline
         MoCo-ViT-IN-1k & 32.5(0.8)&49.8(0.6) & 30.6(1.6)&47.8(1.7) & 33.6(1.5)&48.8(1.5) & 13.4(0.5)&21.9(1.7) & 36.2(0.7)& 48.9(2.0) \\
         MoCo-R50-IN-1k & 34.9(1.6)&51.7(1.1) & 36.5(1.4)&50.6(1.3) & 36.1(1.1)&49.0(1.5) & 20.8(0.8)&30.7(0.8) & 37.9(1.2)&50.6(1.6) \\
         MAE-ViT-IN-1k & 28.6(1.6)&43.9(1.4) & 27.4(1.8)&44.7(1.9) & 28.1(3.0)&43.9(2.7) & 9.6(1.3)&14.8(1.5) & 31.8(1.9)&42.3(1.8) \\
         \hline
         Sup-ViT-IN-1k & 39.1(0.9)&54.0(1.0) & 40.3(1.5)&54.5(2.0) & 40.2(1.4)&53.2(0.8) & 20.3(0.4)&30.0(0.8) & 42.2(0.9)&51.5(1.2) \\
         Sup-R50-IN-1k & 31.1(0.9)&45.6(0.4) & 32.6(1.2)&49.9(0.9) & 31.8(0.7)&47.7(1.0) & 19.3(0.6)&29.8(1.3) & 34.4(0.9)&47.6(1.4) \\
         Sup-ViT-IN-21k & 38.4(2.9)&54.9(2.0) & \textbf{40.7(1.9)}& \textbf{55.3(2.1)} & \textbf{41.6(2.2)} & \textbf{56.1(1.2)} & 18.7(1.5)&28.8(1.5) & 36.6(0.8)&47.5(1.2) \\
         Sup-R50-IN-21k & \textbf{40.0(0.6)} & \textbf{55.1(1.1)} & 40.4(0.8)&54.8(1.6) & 40.4(2.0)&54.8(1.6) & \textbf{23.9(0.9)}& \textbf{33.5(0.8)}  & \textbf{43.2(0.8)} & \textbf{52.8(1.7)} \\
         \hline
         %\multicolumn{3}{c|}{Avg. Over Models} & 35.5& 35.6& 36.5& 18.3&\textbf{37.1}\\
         %\hline
        \end{tabular}
        \caption{WILDS-iWildCam result of OOD and ID validation set model selection. The left and right cell of each learning algorithm denote OOD and ID Macro F1 score. The best performance in each column is highlighted.}
        \label{tab:wilds_iwildcam_all_ood}
        \vspace{-0.6cm}
\end{table}
\begin{table}[t]
\scriptsize
    \centering
    \begin{tabular}{c|c|c|c|c|c|c|c}
    \hline
              &  clipart & infograph & painting & quickdraw & real & sketch & Avg \\
        \hline
        %  ERM & - & - & - & - & - & - & 40.9(0.1) \\
        %  GroupDRO & - & - & - & - & - & - & 33.3(0.2) \\
        %  IRM & - & - & - & - & - & - & 33.9(2.8) \\
        %  CORAL & - & - & - & - & - & - & 41.5(0.1) \\
        %  Mixup & - & - & - & - & - & - & 39.2(0.1) \\
        %  Group-Mix & - & - & - & - & - & - & 37.3\\
        %  \hline
         %Group-Mix-Fixed-Lambda & 89.10 & - & 38.96(1.22) & - & 92.67(0.2) & 32.36(0.8) & NA\\
         %Group-Mix-Random-Lambda & 89.88 & - & 38.96(1.22) & - & 92.67(0.2) & 32.36(0.8) & NA\\
         %LDAM-Mix &90.8(0.2) & - & 36.12(2.20) & - & 92.43(0.3) & 30.69(0.7)  \\
         %Vanilla Mixup (DeiT-S) & - & - & 36.06 & - & 94.61 & 32.40  \\
         %Group-Mix (DeiT-S) & 91.27 & - & 39.80 & - & 95.34 & 31.34\\
         %LDAM-Mix (DeiT-S) & 90.95 & - & 36.56 & - & 95.88 & 31.33 \\
         ERM (MoCo, ViT-B, IN-1k) & 66.8 $\pm$ 0.2       & 24.2 $\pm$ 0.3       & 54.5 $\pm$ 0.1       & 17.3 $\pm$ 0.3       & 65.7 $\pm$ 0.1       & 56.4 $\pm$ 0.1       & 47.5 $\pm$ 0.1\\
         ERM (MoCo, R50, IN-1k) & 62.5 $\pm$ 0.1       & 22.0 $\pm$ 0.2       & 50.5 $\pm$ 0.1       & 14.5 $\pm$ 0.2       & 62.4 $\pm$ 0.1       & 53.3 $\pm$ 0.1       & 44.2  $\pm$ 0.1 \\
         ERM (MAE, ViT-B, IN-1k) & 61.8 $\pm$ 0.1       & 21.5 $\pm$ 0.1       & 49.4 $\pm$ 0.4       & 17.4 $\pm$ 0.4       & 59.6 $\pm$ 0.4       & 51.7 $\pm$ 0.3       & 43.6 $\pm$ 0.2\\
         ERM (Supervised, ViT-B, IN-1k)  & 67.5 $\pm$ 0.2       & 23.6 $\pm$ 0.2       & 54.1 $\pm$ 0.3       & 17.6 $\pm$ 0.2       & 68.7 $\pm$ 0.1       & 54.7 $\pm$ 0.2       & 47.7 $\pm$ 0.1  \\
         ERM (Supervised, R50, IN-1k) & 58.1 $\pm$ 0.3       & 18.8 $\pm$ 0.3       & 46.7 $\pm$ 0.3       & 12.2 $\pm$ 0.4       & 59.6 $\pm$ 0.1       & 49.8 $\pm$ 0.4 & 40.9 $\pm$ 0.1\\
         ERM (Supervised, ViT-B, IN-21k)  & 68.1 $\pm$ 0.3       & 23.1 $\pm$ 0.2       & 54.1 $\pm$ 0.1       & 18.0 $\pm$ 0.5       & 65.6 $\pm$ 0.3       & 55.7 $\pm$ 0.2       & 47.4 $\pm$ 0.2 \\
         ERM (Supervised, R50, IN-21k) & 67.7 $\pm$ 0.1       & \textbf{26.5 $\pm$ 0.2}       & \textbf{57.1 $\pm$ 0.1}       & 16.2 $\pm$ 0.1       & \textbf{73.2 $\pm$ 0.1}       & 58.2 $\pm$ 0.2       & 49.8 $\pm$ 0.1  \\
         \hline
        %  ERM (MoCo, ViT-B, IN-1k, 76.7\%) & 86.29 & - & - & - & - & -\\
        %  ERM (MoCo, R50, IN-1k, 74.6\%) & 85.63 & - & - & - & - & -\\
        %  ERM (MAE, ViT-B, IN-1k, 73.5\%) & 84.79 & - & - & - & - & -\\
        %  ERM (Supervised, ViT-B, IN-1k, 78.2\%) & 83.64 & - & - & - & - & -\\
        %  ERM (Supervised, R50, IN-1k, 76.15\%) & 83.96 & - & - & - & - & -\\
        %  ERM (Supervised, ViT-B, IN-21k, 85.5\%) & 90.82 & - & - & - & - & -\\
        %  ERM (Supervised, R50, IN-21k, 82.0\%) & 85.67 & - & - & - & - & -\\
         GroupDRO (MoCo, ViT-B, IN-1k) & 50.8 $\pm$ 0.3       & 22.0 $\pm$ 0.1       & 43.4 $\pm$ 0.2       & 12.7 $\pm$ 0.1       & 53.3 $\pm$ 0.1       & 43.6 $\pm$ 0.1       & 37.6  $\pm$ 0.0\\
         GroupDRO (MoCo, R50, IN-1k) & 50.7 $\pm$ 0.4       & 18.5 $\pm$ 0.2       & 39.5 $\pm$ 0.2       & 10.0 $\pm$ 0.2       & 53.2 $\pm$ 0.3       & 43.4 $\pm$ 0.3       & 35.9 $\pm$ 0.1\\
         GroupDRO (MAE, ViT-B, IN-1k) & 44.5 $\pm$ 0.5       & 19.5 $\pm$ 0.4       & 32.8 $\pm$ 0.2       & 11.1 $\pm$ 0.4       & 45.6 $\pm$ 0.5       & 36.9 $\pm$ 0.4       & 31.7 $\pm$ 0.1\\
         GroupDRO (Sup., ViT-B, IN-1k)& 60.6 $\pm$ 0.2       & 20.9 $\pm$ 0.4       & 46.8 $\pm$ 0.4       & 13.8 $\pm$ 0.4       & 64.6 $\pm$ 0.1       & 47.6 $\pm$ 0.2       & 42.4 $\pm$ 0.0 \\
         GroupDRO (Sup., R50, IN-1k) & 47.2 $\pm$ 0.5       & 17.5 $\pm$ 0.4       & 33.8 $\pm$ 0.5       & 9.3 $\pm$ 0.3        & 51.6 $\pm$ 0.4       & 40.1 $\pm$ 0.6       & 33.3 $\pm$ 0.2 \\
         GroupDRO (Sup., ViT-B, IN-21k) & 58.8 $\pm$ 0.2       & 20.5 $\pm$ 0.7       & 44.9 $\pm$ 0.4       & 15.4 $\pm$ 0.5       & 59.6 $\pm$ 0.1       & 45.7 $\pm$ 0.5       & 40.8 $\pm$ 0.2 \\
         GroupDRO (Sup., R50, IN-21k)  & 59.1 $\pm$ 0.2       & 22.3 $\pm$ 0.1       & 49.6 $\pm$ 0.1       & 11.0 $\pm$ 0.1       & 66.4 $\pm$ 0.1       & 50.2 $\pm$ 0.2       & 43.1 $\pm$ 0.0 \\
         \hline
         CORAL (MoCo, ViT-B, IN-1k) & 66.8 $\pm$ 0.1       & 16.1 $\pm$ 6.5       & 54.9 $\pm$ 0.1       & 17.9 $\pm$ 0.3       & 65.5 $\pm$ 0.1       & 56.2 $\pm$ 0.6       & 46.2 $\pm$ 1.1\\
         CORAL (MoCo, R50, IN-1k) & 61.4 $\pm$ 0.6       & 20.6 $\pm$ 0.2       & 47.4 $\pm$ 0.2       & 13.8 $\pm$ 0.4       & 59.6 $\pm$ 0.4       & 52.5 $\pm$ 0.4       & 42.6 $\pm$ 0.3 \\
         CORAL (MAE, ViT-B, IN-1k) &61.9 $\pm$ 0.2       & 21.6 $\pm$ 0.1       & 49.4 $\pm$ 0.2       & 16.8 $\pm$ 0.4       & 59.7 $\pm$ 0.1       & 51.7 $\pm$ 0.5       & 43.5 $\pm$ 0.1 \\
         CORAL (Sup., ViT-B, IN-1k) & 67.8 $\pm$ 0.2       & 23.9 $\pm$ 0.2       & 55.1 $\pm$ 0.3       & 17.5 $\pm$ 0.1       & 68.6 $\pm$ 0.2       & 55.2 $\pm$ 0.3       & 48.0 $\pm$ 0.0 \\
         CORAL (Sup., R50, IN-1k) & 59.2 $\pm$ 0.1       & 19.7 $\pm$ 0.2       & 46.6 $\pm$ 0.3       & 13.4 $\pm$ 0.4       & 59.8 $\pm$ 0.2       & 50.1 $\pm$ 0.6       & 41.5 $\pm$ 0.1\\
         CORAL (Sup., ViT-B, IN-21k) & \textbf{69.5 $\pm$ 0.1}       & 24.8 $\pm$ 0.1       & 56.5 $\pm$ 0.1       & \textbf{19.9 $\pm$ 0.5}       & 67.2 $\pm$ 0.1       & 57.5 $\pm$ 0.4       & \textbf{52.1 $\pm$ 1.1}  \\
         CORAL (Sup., R50, IN-21k) & 67.2 $\pm$ 0.4       & 25.2 $\pm$ 0.2       & 55.7 $\pm$ 0.1       & 14.9 $\pm$ 0.3       & 71.5 $\pm$ 0.2       & \textbf{58.6 $\pm$ 0.2}       & 48.8 $\pm$ 0.1  \\
         \hline
         Mixup (MoCo, ViT-B, IN-1k) & 65.0 $\pm$ 0.3       & 23.4 $\pm$ 0.1       & 54.3 $\pm$ 0.2       & 18.1 $\pm$ 0.5       & 63.7 $\pm$ 0.2       & 55.4 $\pm$ 0.3       & 46.7 $\pm$ 0.1\\
         Mixup (MoCo, R50, IN-1k) & 58.7 $\pm$ 0.5       & 19.0 $\pm$ 0.2       & 49.0 $\pm$ 0.4       & 13.2 $\pm$ 0.2       & 58.8 $\pm$ 0.2       & 51.2 $\pm$ 0.4       & 41.6 $\pm$ 0.1  \\
         Mixup (MAE, ViT-B, IN-1k) & 59.7 $\pm$ 0.1       & 20.1 $\pm$ 0.7       & 47.6 $\pm$ 0.4       & 17.0 $\pm$ 0.1       & 55.0 $\pm$ 1.6       & 50.2 $\pm$ 0.2       & 41.6 $\pm$ 0.5 \\
         Mixup (Sup., ViT-B, IN-1k) & 66.9 $\pm$ 0.2       & 23.4 $\pm$ 0.3       & 54.0 $\pm$ 0.4       & 17.1 $\pm$ 0.2       & 67.8 $\pm$ 0.2       & 54.4 $\pm$ 0.3       & 47.3 $\pm$ 0.1 \\
         Mixup (Sup., R50, IN-1k) & 55.7 $\pm$ 0.3       & 18.5 $\pm$ 0.5       & 44.3 $\pm$ 0.5       & 12.5 $\pm$ 0.4       & 55.8 $\pm$ 0.3       & 48.2 $\pm$ 0.5       & 39.2 $\pm$ 0.1\\
         Mixup (Sup., ViT-B, IN-21k)  & 68.4 $\pm$ 0.3       & 23.8 $\pm$ 0.6       & 54.4 $\pm$ 0.7       & 19.8 $\pm$ 0.4       & 65.1 $\pm$ 0.2       & 55.1 $\pm$ 0.6       & 47.8 $\pm$ 0.2      \\
         Mixup (Sup., R50, IN-21k) & 65.7 $\pm$ 0.2       & 24.3 $\pm$ 0.2       & 57.0 $\pm$ 0.5       & 14.9 $\pm$ 0.1       & 70.9 $\pm$ 0.2       & 58.3 $\pm$ 0.2       & 48.5 $\pm$ 0.1   \\
         \hline
         GroupMix (MoCo, ViT-B, IN-1k)& 64.0 $\pm$ 0.5       & 23.6 $\pm$ 0.1       & 54.3 $\pm$ 0.4       & 17.5 $\pm$ 0.3       & 62.9 $\pm$ 0.6       & 55.1 $\pm$ 0.3       & 46.3 $\pm$ 0.2 \\
         GroupMix (MoCo, R50, IN-1k) & 58.8 $\pm$ 0.2       & 19.5 $\pm$ 0.3       & 47.6 $\pm$ 0.8       & 12.9 $\pm$ 0.1       & 57.7 $\pm$ 0.5       & 51.3 $\pm$ 0.4       & 41.3 $\pm$ 0.2\\
         GroupMix (MAE, ViT-B, IN-1k) & 58.6 $\pm$ 0.9       & 20.7 $\pm$ 0.4       & 45.8 $\pm$ 2.5       & 17.2 $\pm$ 0.3       & 56.7 $\pm$ 0.1       & 50.2 $\pm$ 0.2       & 41.5 $\pm$ 0.6\\
         GroupMix (Sup., ViT-B, IN-1k) & 66.8 $\pm$ 0.3       & 23.4 $\pm$ 0.5       & 54.4 $\pm$ 0.9       & 17.3 $\pm$ 0.1       & 67.5 $\pm$ 0.1       & 54.3 $\pm$ 0.4       & 47.3 $\pm$ 0.2   \\
         GroupMix (Sup., R50, IN-1k) & 52.9 $\pm$ 0.2 & 17.2 $\pm$ 0.3 & 43.3 $\pm$ 0.8 & 11.8 $\pm$ 0.4 & 51.5 $\pm$ 0.3 & 47.3 $\pm$ 0.2 & 37.3 $\pm$ 0.2     \\
        GroupMix (Sup., ViT-B, IN-21k)  & 67.8 $\pm$ 0.6       & 23.7 $\pm$ 0.5       & 54.5 $\pm$ 0.5       & 19.4 $\pm$ 0.5       & 65.4 $\pm$ 0.2       & 55.0 $\pm$ 0.6       & 47.6 $\pm$ 0.3\\
         GroupMix (Sup., R50, IN-21k) & 65.6 $\pm$ 0.1       & 24.3 $\pm$ 0.4       & 55.9 $\pm$ 0.7       & 14.3 $\pm$ 0.1       & 70.8 $\pm$ 0.3       & 58.2 $\pm$ 0.2       & 48.2 $\pm$ 0.2 \\
         %Group-Mix (MoCo,ViT-B) & - & - & - & - & - & -  \\
         %Group-Mix (MoCo,ViT-B,warmup) &- & - & - & - & - & -  \\
         %Group-Mix (MAE,ViT-B) &- & - & - & - & - & -  \\
         %Group-Mix (MAE,ViT-B,warmup) &- & - & - & - & - & -  \\
         %Group-Mix (CLIP,ViT-B) &- & - & - & - & - & -  \\
         %Group-Mix (CLIP,ViT-B,warmup) &- & - & - & - & - & -  \\
         %Group-Mix (DeiT,ViT-B) &- & - & - & - & - & -  \\
         %Group-Mix (DeiT,ViT-B,warmup) &- & - & - & - & - & -  \\
         %Group-Mix (ViT-21k,ViT-B) &- & - & - & - & - & -  \\
         %Group-Mix (ViT-21k,ViT-B,warmup) &- & - & - & - & - & - \\
         \hline
    \end{tabular}
    \caption{The full experimental result on DomainNet.}
    \label{tab:domainnet_full}
    \vspace{-0.8cm}
\end{table}

\begin{figure*}[t]
  \vspace{-0.3cm}
  \centering
  \begin{center}
    \newcolumntype{C}{>{\centering\arraybackslash} m{2.76cm}}
  \begin{tabular}{m{0.3cm}@{}C@{}C@{}C@{}C@{}C}
     & (a) DA & (b) Mixup & (c) GroupMix & (d) GroupDRO & (e) CORAL \\    
    \rotatebox{90}{(1) Waterbirds} & \includegraphics[width=0.2\textwidth]{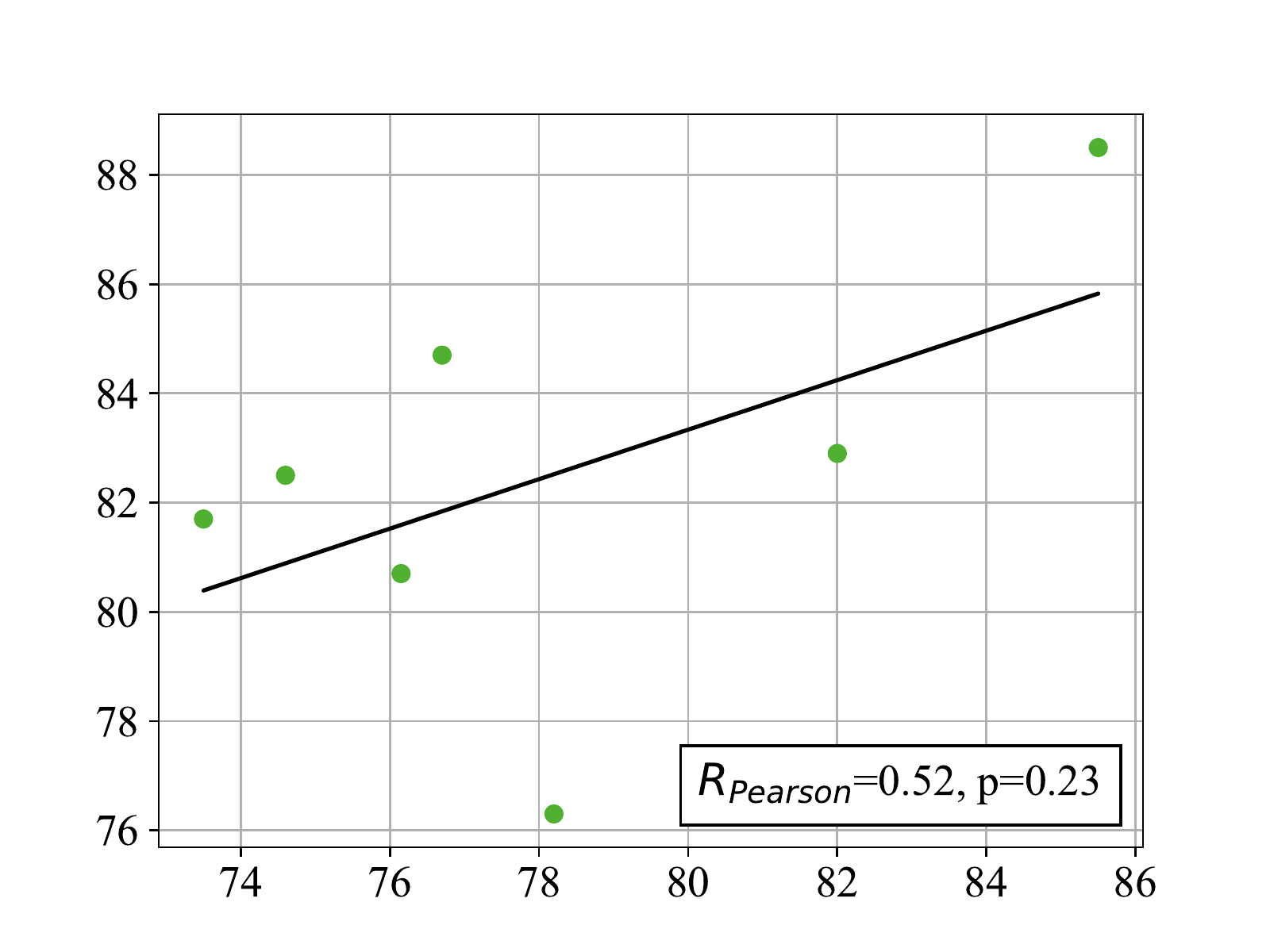}&
    \includegraphics[width=0.2\textwidth]{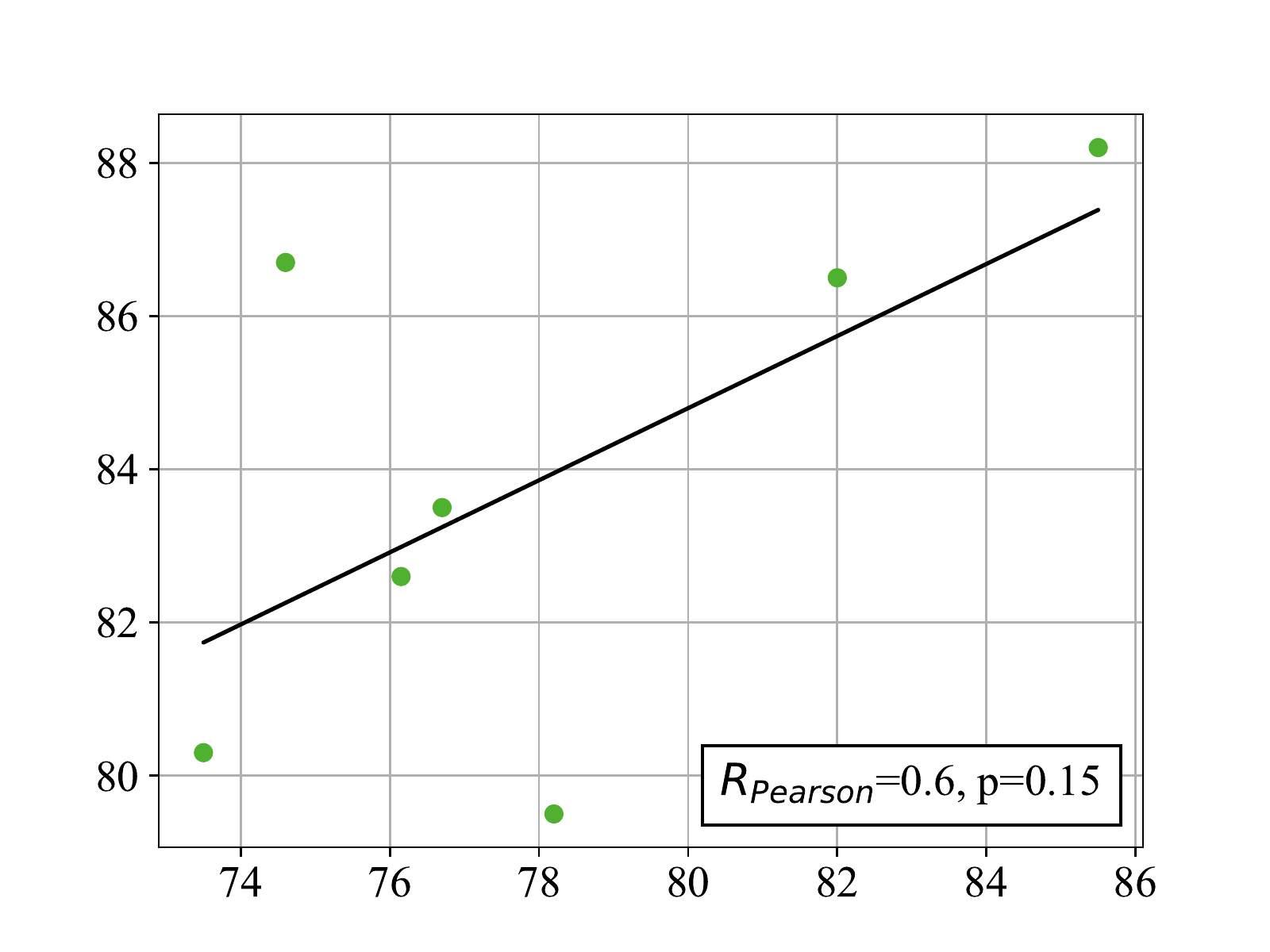}
    &
    \includegraphics[width=0.2\textwidth]{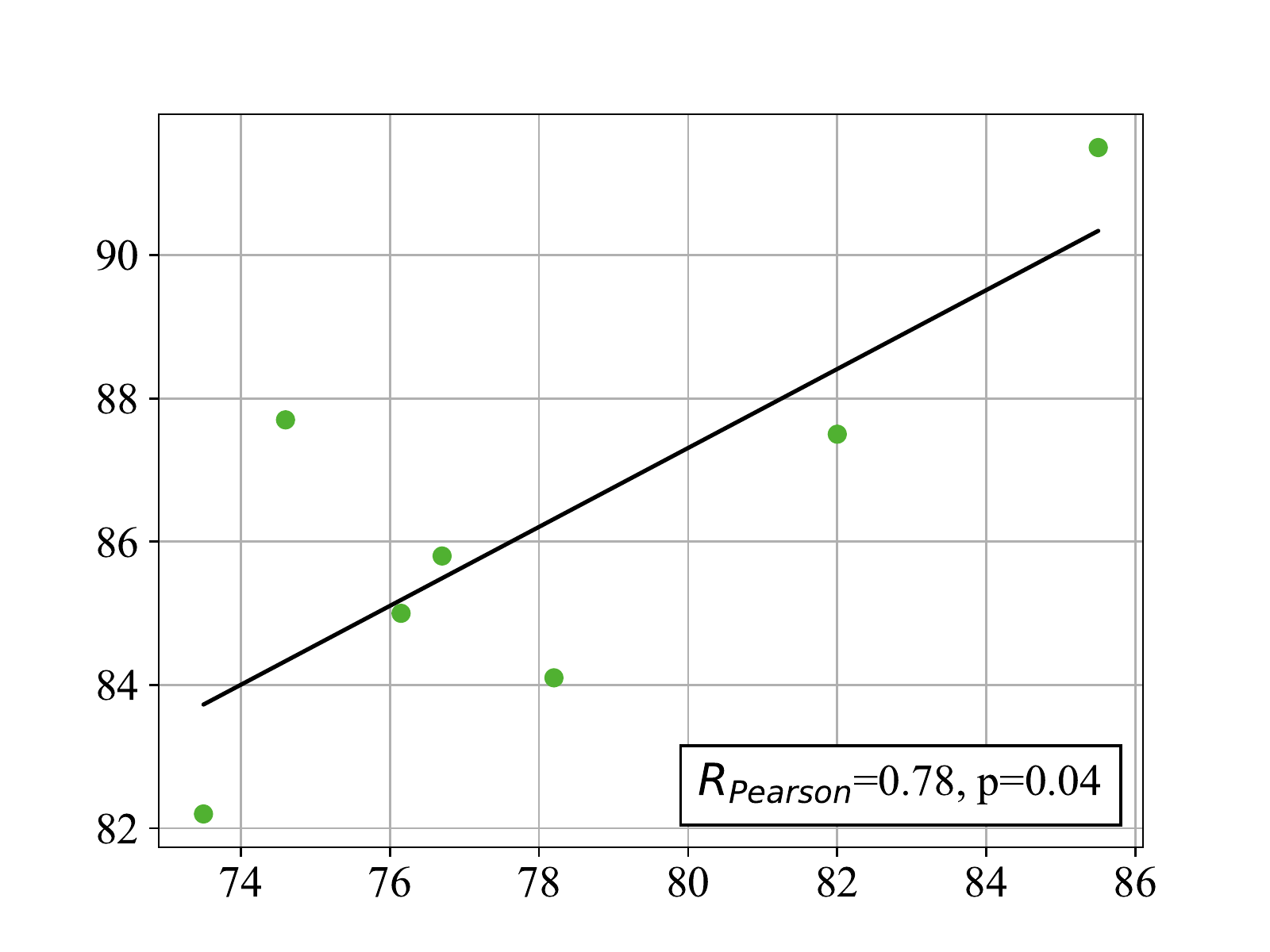}
    &
    \includegraphics[width=0.2\textwidth]{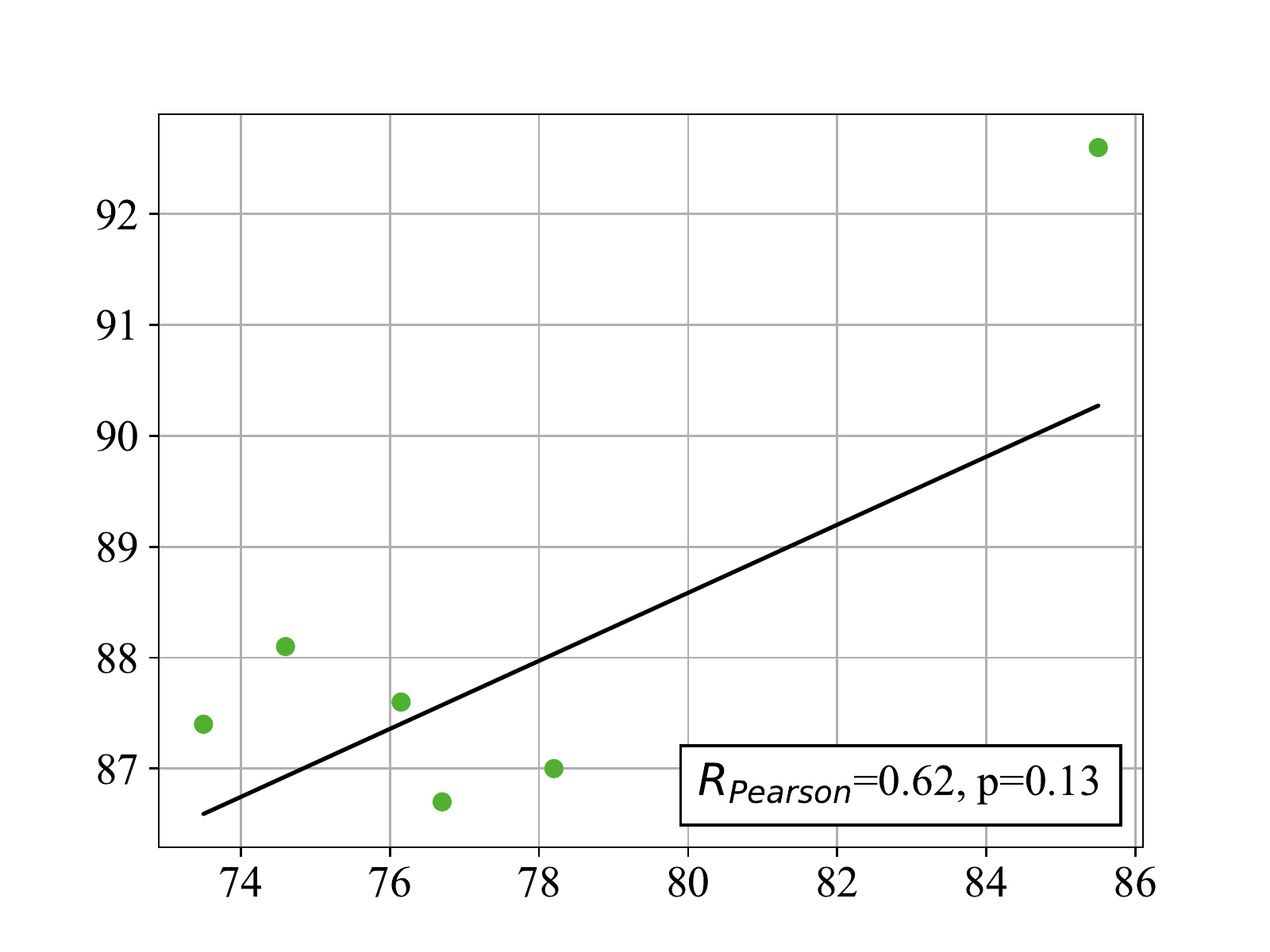}
    &
    \includegraphics[width=0.2\textwidth]{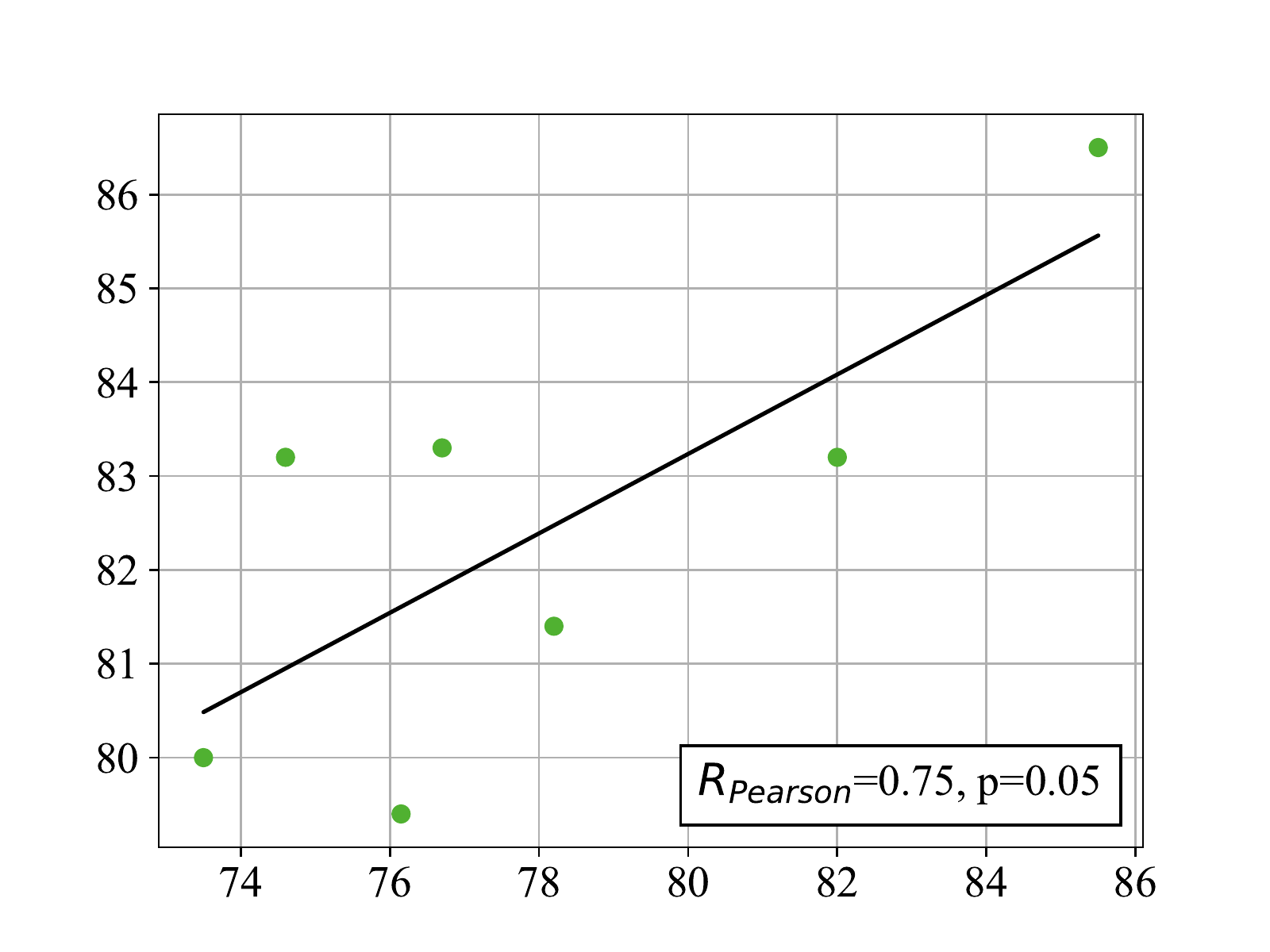}
    \\
    \rotatebox{90}{(2) FMoW} & \includegraphics[width=0.2\textwidth]{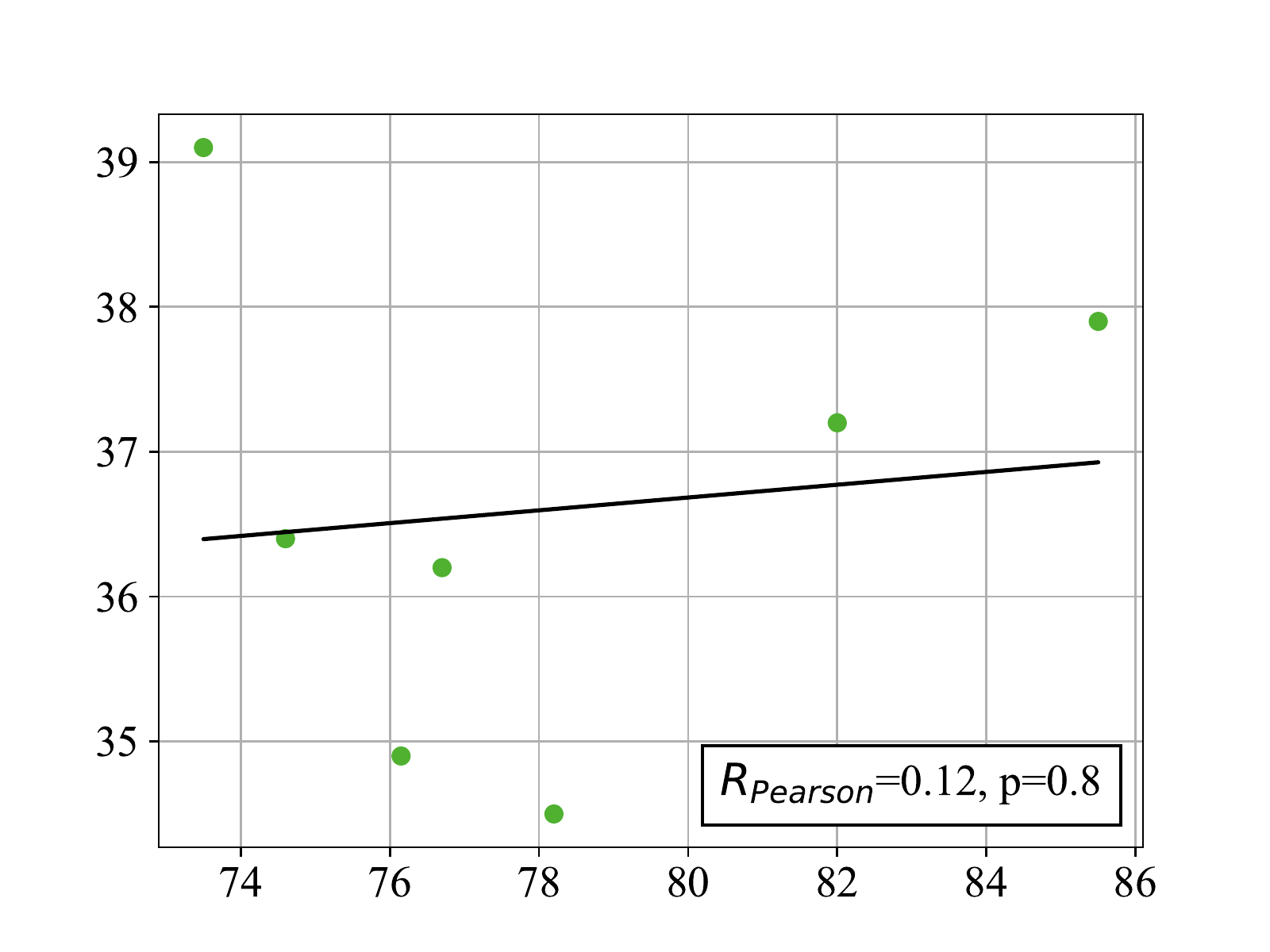}&
    \includegraphics[width=0.2\textwidth]{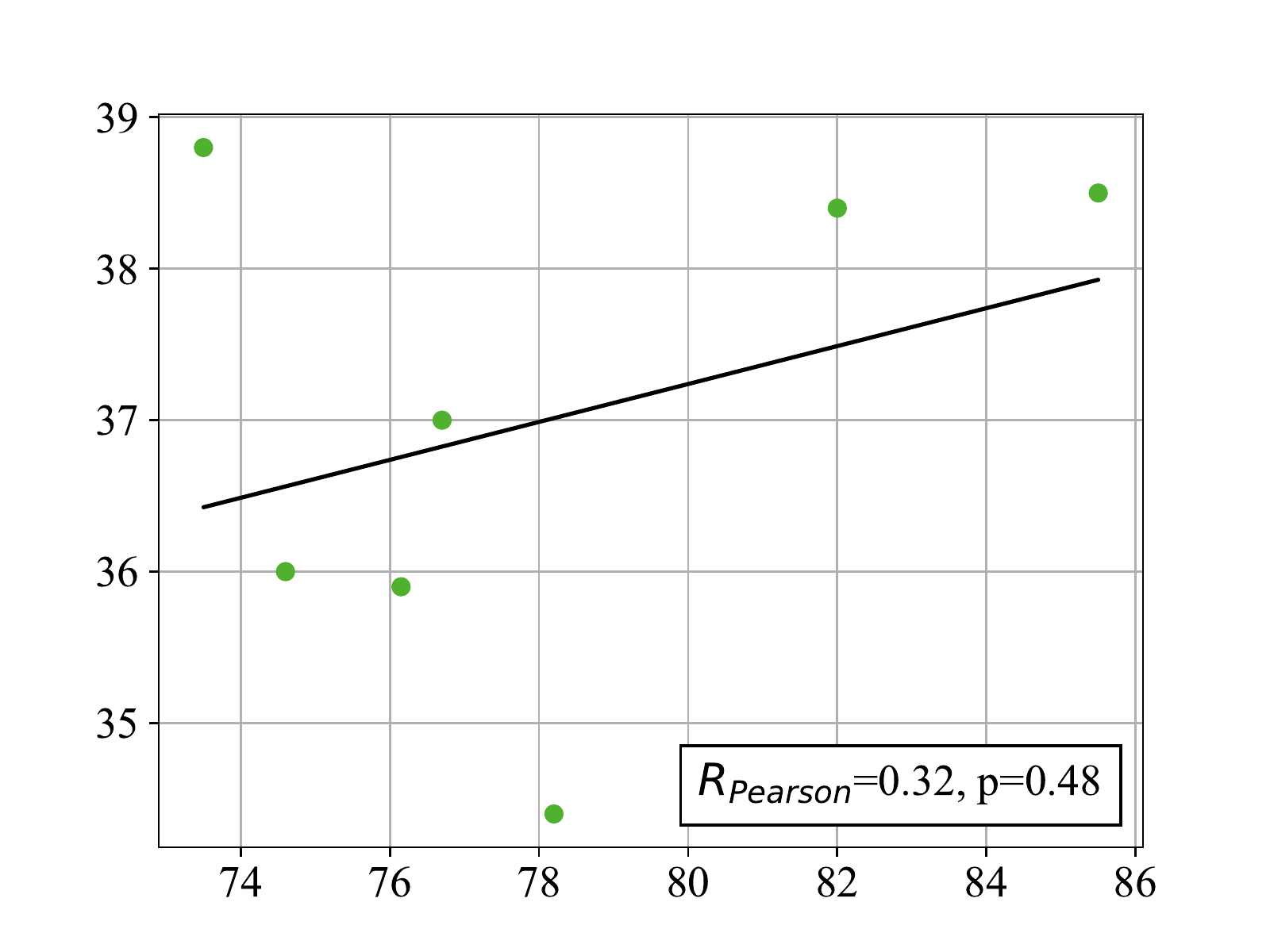}
    &
    \includegraphics[width=0.2\textwidth]{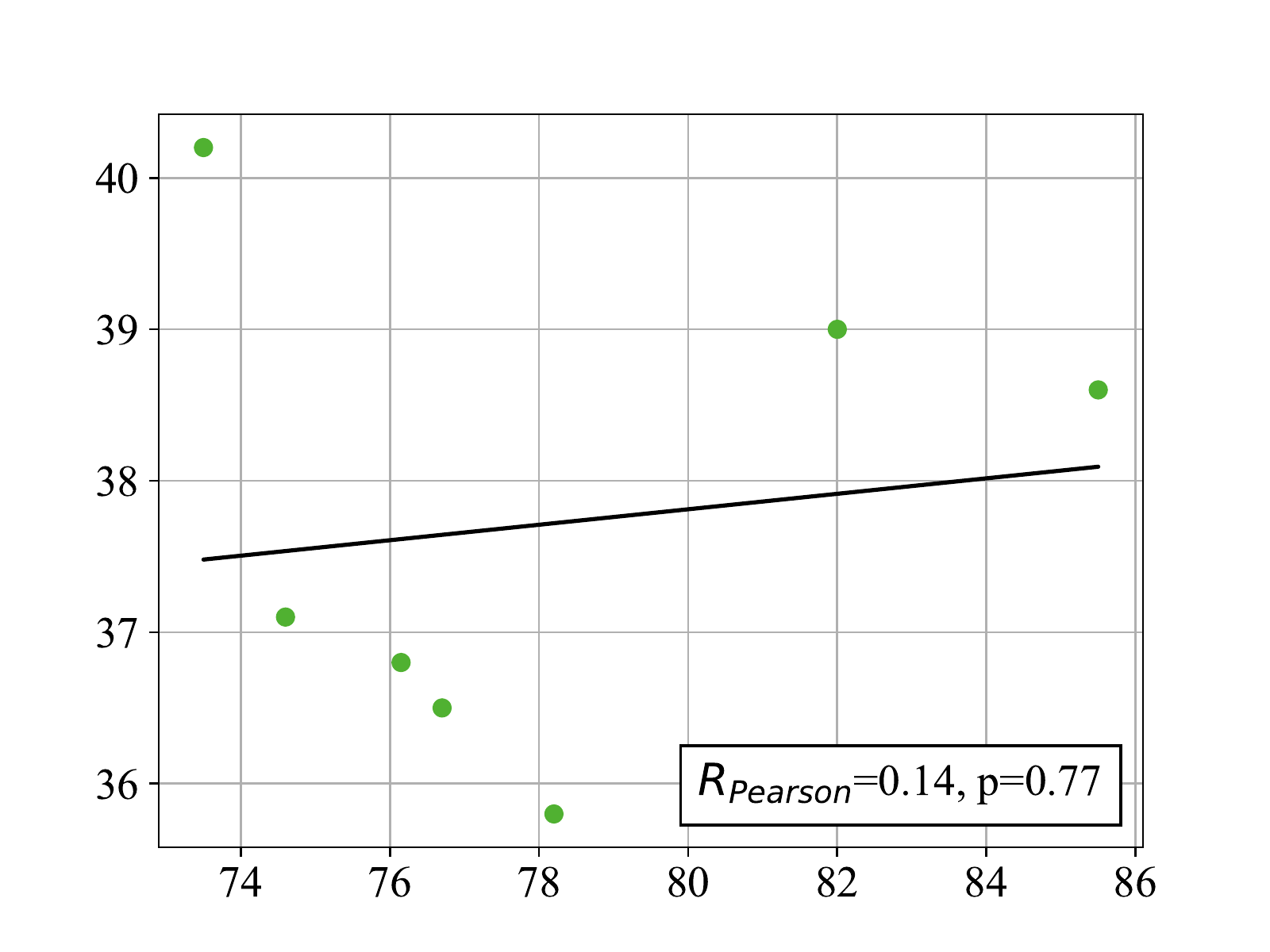}
    &
    \includegraphics[width=0.2\textwidth]{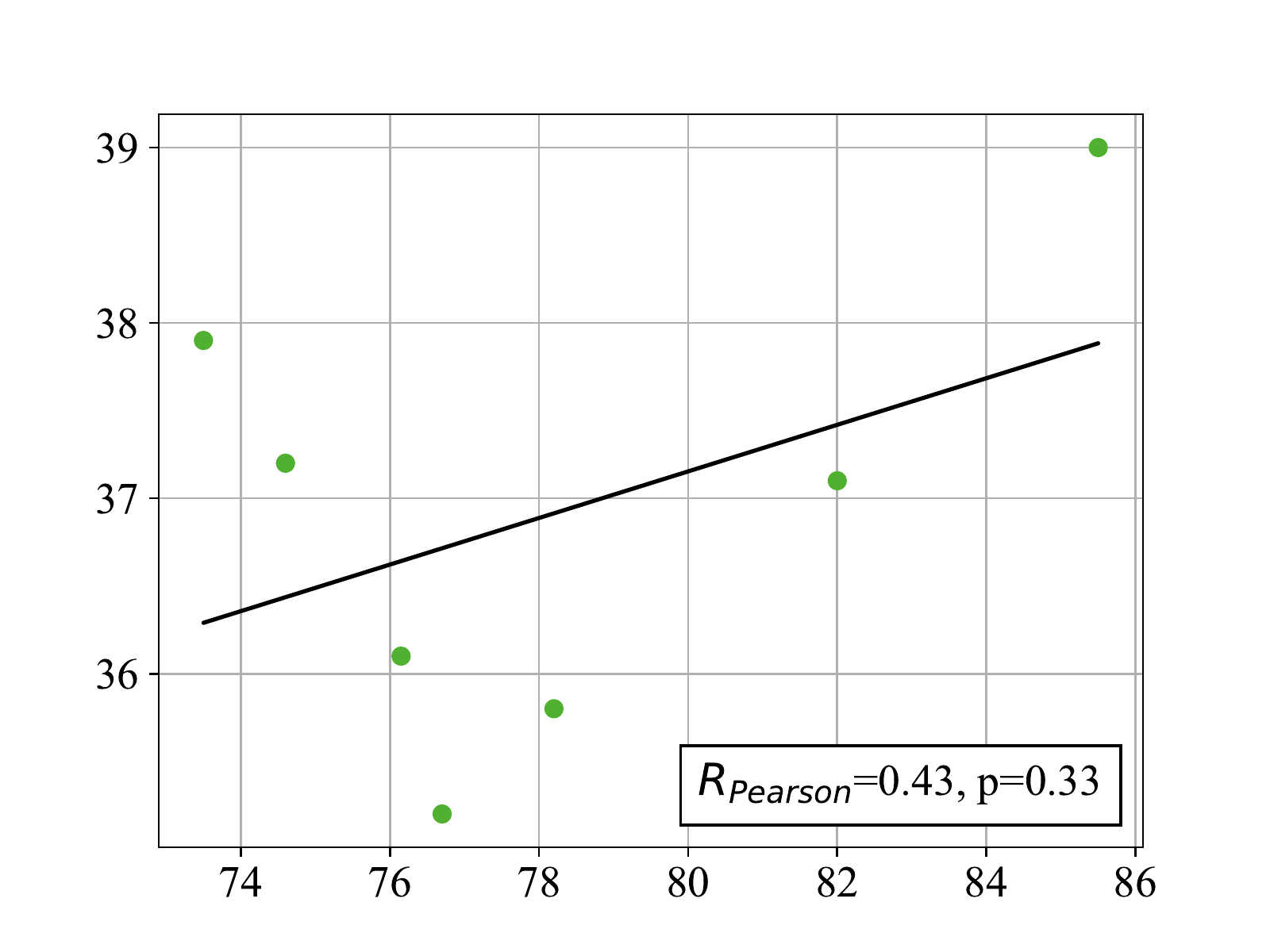}
    &
    \includegraphics[width=0.2\textwidth]{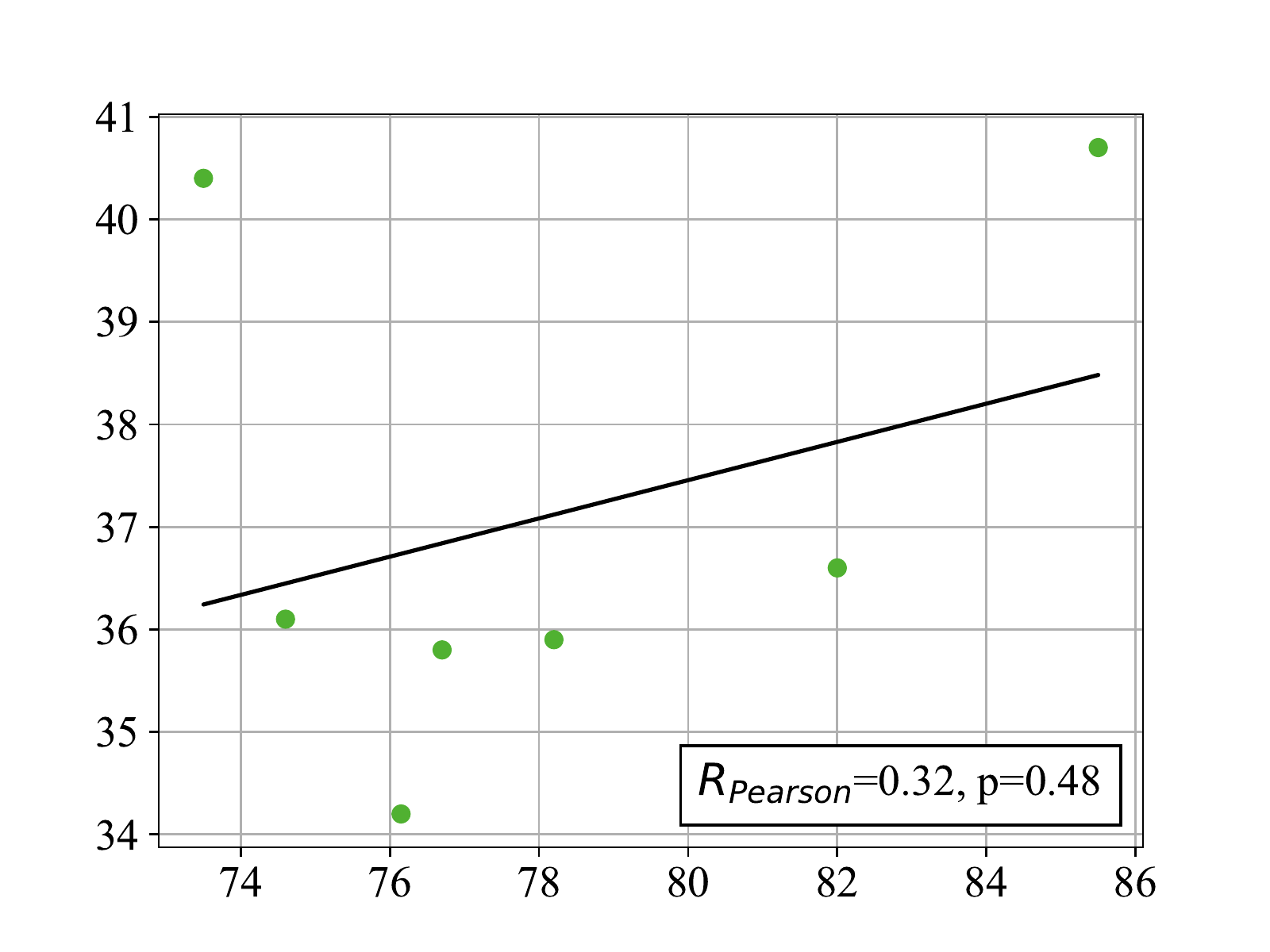}
    \\
    \rotatebox{90}{(3) Camelyon} & \includegraphics[width=0.2\textwidth]{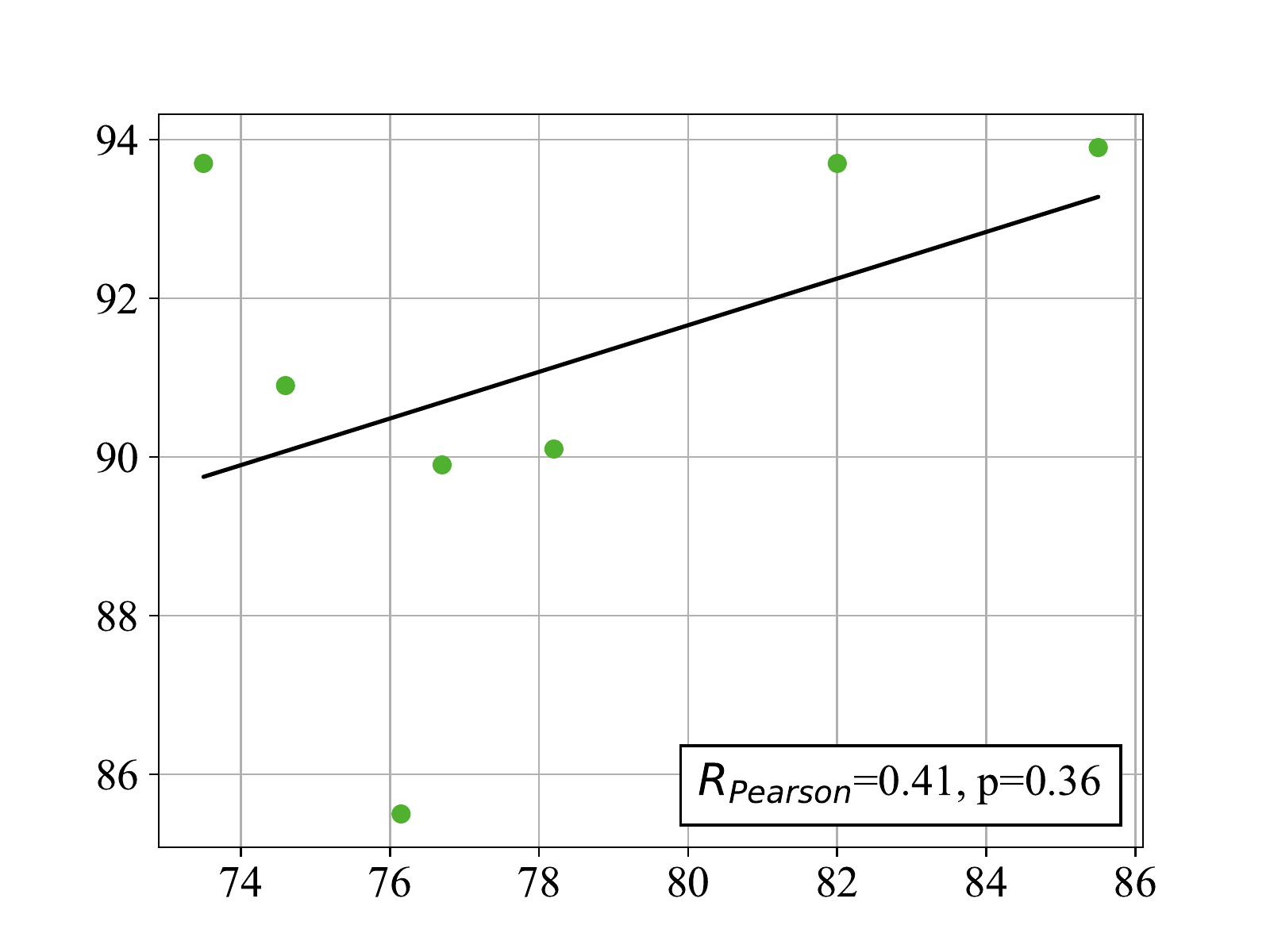}&
    \includegraphics[width=0.2\textwidth]{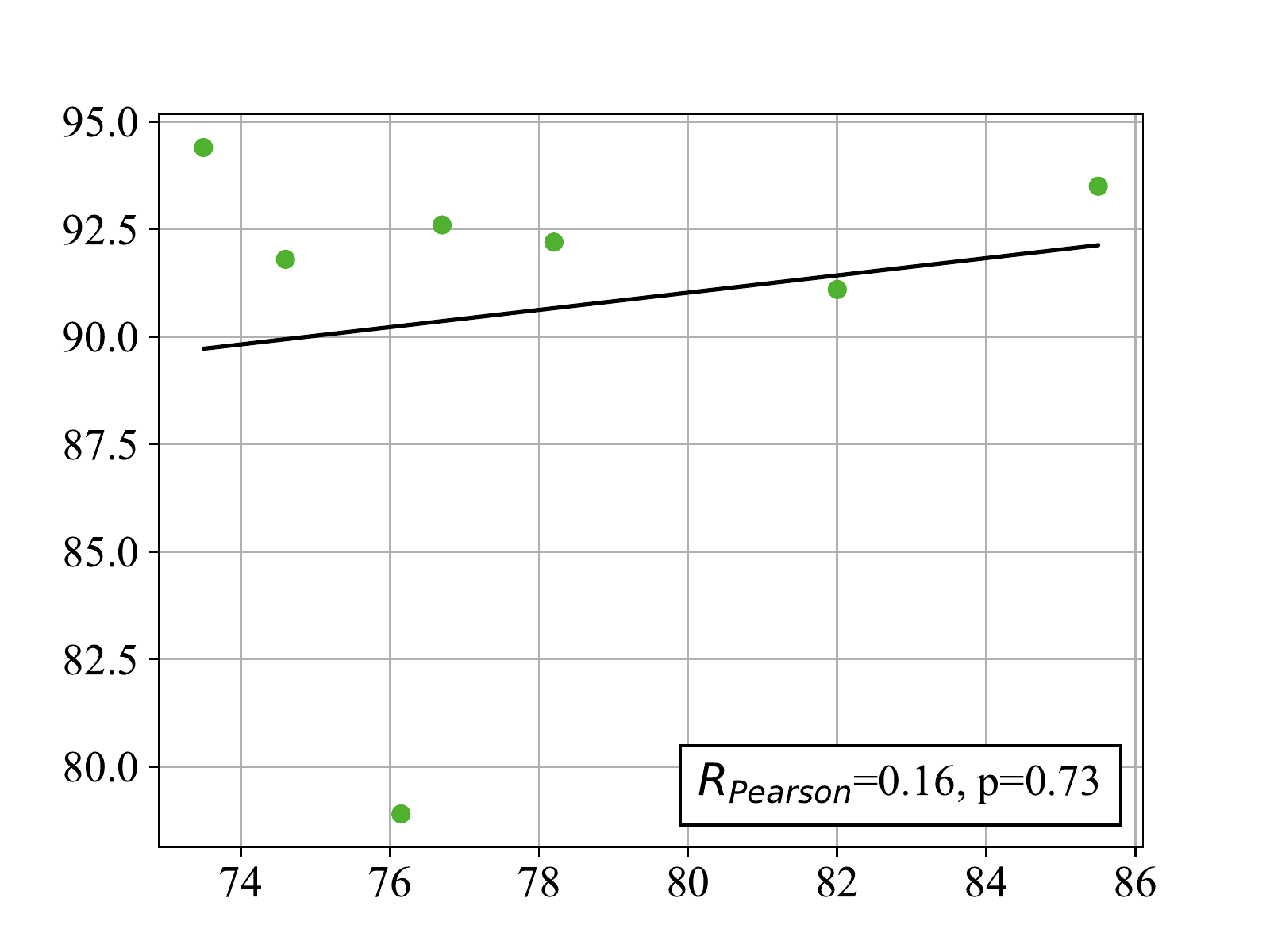}
    &
    \includegraphics[width=0.2\textwidth]{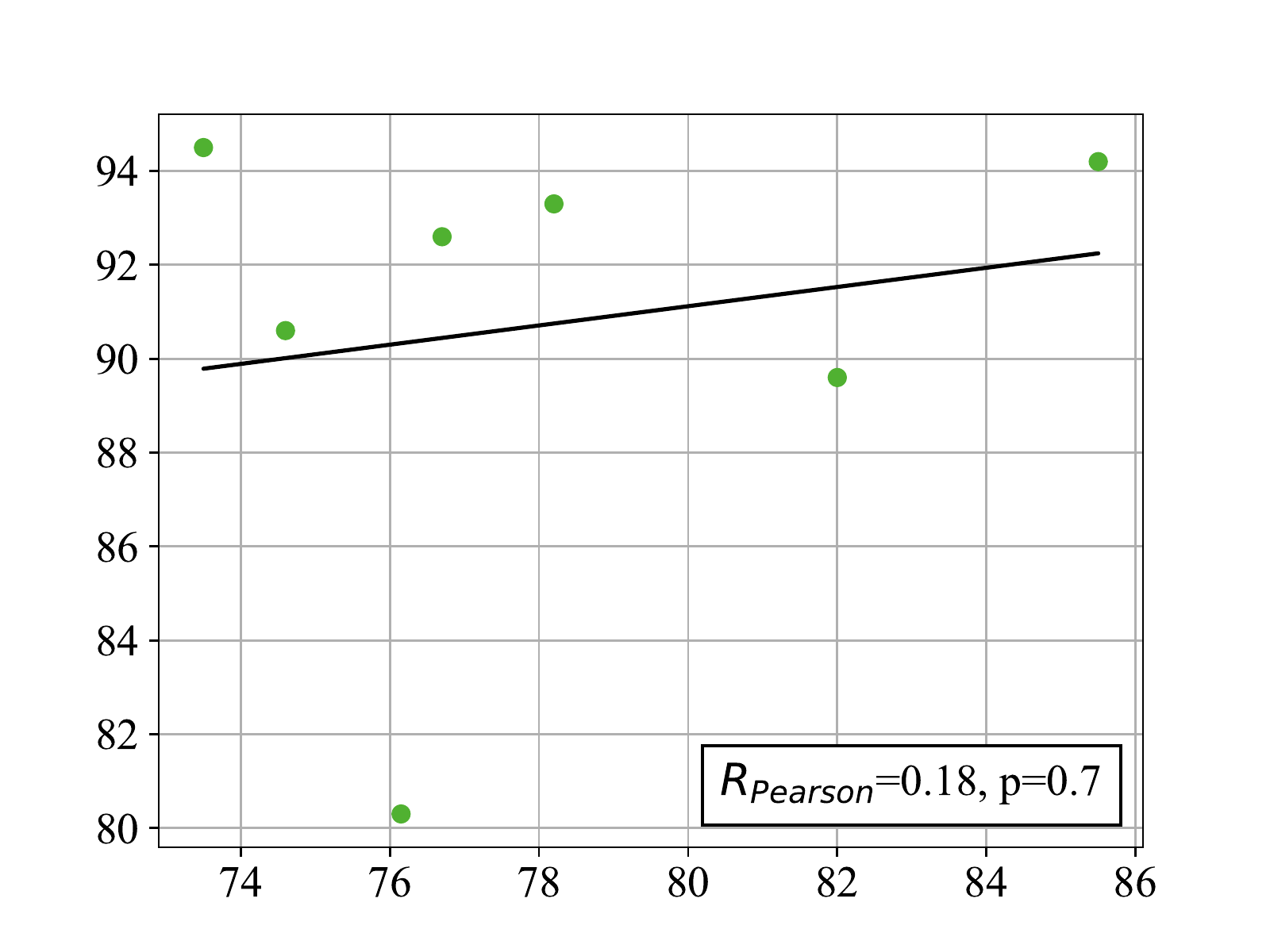}
    &
    \includegraphics[width=0.2\textwidth]{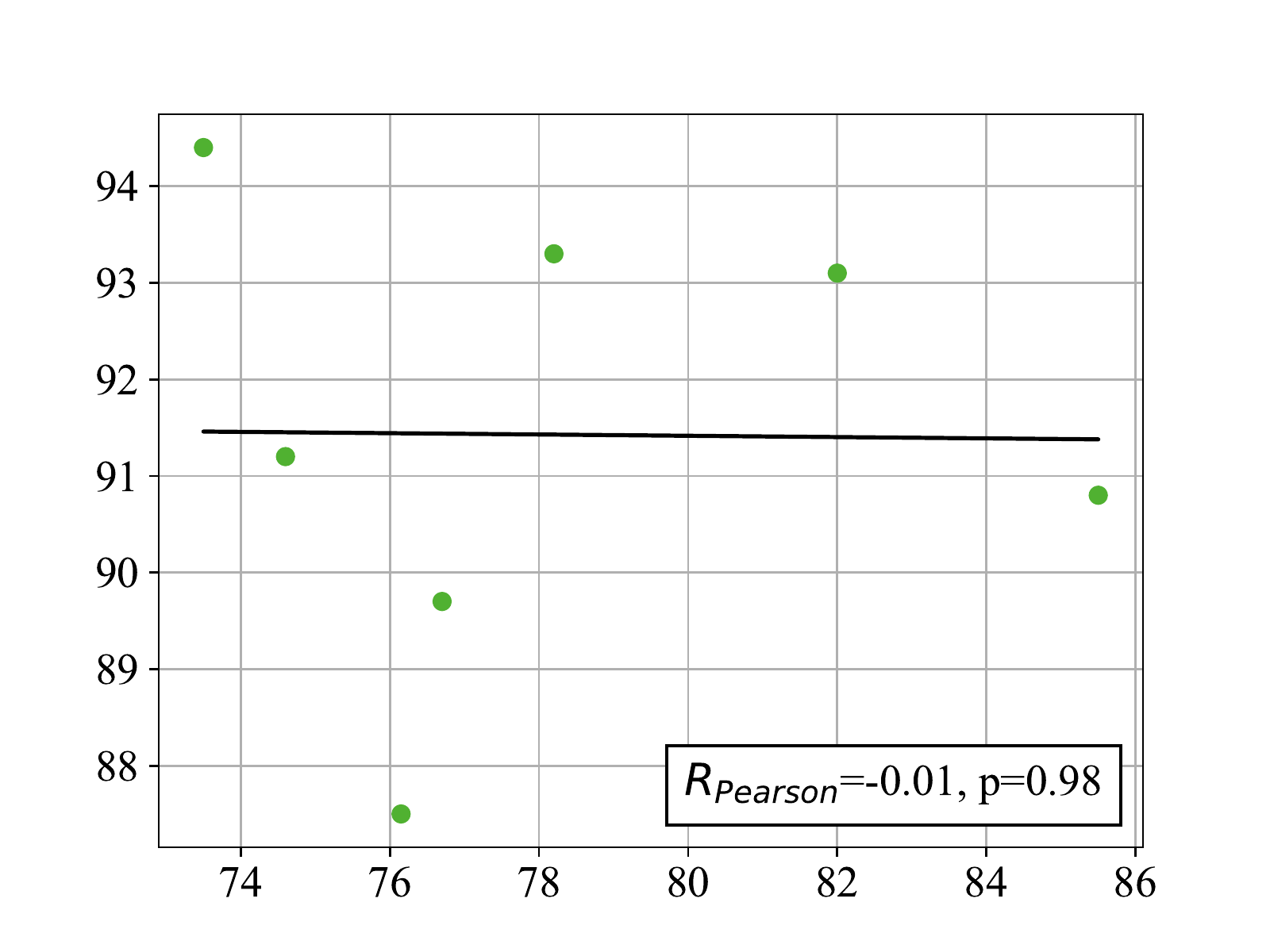}
    &
    \includegraphics[width=0.2\textwidth]{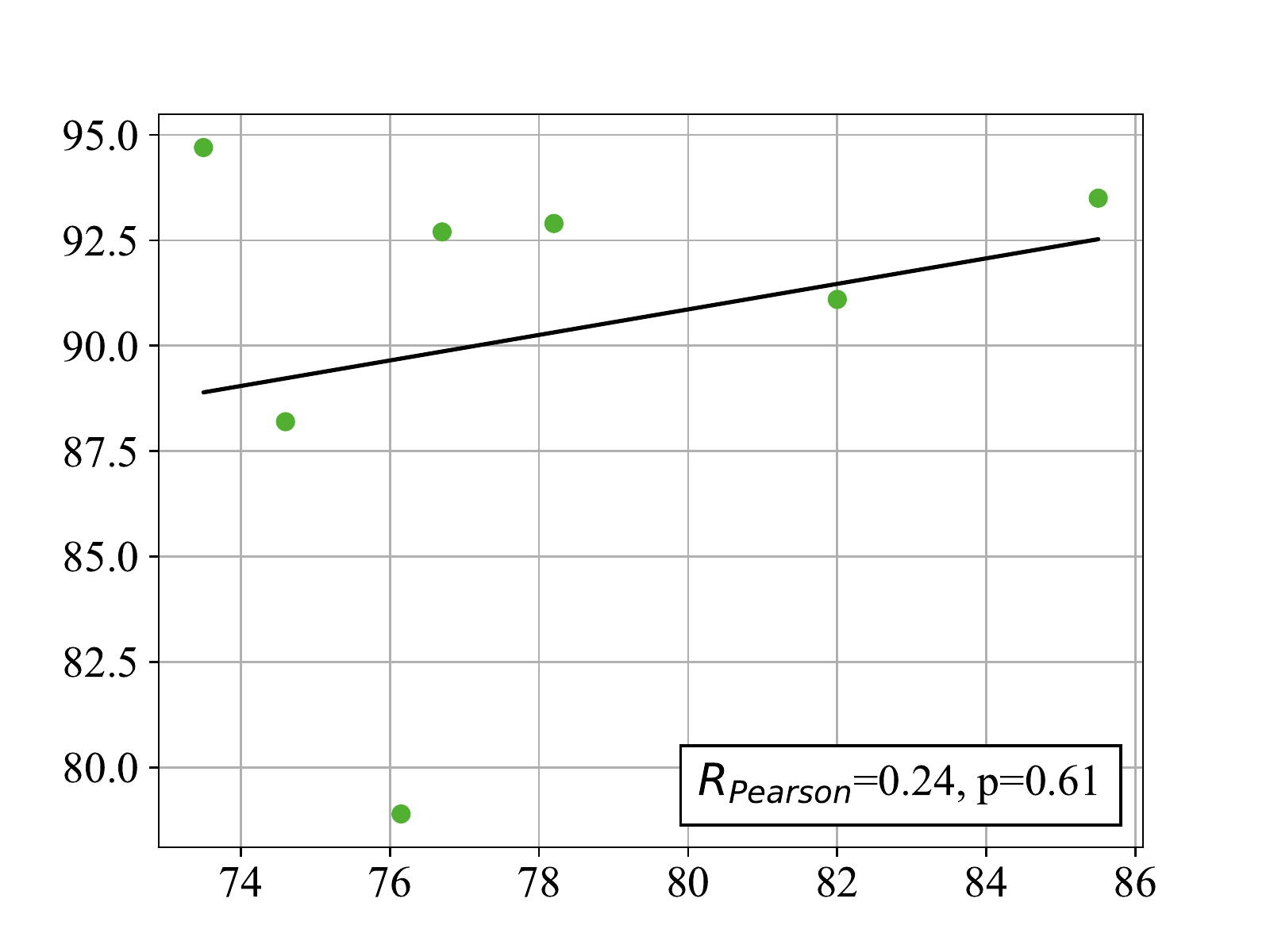}
    \\
    \rotatebox{90}{(4) iWildCam} & \includegraphics[width=0.2\textwidth]{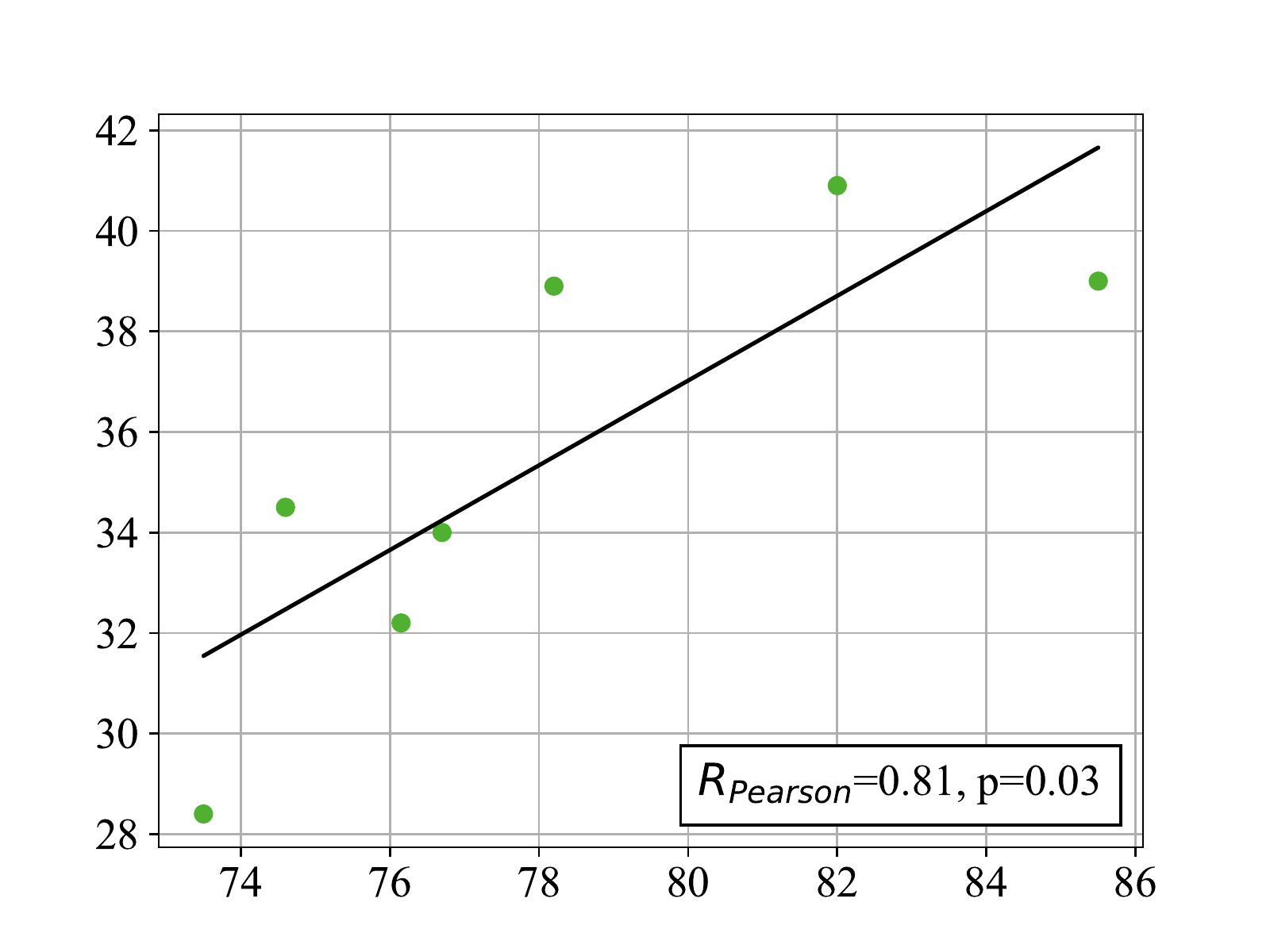}&
    \includegraphics[width=0.2\textwidth]{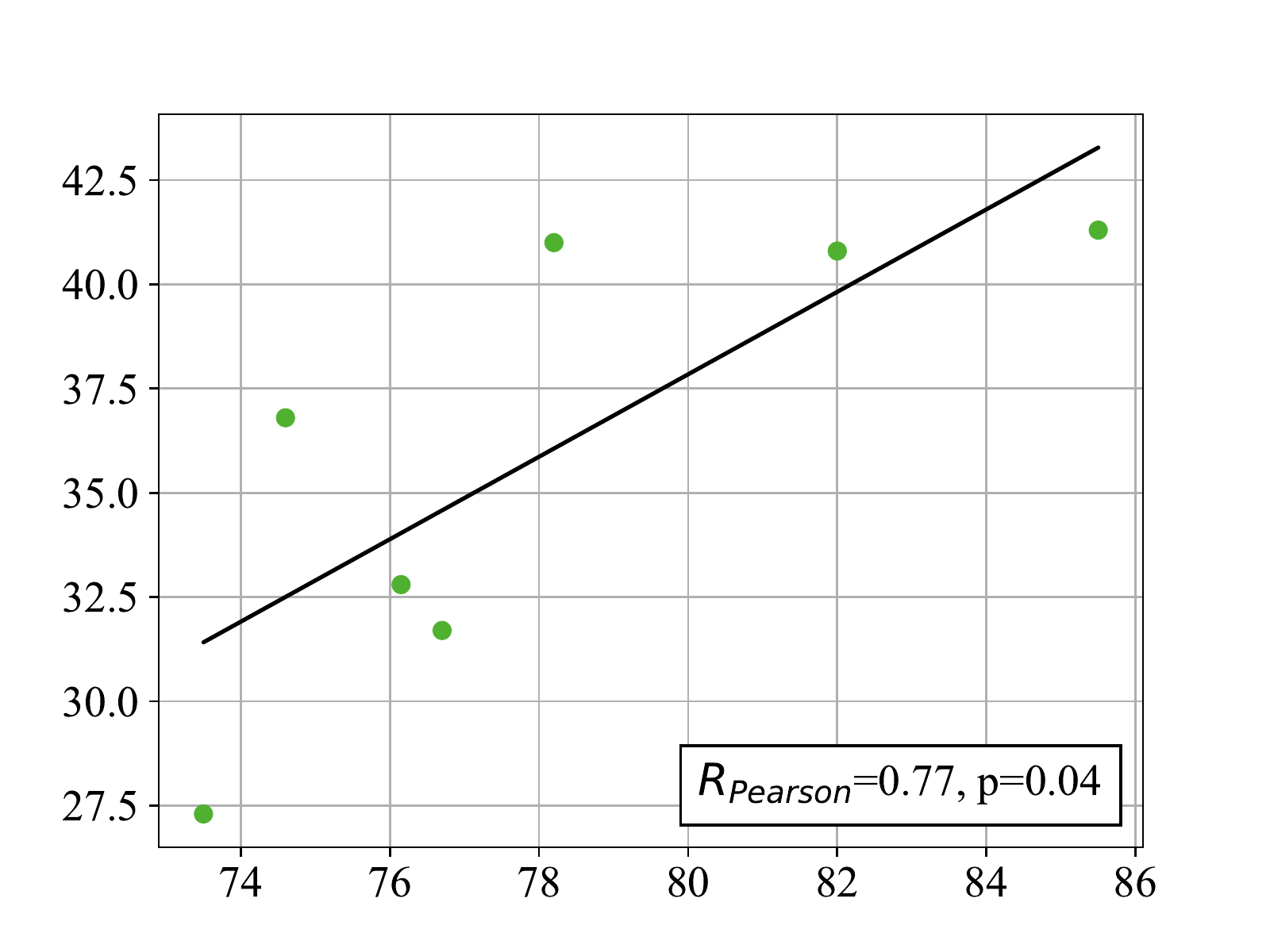}
    &
    \includegraphics[width=0.2\textwidth]{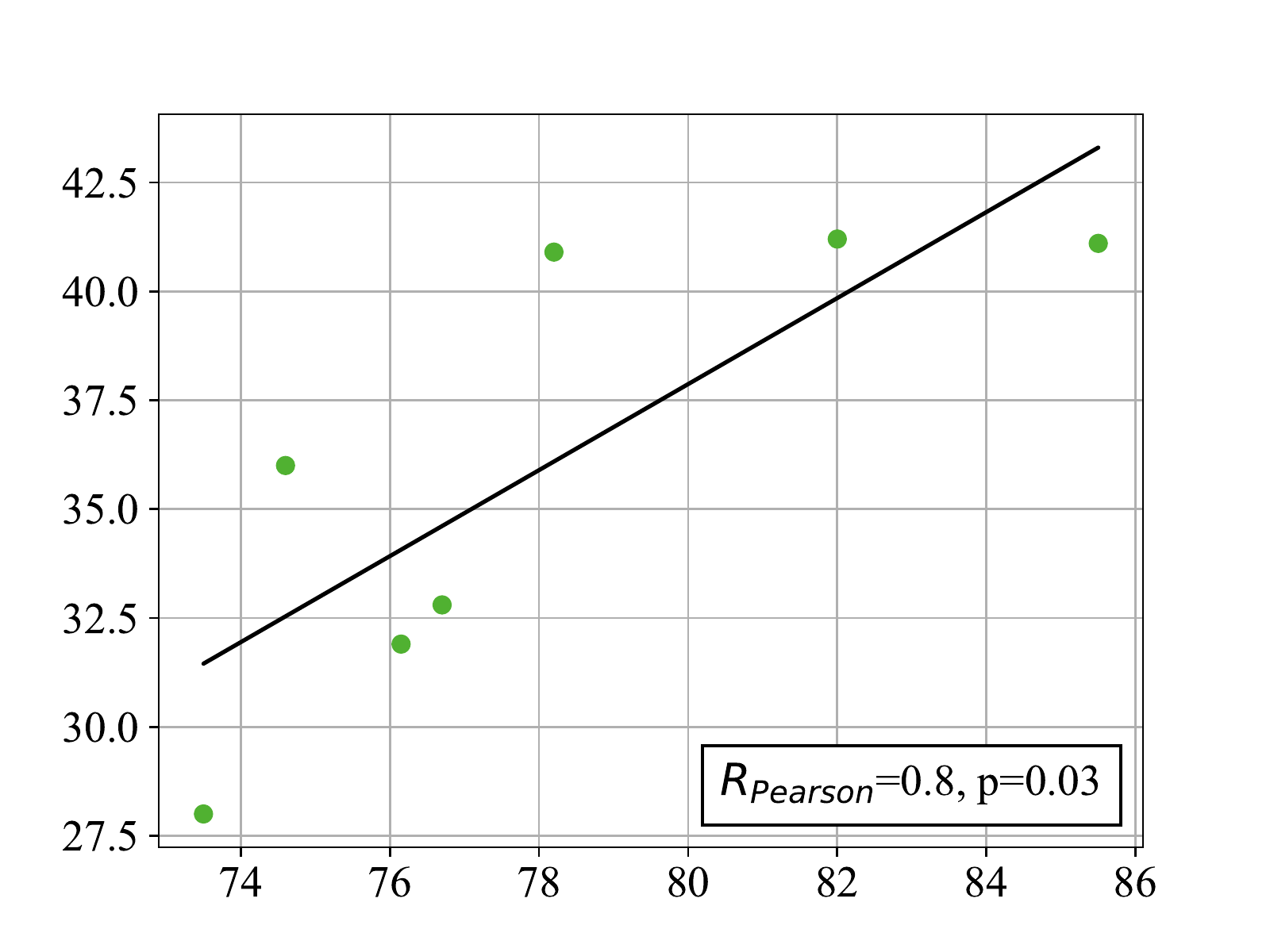}
    &
    \includegraphics[width=0.2\textwidth]{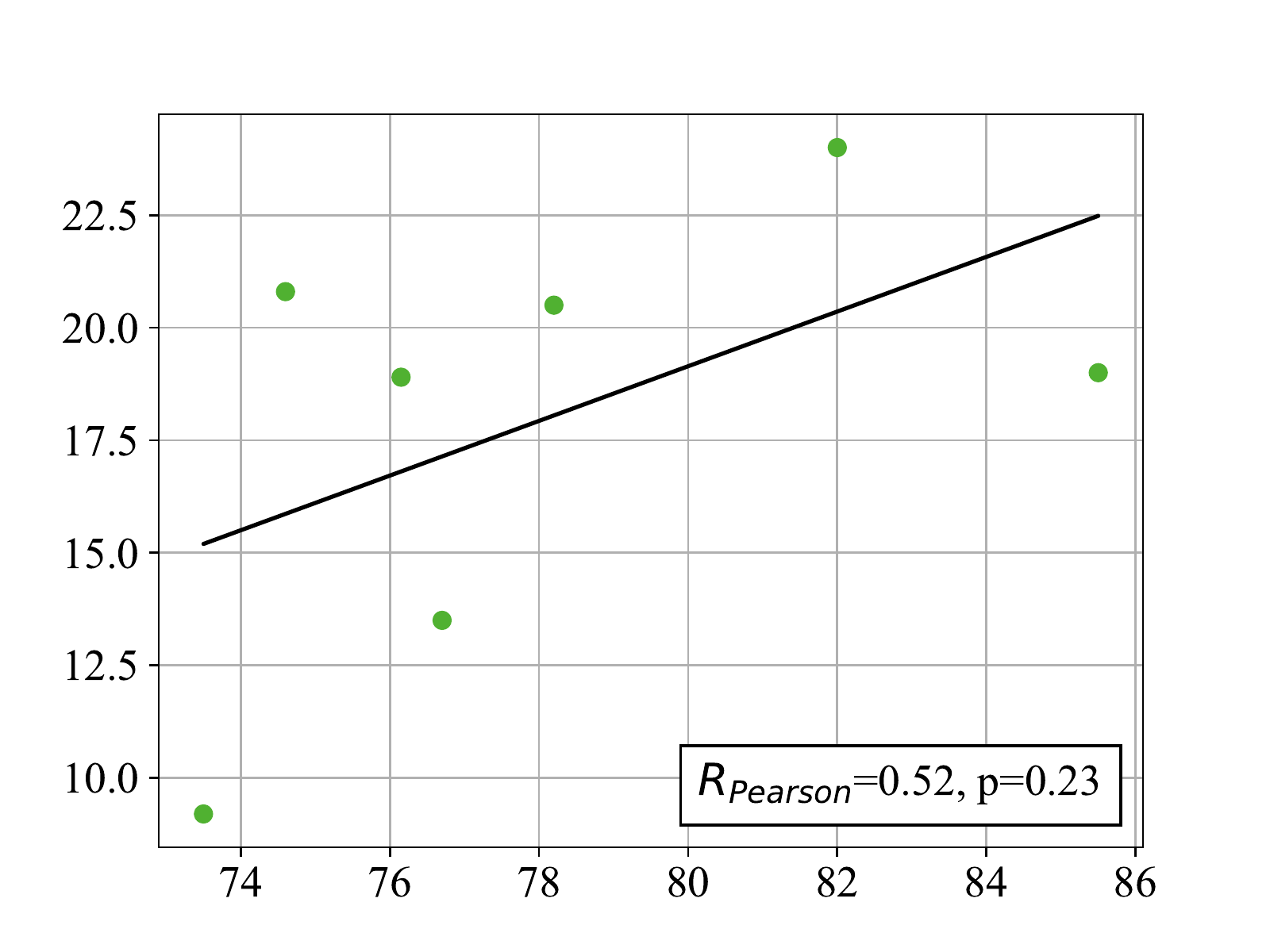}
    &
    \includegraphics[width=0.2\textwidth]{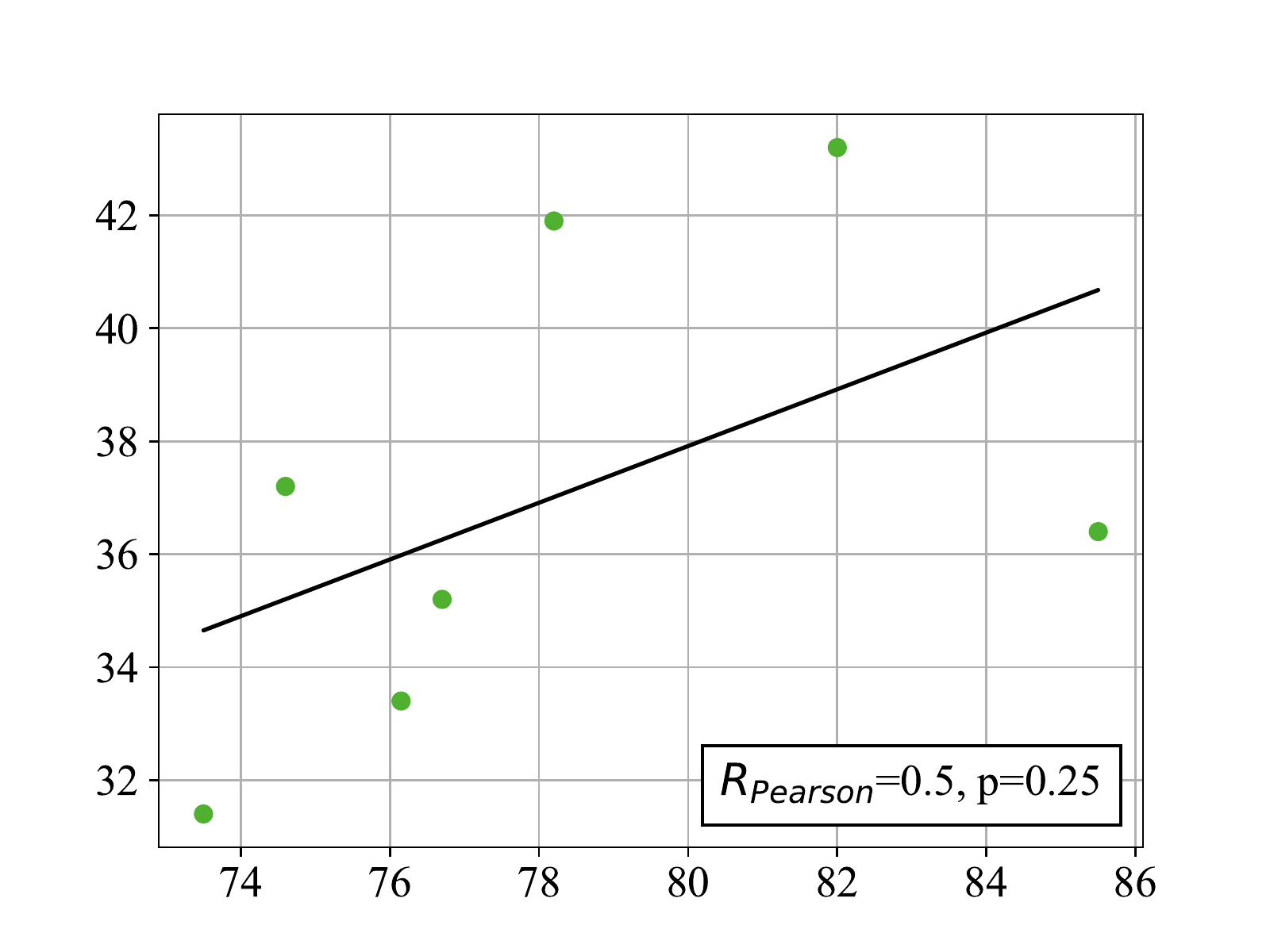}
    \\
    \rotatebox{90}{(5) DomainNet} & \includegraphics[width=0.2\textwidth]{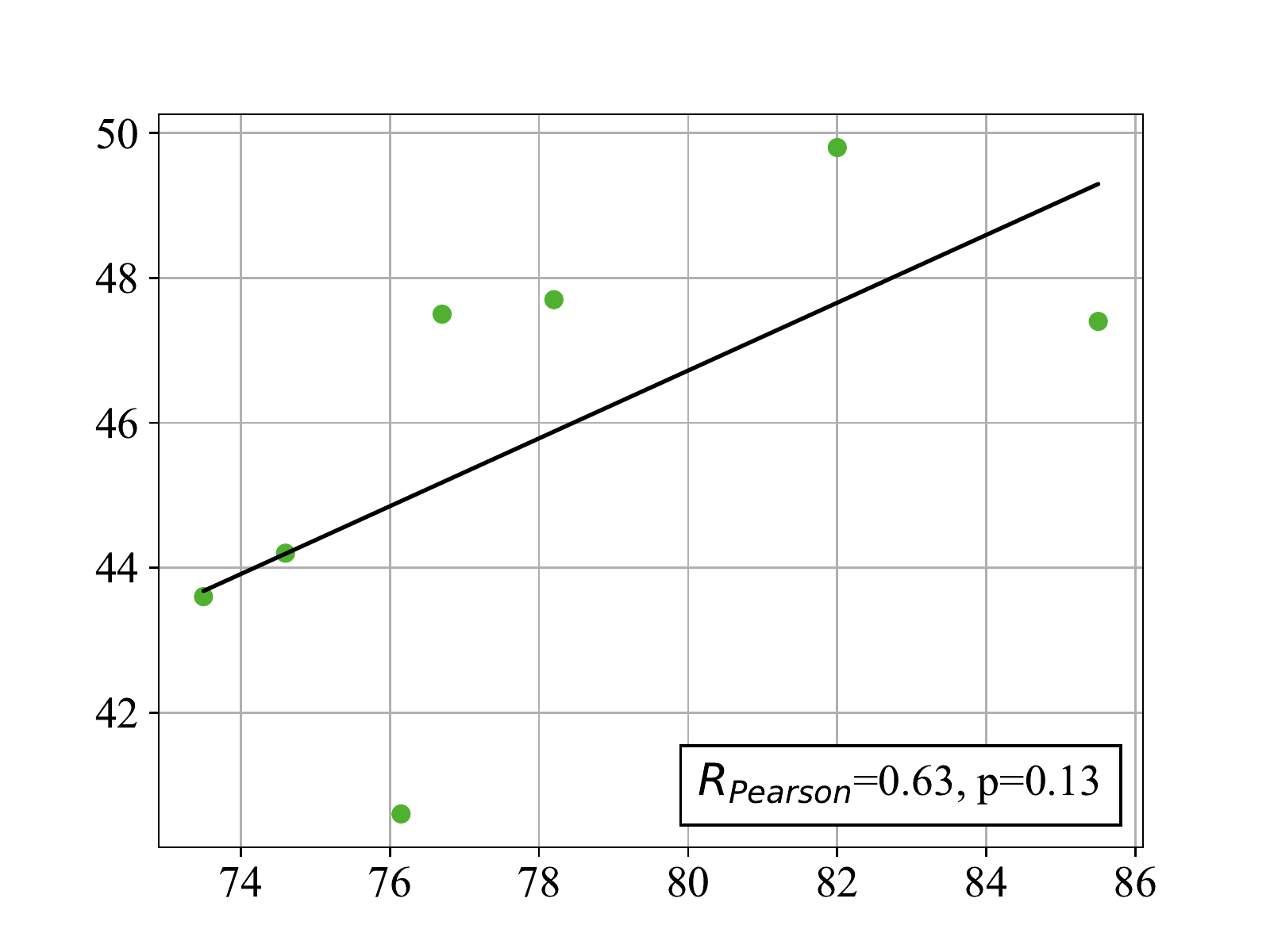}&
    \includegraphics[width=0.2\textwidth]{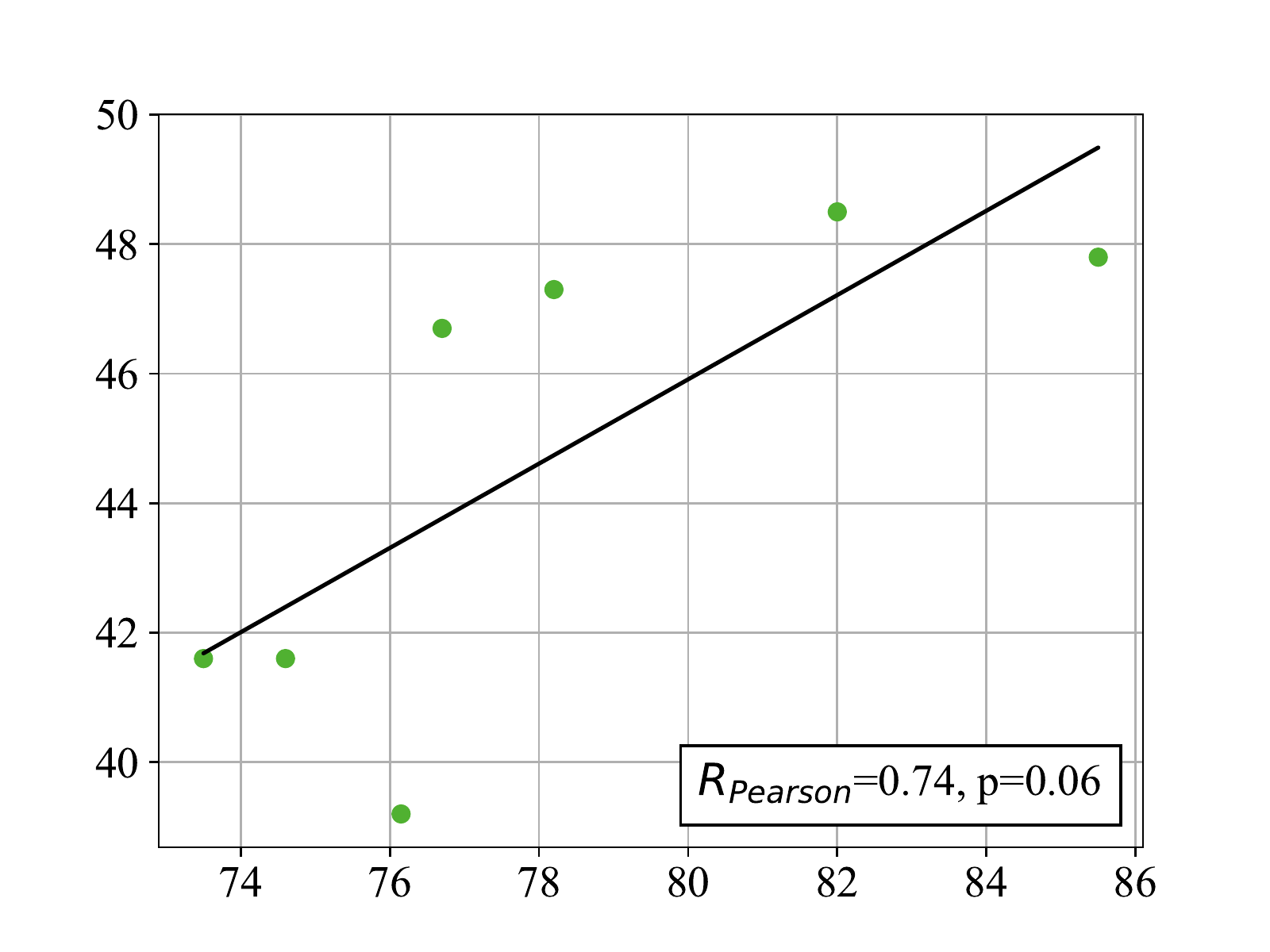}
    &
    \includegraphics[width=0.2\textwidth]{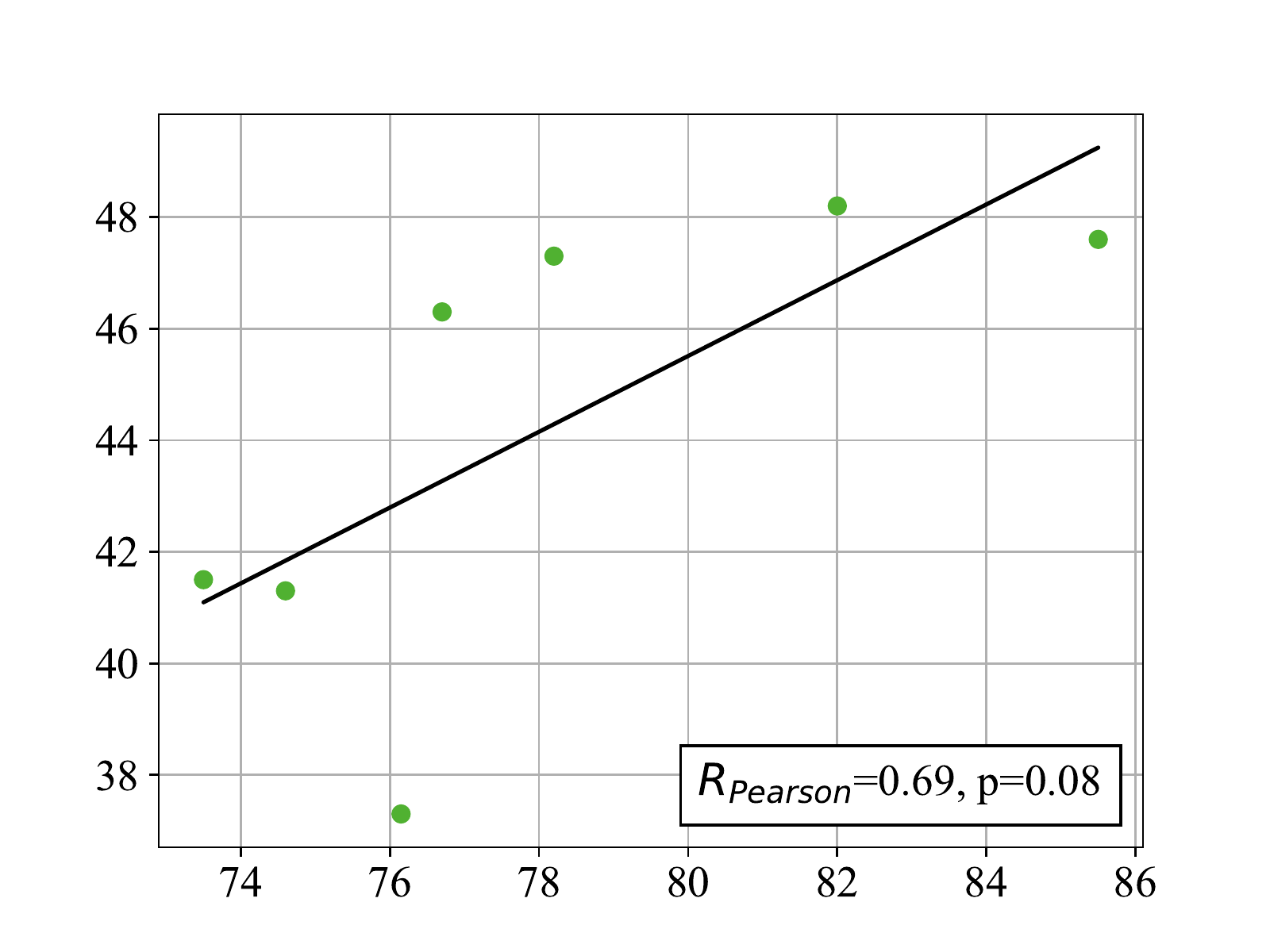}
    &
    \includegraphics[width=0.2\textwidth]{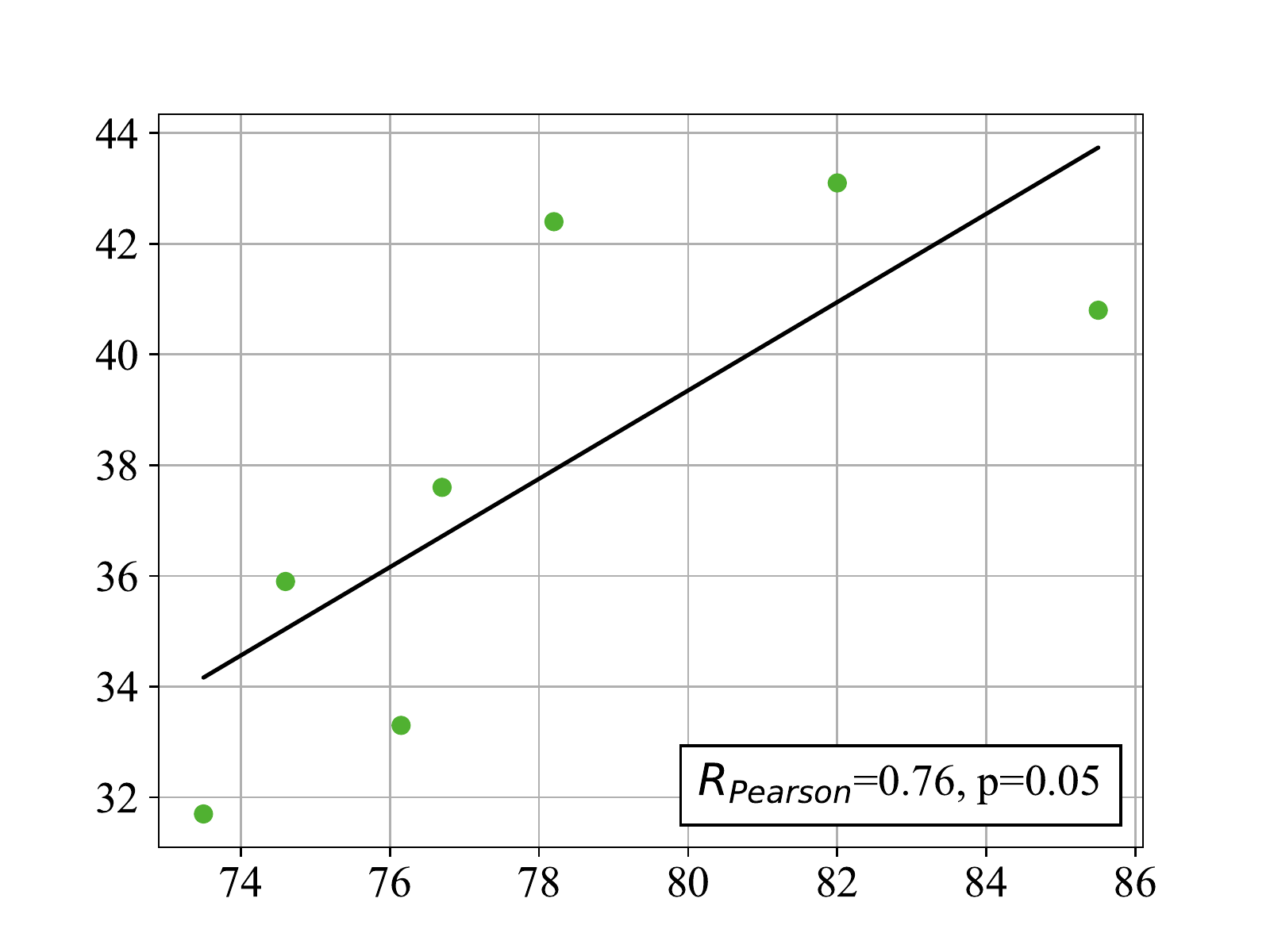}
    &
    \includegraphics[width=0.2\textwidth]{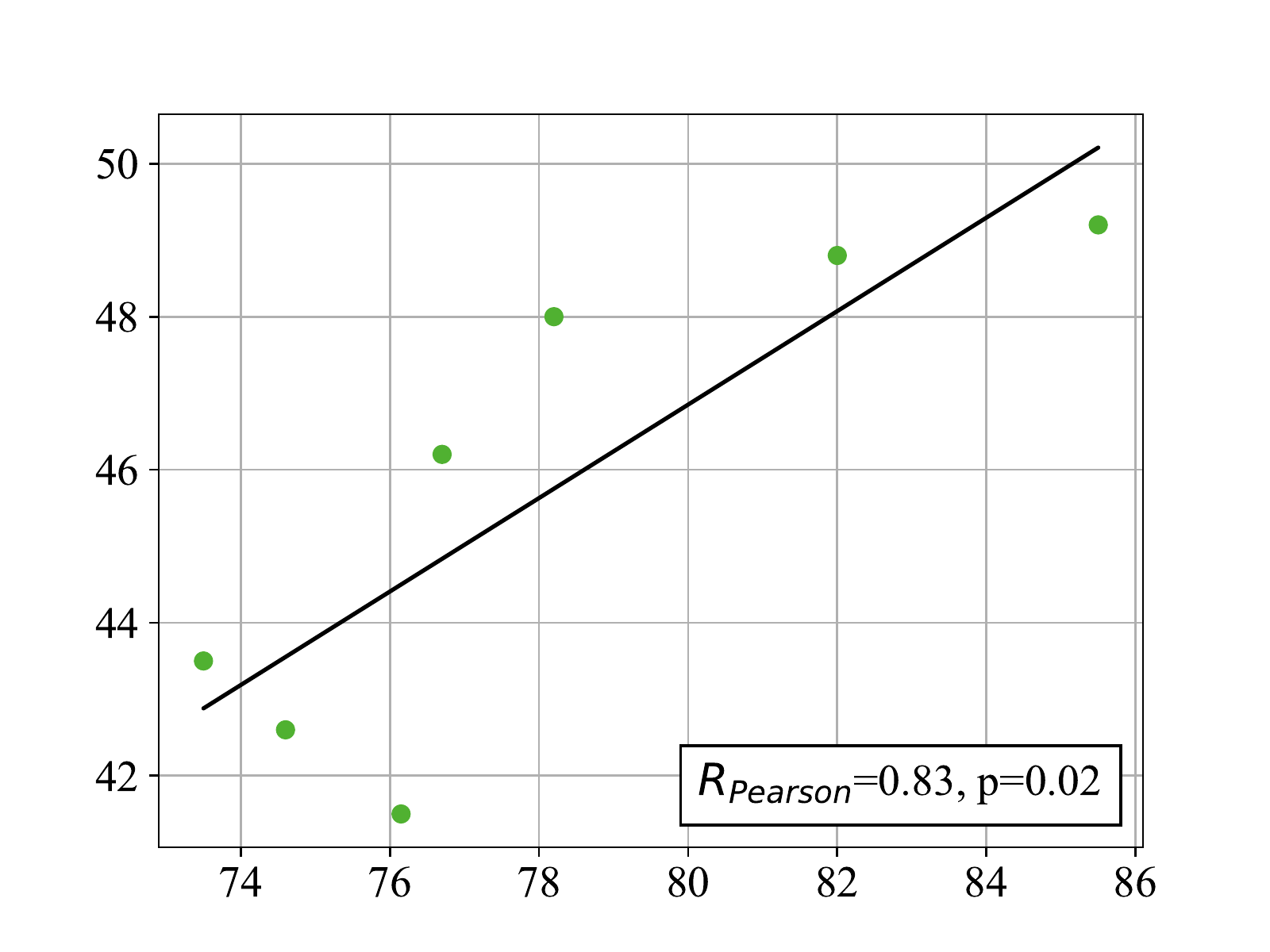}
  \end{tabular}
\end{center}
\vspace{-0.2cm}
  \caption{The correlation analysis between ImageNet accuracy of a pre-trained model and its downstream performance under distribution shift. The x-axis is IN accuracy and the y-axis is the performance of distribution shift in each dataset. The Pearson's and p value of the linear regression are shown in each figure. }
  \label{fig:correlation_analysis}
  \vspace{-0.6cm}
\end{figure*}

\textbf{iWildCam.} The input resolution is set as 448$\times$448, the training epoch as 12 and batch size as 24. The learning rate is searched. The WD is fixed as 0.0, according to existing experiment in \cite{koh2021wilds} and the learning rate is searched from \{0.000001, 0.000002, 0.000004, 0.000008, 0.00001, 0.00002, 0.00004, 0.00008,0.0001\}. The $\alpha$ in mixup is searched from \{0.1,0.2,0.4,1.0\}. For GroupMix, we search the constant $C$ from \{1.0,3.0,10.0,30.0\}. For CORAL, the $\lambda$ is searched from \{0.3,1.0,3.0\}. In the hyperparameter search stage, we train 225 models.  

%For GroupMix, search the learning rate from \{0.000001, 0.000002, 0.000004, 0.000008, 0.00001, 0.00002, 0.00004, 0.00008,0.0001\}, fix the WD to be 0, $\alpha$ from \{0.1,0.2,0.4,1.0\}, $C$ from \{1.0,3.0,10.0,30.0\}. For CORAL, the $\lambda$ in CORAL is searched from \{0.3,1.0,3.0\}. ERM: 9=, Mixup: 9*4=, GroupMix: 9*4*4=, GroupDRO: 9=, CORAL: 9*3=. \textbf{225}

\textbf{Camelyon.} The input resolution is set as 96$\times$96, the training epoch as 10 and batch size as 168. The WD and learning rate are searched from \{0.000001, 0.000003, 0.00001, 0.00003, 0.0001, 0.0003, 0.001, 0.003, 0.01\} and \{0.0,1e-5,1e-4,1e-3,1e-2\}. The $\alpha$ is searched from \{0.1,0.3,1.0\}. For GroupMix, $C$ is searched from \{3.0,10.0,30.0,100.0\}. For CORAL, the $\lambda$ is searched from \{0.3,1.0,3.0\}. So we trained 900 models in total for Camelyon for hyperparameter search.

%For GroupMix, search the learning rate from \{0.000001,0.000003,0.00001,0.00003,0.0001,0.0003,0.001,0.003,0.01\}, WD from \{0.0,1e-5,1e-4,1e-3,1e-2\}, $\alpha$ from \{0.1,0.3,1.0\}, $C$ from \{3.0,10.0,30.0,100.0\}. For CORAL, the $\lambda$ in CORAL is searched from \{0.3,1.0,3.0\}. ERM: 9*5=, Mixup: 9*5*3=, GroupMix: 9*5*3*4=, GroupDRO: 9*5=, CORAL: 9*5*3=. \textbf{900}

\textbf{DomainNet} Following \cite{gulrajani2021in}, the input resolution is set as 224$\times$224 and the training lasts for 5000 update steps. The batch size is 32 for each environment, resulting in a 32*5=160 samples for the whole batch. We search the learning rate from \{0.00001, 0.0001, 0.001\}, WD from \{0.0,1e-5,1e-4\}, $\alpha$ from \{0.1,0.3,1.0\}. The $C$ in GroupMix is searched from \{1.0,3.0,10.0\}, the $\lambda$ in CORAL is searched from \{0.3,1.0,3.0\}. This amounts to 810 trained models in this stage. 

Finally, we use the PyTorch \cite{paszke2019pytorch} code from WILDS \cite{koh2021wilds} and DomainBed \cite{gulrajani2021in} benchmark. All of our experiment is run on a single Nvidia-V100 GPU with 32GB. 

%For GroupMix, search the learning rate from \{0.00001,0.0001,0.001\}, WD from \{0.0,1e-5,1e-4\}, $\alpha$ from \{0.1,0.3,1.0\}, $C$ from \{1.0,3.0,10.0\}. For CORAL, the $\lambda$ in CORAL is searched from \{0.3,1.0,3.0\}. ERM: 3*3=, Mixup: 3*3*3=, GroupMix: 3*3*3*3=, GroupDRO: 3*3=, CORAL: 3*3=. \textbf{810}

\section{The full experimental results}
\vspace{-0.3cm}
Tab.~\ref{tab:waterbirds_fmow} shows the full result of Waterbirds and FMoW in our paper, including the averaged accuracy. On Waterbirds, it is worth noting that the averaged accuracy of GroupMix is generally higher than that of GroupDRO. On FMoW, even though the worse-group accuracy of CORAL is higher than GroupMix, the average accuracy of GroupMix is significantly higher than CORAL's. In conclusion, ERM with GroupMix is quite competitive in both worse-group accuracy and averaged performance. 

Tab.~\ref{tab:wilds_iwildcam_all_ood} shows the result of WILDS with ID and OOD Macro F1. To compare the difference between ID and OOD validation set based model selection, we report the result using ID validation set in the model selection in Tab.~\ref{tab:wilds_iwildcam_all_ood}.B. We observe that CORAL has the drawback of sacrificing the ID accuracy to achieve a high OOD accuracy, whil GroupMix does not reduce the ID accuracy compared with DA and Mixup. Thus, the ERM with data augmentation which uses group information is a strong baseline in both ID and OOD distribution. The table also shows that the two model selection methods do not differ at a significant level. For example, the top-1 models in Tab.~\ref{tab:wilds_iwildcam_all_ood}.A and B overlap in most learning algorithms. So the ID validation set based model selection will not change our general conclusion in the main paper.  

Tab.~\ref{tab:domainnet_full} reports the result of 6 domain generalization accuracy for each experimental setting. It is interesting that on the two most challenging domains, i.e., infograph and quickdraw, ViT and R50 supervised pre-trained on IN-21k have quite different performance. Sup-ViT-IN21k is better at generalizing to quickdraw, the most difficult domain, while Sup-R50-IN21k is better at infograph. This phenomenon indicates that the performance in domain generalization can be further improved by using an ensemble of different neural architectures. 

Fig.\ref{fig:correlation_analysis} shows the result of linear regression between ImageNet (IN) \cite{ILSVRC15} test accuracy of a pre-trained model and its performance in a target task with distribution shift. For self-supervised models, we use the linear probing result as the IN accuracy as reported in their paper \cite{he2021masked,chen2021empirical,he2020momentum}. Fig.\ref{fig:correlation_analysis} shows all the regression analyses of learning algorithms and datasets. For Waterbirds, the correlation between IN accuracy and downstream performance is statistically significant for GroupMix and CORAL. On iWildCam and DomainNet, the correlation is more obvious than on Waterbirds. However, on FMoW and Camelyon, there is no significant correlation between the two performances. This phenemenon further validates our hypothesis that for object recognition task, increasing the performance on the standard dataset (IN) is helpful for downstream tasks under distribution shift. But the benefit of IN performance is no longer valid if the downstream task has quite different visual features such as dense images in FMoW and Camelyon.

\end{document}